\documentclass[10pt,twocolumn,letterpaper]{article}

\usepackage{iccv}
\usepackage{times}
\usepackage{epsfig}
\usepackage{graphicx}
\usepackage{amsmath}
\usepackage{amssymb}

\usepackage{algorithm}
\usepackage{algpseudocode}

\usepackage{booktabs}
\usepackage{colortbl}
\usepackage[table, usenames, dvipsnames]{xcolor}
\usepackage{makecell}
\usepackage[disable]{todonotes}
\usepackage{caption}
\usepackage{subcaption}
\usepackage[sort&compress, square, numbers]{natbib}
\usepackage[accsupp]{axessibility}  

\usepackage[pagebackref=true,breaklinks=true,letterpaper=true,colorlinks,bookmarks=false]{hyperref}
\iccvfinalcopy

\ificcvfinal\fi

\begin{document}

\title{Which Tokens to Use? Investigating Token Reduction in Vision Transformers}

\author{Joakim Bruslund Haurum$^{1}$ \qquad Sergio Escalera$^{2,1}$ \qquad Graham W. Taylor$^{3}$ \qquad Thomas B. Moeslund$^{1}$\\[0.25em]
{\normalsize $^1$ Visual Analysis and Perception (VAP) Laboratory, Aalborg University \& Pioneer Centre for AI, Denmark}\\
{\normalsize$^2$ Universitat de Barcelona \& Computer Vision Center, Spain} \ \
{\normalsize$^3$ University of Guelph \& Vector Institute for AI, Canada}\\
{\tt\small joha@create.aau.dk, sescalera@ub.edu, gwtaylor@uoguelph.ca, tbm@create.aau.dk}
}

\maketitle
\ificcvfinal\thispagestyle{empty}\fi

\begin{abstract}\vspace{-3mm}
Since the introduction of the Vision Transformer (ViT), researchers have sought to make ViTs more efficient by removing redundant information in the processed tokens. 
While different methods have been explored to achieve this goal, we still lack understanding of the resulting reduction patterns and how those patterns differ across token reduction methods and datasets. To close this gap, we set out to understand the reduction patterns of 10 different token reduction methods using four image classification datasets.
By systematically comparing these methods on the different classification tasks, we find that the Top-K pruning method is a surprisingly strong baseline. Through in-depth analysis of the different methods, we determine that: the reduction patterns are generally not consistent when varying the capacity of the backbone model, the reduction patterns of pruning-based methods significantly differ from fixed radial patterns, and the reduction patterns of pruning-based methods are correlated across classification datasets. Finally we report that the similarity of reduction patterns is a moderate-to-strong proxy for model performance. Project page at \url{https://vap.aau.dk/tokens}.
\end{abstract}

\vspace{-3mm}
\section{Introduction}
The Vision Transformer (ViT) \cite{ViT_2021} has in record time seen wide spread adoption within computer vision, ousting Convolutional Neural Networks (CNNs). In order to better understand how ViTs function, prior works have investigated whether ViTs process data in a similar way as CNNs \cite{ViTSeeCNN}, and how different types of supervision affect ViT training \cite{ViTSUpervision}. In this work we investigate the use of \textit{token reduction methods}, which leverage the fact that ViTs can accommodate variable input sequence lengths. These methods aim to make ViTs more efficient by removing redundant tokens and thereby reduce the computational cost of the self-attention operation, which scales quadratically with the number of tokens \cite{ATS_2022, DyViT_2021, ToMe_2023}.

\begin{figure}
    \centering
    \includegraphics[width=\linewidth]{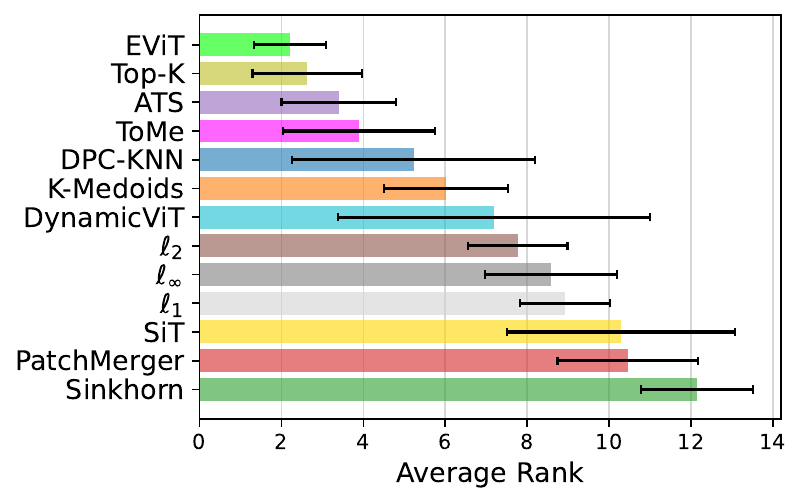}
    \caption{\textbf{Average method rank.} The average rank of each tested method plotted with $\pm$ 1 standard deviation. The Top-K pruning method and its extension, EViT, are found to be the best performing methods.}
    \label{fig:ranked}
\vspace{-3mm}
\end{figure}

However, with some limited exceptions, little has been done to gain deeper insights into how the token reduction process differs across methods or depends on hyperparameters such as backbone capacity or the number of tokens to be kept. Furthermore, the analysis of methods have primarily been constricted to the ImageNet dataset, and is rarely done with structured comparisons to other methods. Consequently, the community does not have a good understanding of how the methods differ from one another. We set out to rectify this by conducting a systematic comparison of 10 recently proposed methods, leveraging thorough experiments to elucidate the inner workings of reduction processes. In this work we make the following contributions:
\begin{itemize}
    \itemsep0em
    \item We conduct the first systematic comparison and analysis of 10 state-of-the-art token reduction methods across four image classification datasets, trained using a single codebase and consistent training protocol.
    \item We find that the Top-K and EViT methods are strong baselines across all datasets.
    \item Extensive experiments providing deeper insights into the core mechanisms of the token reduction methods.
    \item We find that the similarity in reduction patterns is a moderate-to-strong proxy for model performance.
\end{itemize}

\section{Related Works}
\noindent\textbf{Efficient Vision Transformers.}
Since the introduction of the Transformer \cite{Transformer_2017} and Vision Transformer \cite{ViT_2021} a large number of methods have been proposed to make the Transformer model more efficient. Within the computer vision domain several method paradigms have been investigated such as model pruning \cite{SViTE_2021, UPViT_2021, ViTSlim_2022, AdaVIT_2022, CP-ViT} and quantization \cite{QVIT_2022, FQViT_2022}, structured token downsampling inspired by the pooling layers in CNNs \cite{SWIN_2021, PiT_2021, LeViT_2021}, randomly dropping patches \cite{PatchDropout_2023, VATT_2021}, part selection modules \cite{TransFG_2022, MAWS_2021, HAS_2022, RAMSTrans_2021, R2Trans_2022}, and dynamic resizing of input patches \cite{DVT_2021, ReViT_2021, FlexiViT_2022, PSViT_2021, MSViT_2022, SuperViT_2022, CFVIT_2022}. Additionally, the Transformer model uniquely allows for variable input sequence lengths, enabling a new type of method, called token reduction, which operate directly on the token sequence. These methods are the focus of this work.

\noindent\textbf{Token Reduction Methods.} The prior work on sparsification of the input token sequence can be divided into two primary paradigms: token pruning \cite{EViT_2022, DyViT_2021, SPViT_2022, EVoViT_2022, DiversityEVIT_2022, ToRe_2022, ATS_2022, IARED2_2021, SaiT_2022, DPSViT_2022, AViT_2022} and token merging \cite{ SiT_2022, DPCKNN_2022, ToMe_2023, Marin_2023, GroupViT_2022, PatchMerger_2022, TokenLearner_2022, Sinkhorn_2022, CentroidViT_2021}. Pruning-based methods aim to reduce the token sequence by removing tokens, whereas merging-based methods reduce the token sequence by combining tokens. 

Pruning-based methods can be split into dynamic and static keep rate methods, depending on whether the method can dynamically choose how many tokens to prune. Static keep rate pruning methods select a predetermined number of tokens to keep by using the attention scores between the spatial tokens and the global class (CLS) token \cite{EViT_2022, DiversityEVIT_2022} or by predicting per-token keep probabilities \cite{DyViT_2021, SPViT_2022}. In order to not completely discard the information in the pruned tokens, which can contain contextual information regarding \eg location and background, several methods propose merging the pruned tokens into a single token \cite{EViT_2022, SPViT_2022, DiversityEVIT_2022, EVoViT_2022}.
On the other hand, dynamic keep rate pruning methods select an adaptive number of tokens to keep by using either sampling methods \cite{ATS_2022, AViT_2022}, reinforcement learning \cite{IARED2_2021}, or alternating training schemes \cite{SaiT_2022, DPSViT_2022}. 

Similarly, merging-based methods can be split into hard-merging \cite{Marin_2023, ToMe_2023, GroupViT_2022, DPCKNN_2022} and soft-merging \cite{Sinkhorn_2022, SiT_2022, PatchMerger_2022, CentroidViT_2021}, depending on whether the merging operation requires the token assignment to be mutually exclusive. Hard-merging methods typically use commonly known clustering methods such as K-Means \cite{Marin_2023}, K-Medoids \cite{Marin_2023}, and Density-Peak Clustering with K-Nearest Neighbours (DPC-KNN) \cite{DPCKNN_2022}. Other hard-merging based methods have used bipartite-matching of tokens \cite{ToMe_2023} and cross-attention between spatial tokens and learnable cluster centers, called queries \cite{GroupViT_2022}.
The soft-merging based approaches have primarily consisted of soft-clustering methods, which lead to a convex combination of spatial tokens derived from similarity with queries or the spatial tokens themselves \cite{Sinkhorn_2022, PatchMerger_2022, SiT_2022}.

To summarize, while many different token reduction methods have been proposed, scant attention has been given to comparing the methods in a systematic way, nor trying to better understand how the reduction process and reduction patterns (\ie constructed clusters and pruned tokens) are affected by the choice of reduction method, datasets, and model settings. In this work we aim to rectify this by conducting a thorough systematic comparison and in-depth analysis of contemporary token reduction methods.

\section{Experimental Design}
In order to compare the different token reduction methods fairly, we select representative methods from each reduction method paradigm; see Section~\ref{sec:methods}. The methods are chosen based on two criteria: 1) the selected methods should cover the key trends within each paradigm, and 2) the methods should be inserted into the backbone with minimal adjustments to the training loop.

Several ways of incorporating the token reduction operation have previously been used, ranging from a single reduction stage in the middle of the ViT \cite{PatchMerger_2022, SaiT_2022} to after each stage in the ViT \cite{ATS_2022, ToMe_2023, Marin_2023, AViT_2022}. However, the most common approach is to apply the token reduction operation at three predefined stages dividing the ViT into four sections of equal size \cite{EViT_2022, DPCKNN_2022, SPViT_2022, IARED2_2021, GroupViT_2022, DyViT_2021, DiversityEVIT_2022, SiT_2022}. This is the setting which we will follow. At each stage a ratio of the tokens, $r$, are kept for further processing, where $r\in\{0.25, 0.50, 0.70, 0.90\}$ is denoted the \textit{keep rate}.

\subsection{Methods}\label{sec:methods}
To ensure diversity of methods we select three representative and top-performing methods from each paradigm. For the Dynamic Keep Rate Pruning paradigm, however, we only select one as the other methods either did not converge during training with the training settings described in Section~\ref{sec:trainsets} (A-ViT \cite{AViT_2022}), or required substantial modifications to the training loop (IA-RED$^2$ \cite{IARED2_2021}, DPS-ViT \cite{DPSViT_2022}, and SaiT \cite{SaiT_2022}). Each method is described in Section~\ref{sec:StaticPrune}--\ref{sec:SoftClustering}. We also introduce a set of baseline pruning methods with a fixed image-centered radial pattern, see Section~\ref{sec:fixedPattern}. These are based on observations made by Yin \etal \cite{AViT_2022} and Rao \etal \cite{DyViT_2021} who found that the averaged reduction patterns display a radial pattern focused on the image center.
All methods are implemented in a single codebase based on official model implementations when possible.

\subsubsection{Fixed Pattern Pruning Baseline}\label{sec:fixedPattern}
We introduce a set of baseline methods with a fixed reduction pattern based on the distance of each token to the center of the image, measured using the $\ell_p$-norm. Specifically, we consider fixed patterns created by using the $\ell_1$, $\ell_2$, and $\ell_\infty$ norms. In order to create the fixed patterns such that only $r^s$ tokens are kept at reduction stages $s\in\{1,2,3\}$, we prune tokens based on their distance to the image center, setting the threshold such that the absolute difference between the kept tokens and $Pr^s$ is minimized, where $P$ is the initial amount of spatial tokens.

\subsubsection{Static Keep Rate Pruning}\label{sec:StaticPrune}
\noindent\textbf{Top-K} is a commonly used pruning baseline, where the attention between the $P$ spatial tokens and the CLS token is used. At reduction stage $s$ the method simply selects the $K_s$ most attended to tokens, where $K_s=Pr^s$.

\noindent\textbf{EViT} \cite{EViT_2022} extends Top-K pruning by creating a single ``fused'' token at each stage $s$. The fused token is computed  by averaging the $P_s-K_s$ pruned tokens weighted by their CLS token attention scores, where $P_s$ is the number of tokens at stage $s$ before the reduction is applied.

\noindent\textbf{DynamicViT} \cite{DyViT_2021} prunes the tokens by constructing a binary decision mask $\mathbf{D}_s$ based on keep probabilities produced by a small prediction module. The Gumbel-Softmax trick \cite{GumbelSoftmax} is used to ensure training is differentiable, while during inference the $K_s$ most probable tokens are kept. An extra loss is needed to ensure that $\mathbf{D}_s$ only keeps $K_s$ tokens.

\subsubsection{Dynamic Keep Rate Pruning}
\noindent\textbf{ATS} \cite{ATS_2022} is a sampling-based pruning method which selects a variable amount of tokens at each reduction stage. This is achieved by applying the inverse transform sampling (ITS) on the cumulative distribution function (CDF) of the CLS token attention scores and uniformly sampling the CDF $Pr^s$ times. In case a token is assigned a high attention score by the CLS token it may be sampled multiple times by the ITS operation, in which case only a single copy is kept. Thereby, ATS can sample fewer than $Pr^s$ tokens at stage $s$.

\subsubsection{Hard-Merging}

\noindent\textbf{ToMe} \cite{ToMe_2023} is a recent token merging method, where the set of tokens are split into a bipartite graph with equal sized partitions $A$ and $B$, where edges are constructed by drawing a single edge for each node in $A$ to the node in $B$ with the highest cosine similarity. The $P_s(1-r^s)$ highest valued edges are kept and connected nodes are merged by averaging the token features, followed by combining set $A$ and $B$ again. It should be noted that the ToMe method is constrained such that $r$ cannot be below 50\% as nodes in set $A$ are only allowed a single edge. In order to align the nomenclature with clustering methods, we denote the output of the ToMe method as cluster centers.

\noindent\textbf{K-Medoids} \cite{Marin_2023} is an iterative hard-clustering baseline where the $Pr^s$ cluster centers are set to be the cluster element which minimizes the $\ell_2$ distance to all other elements in the cluster. The method iteratively updates the clusters by assigning tokens to the cluster with the closest cluster center. We initialize the cluster centers based on the CLS token attention scores as proposed by Marin \etal \cite{Marin_2023}. 

\noindent\textbf{DPC-KNN} \cite{DPCKNN_2016} is a two-step clustering approach, where first the density of each token is computed based on the distance to the $k$-nearest neighbours, followed by determining the minimum distance to a point with higher density. The two measures are combined and the cluster centers are defined to be the $Pr^s$ tokens with the highest combined scores. The final cluster representation is obtained by averaging the elements assigned to the cluster. Zeng \etal \cite{DPCKNN_2022} proposed to add a small linear layer which predicts the importance for each token, making the cluster representation a weighted average of the cluster elements.

\subsubsection{Soft-Merging}\label{sec:SoftClustering}
\noindent\textbf{SiT} \cite{SiT_2022} is a recent soft-clustering method, where a small network predicts an assignment matrix $\mathbf{A}_s$, representing a convex combination of the $Pr^{s-1}$ input tokens to construct $Pr^s$ clusters. Specifically $\mathbf{X}_s = \mathbf{X}_{s-1}\mathbf{A}_s$, where $\sum_{i=1}^{Pr^s}\mathbf{A}_s[i,j] = 1$ for $j=1,2,\ldots,{Pr^{s-1}}$ and $\mathbf{X}$ is the token feature representations.

\noindent\textbf{Sinkhorn} \cite{Sinkhorn_2022} is a query-based clustering method, where unlike in SiT, the cluster centers, called queries, are randomly initialized learnable vectors. The assignment matrix is constructed by applying the Sinkhorn-Knopp algorithm on the cosine similarities between the tokens and queries. 

\noindent\textbf{PatchMerger} \cite{PatchMerger_2022} is a query-based clustering method, similar to Sinkhorn, where the assignment matrix is constructed by calculating the dot product between the queries and tokens. This is followed by a softmax operation to ensure the assignment matrix results in a convex combination.

 \subsection{Datasets}
 Previously, methods have only been tested on the ImageNet \cite{ILSVRC} classification dataset and primarily against the backbone model with no token reduction methods. In order to gain diverse insights into the methods, we analyse and compare the different token reduction methods using four distinct classification datasets: ImageNet, NABirds \cite{NABirds}, COCO \cite{COCO}, and NUS-WIDE \cite{NUS_WIDE}.
 
\begin{figure}[!htp]
     \centering
     \begin{subfigure}[b]{0.23\textwidth}
         \centering
         \includegraphics[width=\textwidth]{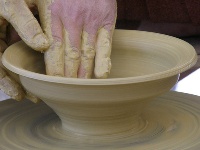}
         \caption{ImageNet}
     \end{subfigure}
     \hfill
     \begin{subfigure}[b]{0.23\textwidth}
         \centering
         \includegraphics[width=\textwidth]{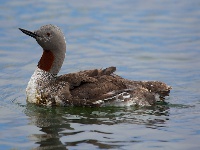}
         \caption{NABirds}
     \end{subfigure}
     \hfill
     \begin{subfigure}[b]{0.23\textwidth}
         \centering
         \includegraphics[width=\textwidth]{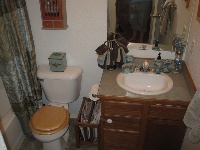}
         \caption{COCO}
     \end{subfigure}
     \hfill
     \begin{subfigure}[b]{0.23\textwidth}
         \centering
         \includegraphics[width=\textwidth]{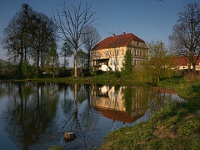}
         \caption{NUS-WIDE} 
     \end{subfigure}
        \caption{\textbf{Dataset examples.} Randomly selected images from the four considered image classification datasets.}
        \label{fig:dataset}
\end{figure}

 ImageNet and NABirds are used to evaluate the commonly used multi-class classification task. ImageNet is the most commonly used vision classification dataset, consisting of 1000 diverse classes across 1.2 million images. In contrast, the NABirds dataset represents a much more fine-grained classification task, consisting of 48,000 images and 555 bird classes. We also compare methods on the COCO and NUS-WIDE multi-label classification tasks, representing the case where more than one class of interest can be present simultaneously. COCO and NUS-WIDE consists of 80--81 classes of common object and animals across 122k to 220k images, respectively. In contrast to ImageNet, where the object of interest is often in the center of the image, the NABirds, COCO, and NUS-WIDE represent scenarios where the distinguishing attributes are not necessarily in the center of the image, or there may be more than one object of interest, respectively. Example images of each dataset are shown in Fig.~\ref{fig:dataset}. Classification performance on ImageNet and NABirds is measured using the standard Top-1 accuracy metric \cite{ILSVRC, TransFG_2022}, and for COCO and NUS-WIDE we report the mean Average Precision score (mAP) \cite{ASL}.

 \subsection{Training Details}\label{sec:trainsets}
 All methods are inserted into an DeiT backbone pre-tained on ImageNet without distillation \cite{DeiT_2021} at the 4th, 7th, and 10th transformer blocks. The DeiT backbone was chosen as it is the most commonly used throughout the token reduction literature. We consider both the Tiny, Small, and Base DeiT backbone capacities, denoted DeiT-$\{\text{T}, \text{S}, \text{B}\}$, respectively. 

 For all methods we based our hyperparameter settings on those presented by Rao \etal \cite{DyViT_2021}. A hyperparameter search over the learning rate warmup period, backbone learning rate scaler, and backbone freeze period was initially conducted on the ImageNet dataset using the DeiT-S backbone, training for 30 epochs. The best performing setting for each $r$ was used for training the DeiT-T and DeiT-B variants. It should be noted that for the DynamicViT and SiT methods we do not include the distillation losses used in the original papers, as we find that the effect is minimal and instead choose to keep the training procedure consistent.
 
 For NABirds, COCO, and NUS-WIDE, we fine-tuned the DeiT-S baseline in a similar manner, and for each token reduction method compared the ImageNet hyperparameter setting and fine-tuned DeiT-S hyperparameters. The best set of hyperparameters was used to train the DeiT-T and DeiT-B variants. The NABirds models were trained for 50 epochs with minimal augmentation and no label smoothing, following the guidelines from He \etal \cite{TransFG_2022}. COCO and NUS-WIDE models were trained for 40 epochs with the Asymmetric Loss \cite{ASL} following the multi-label classification guidelines from Ben-Baruch \etal \cite{ASL}. The specific hyperparameter values can be found in the supplementary materials (\supp).

 Lastly, it should be noted that for all datasets we trained the models at a resolution of $224\times224$. This is non-standard for the NABirds, COCO, and NUS-WIDE datasets (normally $448\times 448$). This choice was made to keep reduction patterns comparable, and because the aim was not to push the state-of-the-art in accuracy, but rather to train a set of models from which we can gain deeper insights.
\begin{table*}[!htp]
\caption{\textbf{Performance of Token Reduction methods with DeiT-S backbone.} Model performance is measured across varying keep rates, $r$, denoted in percentage of tokens kept at each reduction stage. Scores exceeding the DeiT baseline are noted in \textbf{bold}, measured in Top-1 accuracy for ImageNet \& NABirds and mean Average Precision for COCO \& NUS-WIDE. The three best performing methods per keep rate are denoted in descending order with \colorbox{red!25}{red}, \colorbox{orange!25}{orange}, and \colorbox{yellow!25}{yellow}, respectively. Similarly, the three worst performing methods are denoted in descending order with \colorbox{blue!10}{light blue}, \colorbox{blue!20}{blue}, and \colorbox{blue!30}{dark blue}. Results with the DeiT-B and DeiT-T backbones are available in the \supp}
\label{tab:perfSmall}
\centering
\resizebox{\textwidth}{!}{%
\begin{tabular}{@{}lcccclcccclcccclcccc@{}}
\toprule
 & \multicolumn{4}{c}{ImageNet} &  & \multicolumn{4}{c}{NABirds} &  & \multicolumn{4}{c}{COCO} &  & \multicolumn{4}{c}{NUS-WIDE} \\ \cmidrule(lr){2-5} \cmidrule(lr){7-10} \cmidrule(lr){12-15} \cmidrule(lr){17-20}
DeiT-S & \multicolumn{4}{c}{79.85} & & \multicolumn{4}{c}{80.57} & & \multicolumn{4}{c}{78.11} & & \multicolumn{4}{c}{63.23} \\ \midrule\midrule
$r$  (\%) & 25 & 50 & 70 & 90 &  & 25 & 50 & 70 & 90 &  & 25 & 50 & 70 & 90 &  & 25 & 50 & 70 & 90 \\ \midrule

$\ell_1$ & 70.05 & 74.47 & \cellcolor{blue!10}77.25 & 79.17 &  & 62.90 & 70.52 & 77.10 & 80.08 &  & 61.28 & 69.49 & 74.31 & 77.09 &  & \cellcolor{blue!10}54.44 & 59.60 & 61.98 & 62.91 \\
$\ell_2$ & 70.54 & 74.86 & 77.41 & 79.27 &  & 64.28 & 72.09 & 77.53 & 80.11 &  & 62.23 & 70.30 & 74.66 & 77.19 &  & 55.31 & 60.22 & 62.07 & 62.71 \\
$\ell_\infty$ & 70.58 & \cellcolor{blue!20}74.03 & 77.48 & 79.23 &  & 63.36 & 70.19 & 77.23 & 79.96 &  & 61.50 & 69.11 & 74.73 & 77.27 &  & 55.10 & 59.34 & 62.11 & 62.77 \\ \midrule
Top-K & 72.91 & 77.82 & \cellcolor{yellow!25}79.22 & \cellcolor{orange!25}\textbf{79.87} &  & \cellcolor{orange!25}76.28 & \cellcolor{orange!25}80.38 & \cellcolor{yellow!25}\textbf{80.70} & \textbf{80.60} &  & \cellcolor{yellow!25}70.14 & \cellcolor{red!25}75.84 & \cellcolor{red!25}77.50 & \cellcolor{red!25}78.09 &  & 59.32 & \cellcolor{yellow!25}61.98 & \cellcolor{orange!25}62.69 & \cellcolor{red!25}\textbf{63.26} \\
EViT & \cellcolor{yellow!25}74.17 & \cellcolor{orange!25}78.08 & \cellcolor{orange!25}79.30 & \cellcolor{orange!25}\textbf{79.87} &  & \cellcolor{red!25}76.74 & \cellcolor{yellow!25}80.28 & \cellcolor{orange!25}\textbf{80.73} & \cellcolor{yellow!25}\textbf{80.64} &  & \cellcolor{orange!25}71.28 & \cellcolor{orange!25}75.78 & \cellcolor{red!25}77.50 & \cellcolor{orange!25}78.07 &  & \cellcolor{yellow!25}59.69 & 61.89 & \cellcolor{yellow!25}62.67 & \cellcolor{orange!25}\textbf{63.25} \\
DynamicViT & \cellcolor{blue!30}60.32 & 77.84 & 79.17 & \cellcolor{yellow!25}79.79 &  & 70.60 & \cellcolor{red!25}\textbf{80.62} & \cellcolor{red!25}\textbf{80.77} & \cellcolor{red!25}\textbf{80.84} &  & \cellcolor{blue!30}39.18 & 69.02 & 75.43 & 77.69 &  & \cellcolor{blue!30}39.20 & \cellcolor{blue!20}57.83 & 61.96 & 63.16 \\
ATS & 72.95 & 77.86 & 79.09 & 79.63 &  & \cellcolor{yellow!25}73.46 & 78.89 & 80.36 & 80.55 &  & 70.13 & 75.66 & \cellcolor{yellow!25}77.23 & 77.83 &  & \cellcolor{orange!25}60.20 & \cellcolor{red!25}62.35 & \cellcolor{red!25}62.93 & \cellcolor{yellow!25}63.18 \\ \midrule
ToMe & - & \cellcolor{red!25}78.29 & \cellcolor{red!25}79.63 & \cellcolor{red!25}\textbf{79.92} &  & - & 74.99 & 80.05 & \cellcolor{orange!25}\textbf{80.68} &  & - & 74.99 & \cellcolor{orange!25}77.36 & 77.88 &  & - & 61.51 & 62.50 & 62.89 \\
K-Medoids & 68.94 & 76.44 & 78.74 & 79.73 &  & 65.28 & 76.95 & 79.75 & 80.46 &  & 66.26 & 74.15 & 76.76 & \cellcolor{yellow!25}77.94 &  & 57.78 & 61.48 & 62.47 & 63.12 \\
DPC-KNN & \cellcolor{red!25}75.01 & \cellcolor{yellow!25}77.95 & 78.85 & 79.54 &  & 68.77 & 74.14 & 76.70 & 78.88 &  & \cellcolor{red!25}72.15 & \cellcolor{yellow!25}75.70 & 77.06 & 77.74 &  & \cellcolor{red!25}60.78 & \cellcolor{orange!25}62.11 & \cellcolor{yellow!25}62.67 & 62.93 \\ \midrule
SiT & \cellcolor{orange!25}74.65 & 77.16 & 77.52 & \cellcolor{blue!10}77.71 &  & \cellcolor{blue!10}62.82 & \cellcolor{blue!10}62.02 & \cellcolor{blue!20}60.72 & \cellcolor{blue!20}58.50 &  & \cellcolor{blue!10}57.65 & \cellcolor{blue!20}57.33 & \cellcolor{blue!20}57.11 & \cellcolor{blue!20}57.13 &  & 57.95 & \cellcolor{blue!10}58.84 & \cellcolor{blue!20}59.29 & \cellcolor{blue!20}59.59 \\
PatchMerger & \cellcolor{blue!10}69.44 & \cellcolor{blue!10}74.17 & \cellcolor{blue!20}75.80 & \cellcolor{blue!20}76.75 &  & \cellcolor{blue!30}47.26 & \cellcolor{blue!20}61.34 & \cellcolor{blue!10}65.45 & \cellcolor{blue!10}68.24 &  & 62.24 & \cellcolor{blue!10}68.09 & \cellcolor{blue!10}70.75 & \cellcolor{blue!10}72.12 &  & 55.82 & 59.27 & \cellcolor{blue!10}60.46 & \cellcolor{blue!10}61.20 \\
Sinkhorn & \cellcolor{blue!20}64.26 & \cellcolor{blue!30}64.07 & \cellcolor{blue!30}64.02 & \cellcolor{blue!30}64.09 &  & \cellcolor{blue!20}48.89 & \cellcolor{blue!30}50.19 & \cellcolor{blue!30}51.46 & \cellcolor{blue!30}51.22 &  & \cellcolor{blue!20}56.93 & \cellcolor{blue!30}56.68 & \cellcolor{blue!30}56.85 & \cellcolor{blue!30}56.65 &  & \cellcolor{blue!20}50.59 & \cellcolor{blue!30}50.67 & \cellcolor{blue!30}50.63 & \cellcolor{blue!30}50.21 \\ \bottomrule
\end{tabular}%
}
\end{table*}
\section{Results}
We report performance on the four image classification datasets considered with the DeiT-S backbone results in Table~\ref{tab:perfSmall}, results for the DeiT-T and DeiT-B backbones in the \supp, and the average rank of the methods in Figure~\ref{fig:ranked}. Across all backbone capacities we note two trends: 1) pruning-based methods with learned reduction patterns are consistently among the top-3 methods across all datasets and 2) soft-clustering methods are consistently among the bottom three methods across all datasets. 

We also find that with the DeiT-T and DeiT-S backbones the hard-merging methods ToMe and DPC-KNN regularly outperform all other methods, especially on the ImageNet dataset and when only 25-50\% of the tokens are kept. However, with the DeiT-B backbone, we observe that the pruning-based methods with learned reduction patterns outperform even the hard-merging methods. Looking closer into the pruning-based methods we observe that the fixed-pattern $\ell_p$ methods are competitive when 90\% of tokens are kept, but at lower keep rates the performance drops significantly. For the learned approaches, we find that the DynamicViT method is the most unstable of the tested methods, often being in the bottom three methods when the keep rate is lower than 90\%. Similarly, we find that the performance of the ATS method is very dataset dependent, with great performance on the challenging NUS-WIDE dataset, but average performance on all other datasets. It should be noted that the ATS method manages to do so while on average using 50-90 and 10-30 fewer tokens than the other methods when the keep rate is set to 90\% and 70\%, respectively. This is discussed in the \supp. 

Comparatively, with a keep rate of 50-90\% the Top-K method is the best performing method 36\% of the time and in the top-3 methods 83\% of the time. This contradicts previous results \cite{ATS_2022}, and indicates that the Top-K method is a very strong baseline. However, at a keep rate of 25\% we find that the fused tokens in the EViT method can lead to an improvement of up to 2 percentage points over the Top-K method. This is also evident in Figure~\ref{fig:ranked}, where on average the EViT and Top-K methods are the two best ranked methods. Lastly, we note that when 90\% of tokens are kept, the Top-K, EViT, DynamicViT, ATS, and ToMe methods outperform the DeiT baselines by up to 0.5 percentage points. 
\section{In-Depth Analysis of Reduction Patterns}
In order to gain deeper insights into the token reduction process, we pose a set of research questions dedicated to uncovering the underlying core mechanisms of the investigated methods. We calculate the defined metrics per dataset and aggregate across all datasets, unless otherwise noted. 

Per-dataset results and examples of token reduction patterns can be found in the \supp. 

\begin{figure*}[!htp]
    \centering
    \includegraphics[width=\linewidth]{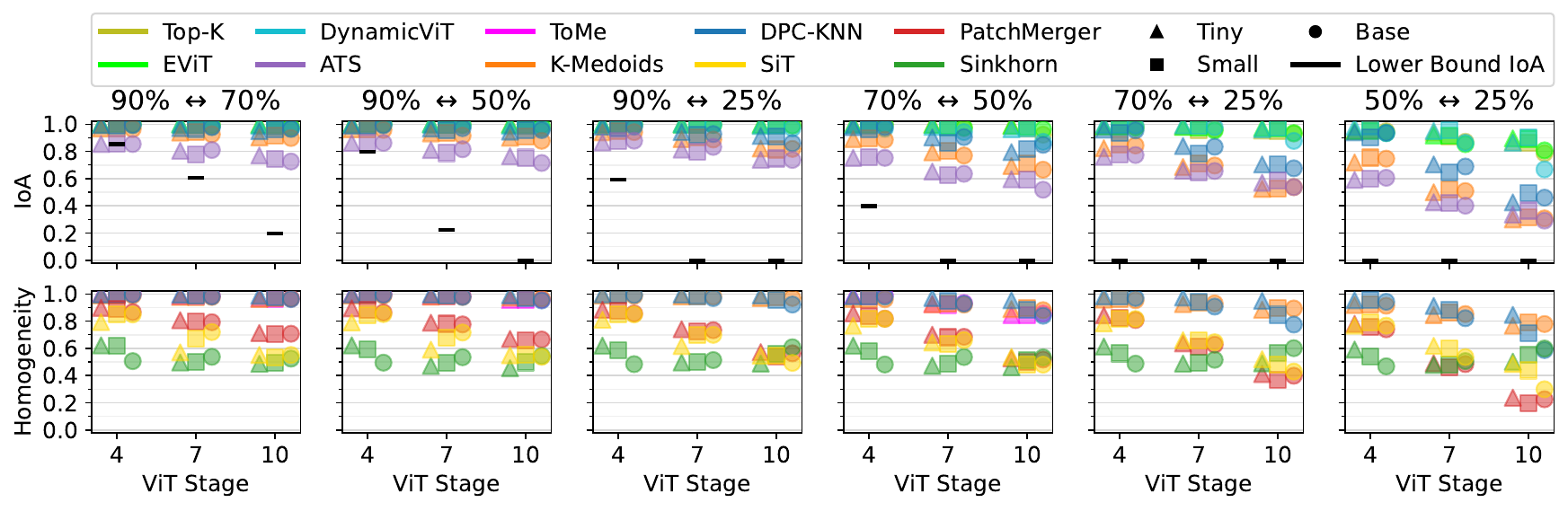}
    \caption{\textbf{Effect of varying $r$.} We quantify how similar the reduction patterns are when just the keep rate $r$ is varied. This is quantified with the Intersection over Area (IoA) and Homogeneity for pruning- and merging-methods, respectively. For the IoA metric we can derive the lower bound IoA given different keep rate values; see \supp. Note the overlap in pruning methods: Top-K, EViT, and DynamicVit, and merging methods: ToMe, DPC-KNN, and K-Medoids.}
    \label{fig:hypoSubset}
\end{figure*}

\subsection{Are Reduction Patterns Consistent when Varying the Keep Rate $r$?}\label{sec:hypoSubset}
A common assumption is that the token reduction methods will select tokens from the most representative regions of the image \cite{DyViT_2021}. Assuming this to be true, one would expect that the reduction patterns are consistent (\ie the same set of tokens are merged or pruned) when: 1) reducing the keep rate $r$ and 2) when varying the backbone capacity (see Section~\ref{sec:hypoCapacity}).

In order to evaluate whether the reduction patterns are consistent under varying keep rates, we consider reduction patterns $M_1$ and $M_2$ from models trained with keep rates $r_1$ and $r_2$, respectively, where $r_1\neq r_2$. We only compare within the same reduction method and backbone capacity. 

For pruning-based methods we evaluate using the Intersection over Area (IoA) between the reduction patterns, \ie how large a ratio of the tokens in $M_2$ are present in $M_1$, assuming $r_1>r_2$. For merging-based methods we evaluate using the Homogeneity of the constructed clusters \cite{Homogeneity}. Homogeneity is a measure of how consistent the class assignment is within each cluster, \ie whether the elements of each cluster in $M_1$ are assigned to the same clusters in $M_2$, assuming $r_1>r_2$. Further details on IoA and Homogeneity can be found in the \supp.

For the hard-clustering methods DPC-KNN and K-Medoids we evaluate using IoA in addition to Homogeneity, by treating the cluster centers as kept tokens. For evaluation of soft-merging methods, we define $M$ by assigning each token to the cluster with the highest assignment score.  

We plot our findings in Figure~\ref{fig:hypoSubset}. First we find that when lowering the keep rates, the IoA of Top-K, EViT, and DynamicViT (\ie the fixed keep rate pruning methods) are consistently high. However, for the hard-clustering methods and the dynamic keep rate ATS we observe that the IoA quickly drops across all reduction stages, towards the lower bound IoA values, indicating the extracted reduction patterns differ significantly. Secondly, we find that the Homogeneity of the hard-merging methods is consistently high, while it is significantly lower for the soft-merging methods.

From this we can conclude that pruning-based methods, with the exception of ATS, produce consistent reduction patterns when varying $r$. Similarly, we find that the hard-merging methods select consistent clusters, but with inconsistent cluster centers, while soft-merging methods produce inconsistent clusters when varying $r$.

\begin{figure*}[!htp]
    \centering
    \includegraphics[width=\linewidth]{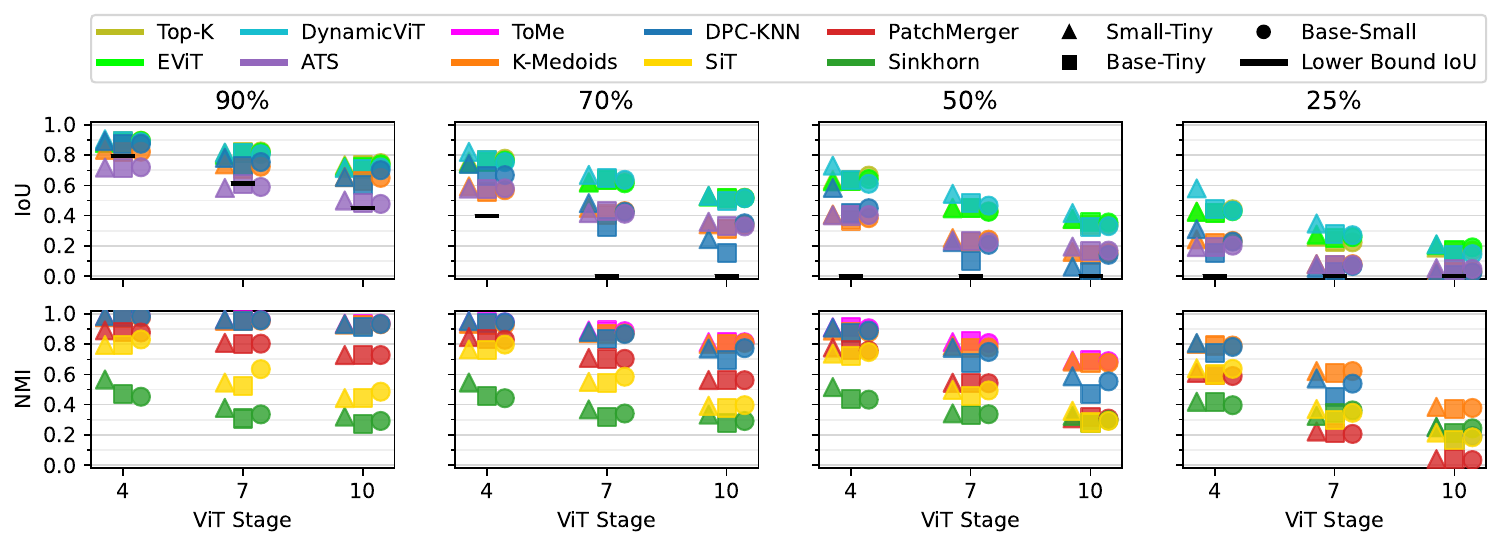}
        \caption{\textbf{Effect of varying the backbone capacity.} We quantify similarity of the reduction patterns when the backbone is varied. This is measured with the Intersection over Union (IoU) and Normalized Mutual Information (NMI) for pruning and merging methods, respectively. For the IoU metric we can derive the lower bound IoU given different keep rate values; see \supp. Note the IoU for the ATS method can be lower than the lower bound IoU, as it is a dynamic keep rate method.}
    \label{fig:hypoCapacity}
\end{figure*}

\subsection{Are Reduction Patterns Consistent when Varying Model Capacity?} \label{sec:hypoCapacity}
In order to evaluate whether the reduction patterns are affected by the backbone model capacity, we consider reduction patterns $M_1$ and $M_2$ from the same token reduction method trained with $r_1=r_2$ but varying backbone capacity. 

For pruning-based methods we evaluate using Intersection over Union (IoU) to gauge the similarity of $M_1$ and $M_2$. For merging-based methods we evaluate using the Normalized Mutual Information (NMI) \cite{NMI}. Further details on IoU and NMI can be found in the \supp.

As seen in Figure~\ref{fig:hypoCapacity}, we find that for pruning-based methods the similarity of the reduction patterns is very low for all keep rates. Similar to the observations made in Section~\ref{sec:hypoSubset} we observe that the IoU of the DPC-KNN, K-Medoids, and ATS methods is especially low. This indicates that the reduction patterns for pruning-based methods are inconsistent as the backbone model capacity is varied. However, for the hard-merging methods we observed that the clusters are consistent across backbone capacities at all reduction stages when $r>25\%$. The same can be observed for the soft-merging based PatchMerger method when $r>50\%$. 

From this we can conclude that the reduction patterns of pruning-based methods are inconsistent when varying the backbone capacity. For merging-based methods we find that the ToMe, DPC-KNN, and K-Medoids methods are consistent as long as $r>25\%$, while PatchMerger is consistent for $r>50\%$. We can again conclude that the hard-merging methods select consistent clusters, but with varying cluster centers, while soft-merging methods produce inconsistent clusters, as was observed in Section~\ref{sec:hypoSubset}.

\begin{figure*}[!htp]
    \centering
    \includegraphics[width=\linewidth]{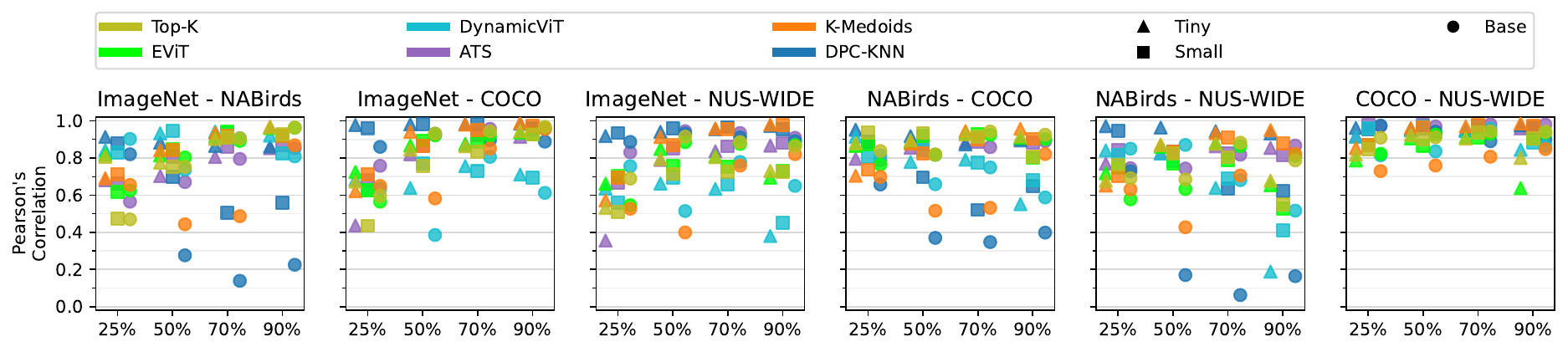}
    \caption{\textbf{Cross-Dataset reduction pattern similarity.} We quantify how similar the reduction patterns are across datasets for pruning-based methods, by measuring the correlation between the dataset averaged reduction patterns.}
    \label{fig:hypoHeatmap}
\end{figure*}

\subsection{Do Reduction Patterns Differ Across Datasets?}
Little is known about the behaviour of the reduction patterns across different image datasets. One open question is whether there are strong commonalities in the reduction patterns from different datasets, or whether the reduction patterns differ across datasets. In order to do such a comparison, we have to consider the global trends, as per-example comparisons cannot be made. We denote the dataset averaged reduction pattern as $\bar{M}$, which is obtained by computing how many ViT stages each token is passed through, averaged over the validation data splits.

We evaluate the similarity of the averaged reduction patterns across datasets, $\bar{M_1}$ and $\bar{M_2}$, by considering reduction patterns from the same token reduction method trained with keep rate $r$ and backbone capacity, but obtained from different datasets. In order to quantify the similarity we draw inspiration from the saliency detection field, specifically the analysis of different metrics by Bylinskii \etal \cite{SaliencyMetrics}. We use the common Pearson's correlation coefficient, and report results with other common saliency metrics in the \supp.

As seen in Figure~\ref{fig:hypoHeatmap} and following the rule of thumb guidelines by Hinkle
\etal \cite{hinkle2003applied}, we find a moderate-to-high correlation of the averaged reduction patterns for nearly all methods across all datasets and keep rates. The exceptions are the DPC-KNN, K-Medoids, and DynamicViT methods, which are found to have spurious lower (but still positive) correlation scores for several dataset pairs, indicating the averaged reduction patterns are less consistent. The lowest correlation scores are obtained by the DPC-KNN method, though this may be attributed to the inconsistent cluster centers as described in Section~\ref{sec:hypoSubset}-\ref{sec:hypoCapacity}. However, it is not intuitive that the reduction patterns are highly correlated across datasets, as one would expect that due to the significant differences across the investigated datasets imposed by the difference in type of classification task and its granularity. Nonetheless, the results indicate that on average the different token positions are used equally often across datasets. This might be due to biases in the image capturing process \eg the sky is always in the top half of the image and the object of interest is in the lower half and center of the image (as seen in Figure~\ref{fig:dataset}). We conclude that in general the reduction patterns do not differ significantly across datasets.

\subsection{Do Pruning-based Reduction Patterns Differ from Fixed Patterns?}
As discovered in prior work \cite{DyViT_2021, AViT_2022}, when averaging reduction patterns across a dataset the tokens near the image center are kept for longer, resembling a radial selection function. Therefore, it is reasonable to question how similar the per-example reduction patterns are to the fixed patterns from the $\ell_p$ methods. If they are similar, one could in principle do away with learning the adaptive reduction patterns.

We find that all learned pruning-based reduction patterns have a very low IoU with the fixed $\ell_p$ patterns at all reduction stages, shown in detail in the \supp.
As the $\ell_p$ patterns gradually focus on the tokens close to the image center, this indicates that the learned reduction patterns are not focused on the center. 
Instead the learned reduction patterns use information from across the entire image at all stages. We can therefore conclude that the learned pruning-based reduction patterns differ from fixed radial patterns. 

\begin{figure*}[!htp]
    \centering
    \includegraphics[width=\linewidth]{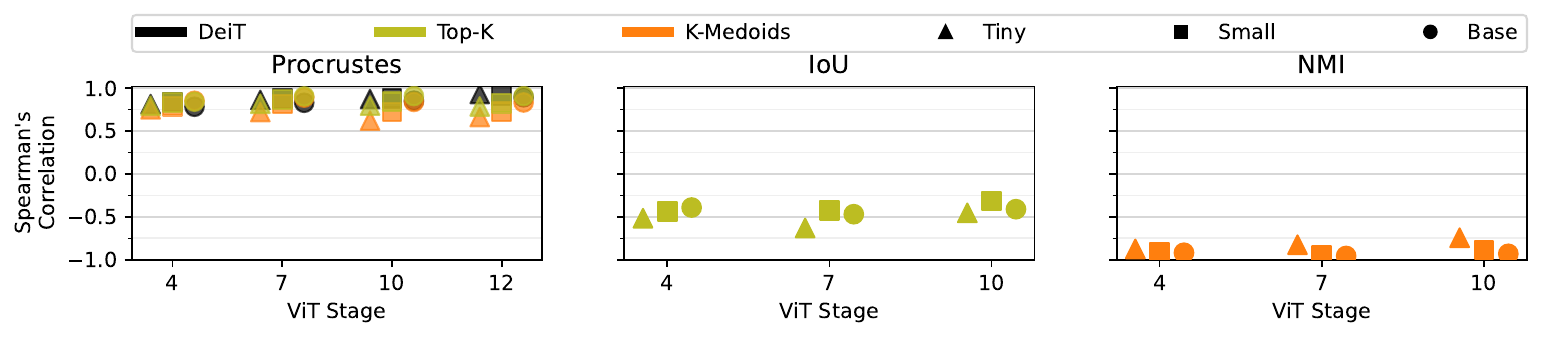}
    \caption{\textbf{Reduction patterns as performance proxies.} We determine whether reduction pattern similarity and feature alignment are good proxies for difference in model performance, by measuring the Spearman's ranked correlation between difference in performance and the orthogonal Procrustes distance, IoU, and NMI. Please note that Procrustes is a distance measure, whereas IoU and NMI are similarity measures.}
    \label{fig:hypoProxy}
\end{figure*}

\subsection{Are Reduction Patterns Good Proxies for Model Performance?}
A key practical question is whether a pair of reduction patterns can be used to predict the difference in model performance. This is investigated by determining the correlation between $f(A)-f(B)$ and $d(M_A,M_B)$, where $f$ is a performance measure (\ie accuracy or mAP), $d$ is a similarity measure, $A$ and $B$ are two models, and $M_A$ and $M_B$ are their reduction patterns. We use the Spearman's ranked correlation coefficient to measure the correlation. This approach was originally proposed by Ding \etal \cite{Procrustes_2021}. We set $d$ to be IoU and NMI for pruning- and merging-based methods, respectively, and constrain $A$ to be the Top-K and K-Medoids models as they are both strong baselines. Additionally, we measure the feature alignment between $A$ and $B$ by defining $d$ to be the orthogonal Procrustes distance, which Ding \etal \cite{Procrustes_2021} found to be a better metric of feature alignment than the typically used Centered Kernel Alignment \cite{CKA_2019} and Projection-Corrected Canonical Correlation Analysis \cite{CCA_2018}. For the feature alignment test we allow $A$ to be Top-K, K-Medoids, and DeiT, and calculate the alignment using the CLS token after the three reduction stages and the final ViT stage.

We find that for all model capacities the orthogonal Procrustes distance and NMI are highly correlated with the difference in model performance, while the IoU metric is moderately correlated, see Figure~\ref{fig:hypoProxy}. The fact that NMI is a better proxy than IoU indicates that the merging-based methods are more sensitive to the construction of the clusters, whereas pruning-based methods are less sensitive to the selection of specific tokens.

Lastly, the reason the Procrustes distance is a good proxy may be grounded in the fact all tested methods use a pre-trained DeiT backbone. Therefore, as long as the CLS token does not change during training of the token reduction method, feature alignment should be a good proxy. This has previously been the motivation for distillation losses \cite{DyViT_2021, SiT_2022}. However, all tested methods were trained without such losses, indicating that the well-performing models have inherently retained high CLS token alignment.
\section{Limitations}
We deliberately set certain limitations in order to keep the experiment complexity manageable. First of all, we only consider the image classification task where we extend the analysis from just ImageNet to three additional datasets. Therefore, extending our analysis to additional tasks such as video classification and action recognition is seen as out of scope and we leave this for the future. Secondly, we restricted our training scheme to only consider an ImageNet pre-trained backbone. This is common practice in the literature. Training from scratch on datasets such as ImageNet, OpenImages \cite{OpenImages}, or Visual Genome \cite{VisualGenome} would have been prohibitive, and we consider the fine-tuning setting to be more realistic when generalizing to datasets other than ImageNet. Thirdly, we do not investigate how interpretable or robust the different token reduction methods are, though this would be of great interest in the future. Lastly, while the main motivation for the token reduction methods has been to make ViTs more efficient, this work does not evaluate the efficiency of the tested methods. This is a deliberate choice as this work focuses on establishing a systematic comparison of methods with regards to classification performance, as well as gaining deeper insights into the core mechanisms of the tested methods. Furthermore, efficiency is not a simple thing to measure due to confounding factors such as hardware, implementation, and infrastructure as discussed by Dehghani \etal \cite{efficiency_2022}.
\section{Conclusion}
In this work we presented the first comprehensive and systematic analysis of 10 state-of-the-art token reduction methods. We find that the simple Top-K pruning approach is a very strong baseline across all tested image classification datasets, only outperformed by the EViT method, a straight-forward extension of Top-K. We conducted the first analysis of how the reduction patterns are affected by choice of dataset, number of tokens to be kept, and model capacity. We observe that varying the backbone has a large effect on the reduction patterns, whereas when the keep rate is varied the reduction patterns are very consistent. We also found a moderate-to-strong correlation of the average reduction patterns across datasets, and that the similarity of reduction patterns between methods is a moderate-to-strong proxy for model performance. We hope these findings will help inform future research in token reduction methods.

\noindent\textbf{Acknowledgements:} This work was supported by the Pioneer Centre for AI (DNRF grant number P1), partially supported by the Spanish project PID2022-136436NB-I00 and by ICREA under the ICREA Academia programme, and the Canada CIFAR AI Chairs.

\appendix
\section{Overview of Supplementary Materials}
In these supplementary materials we describe in further detail aspects of the training process, performance of the tested models, per-dataset results of the in-depth analysis, and visual examples of the reduction patterns. We refer to both the relevant sections in the main manuscript as well as sections in the supplementary materials. Specifically the following will be described:

\begin{itemize}
    \itemsep0em
    \item Hyperparameters for the four classification datasets (Section~3.3 / \ref{sec:hypers}). 
    \item Performance of token reduction methods using the DeiT-T and DeiT-B backbones (Section~4 / \ref{sec:resultsSupp}).  
    \item Per-dataset analysis of the dynamic keep rate in the ATS method (Section~4 / \ref{sec:atsKeepRate}). 
    \item Description of the pattern reduction similarity measures (Section~5.1--5.2 / \ref{sec:patternMetrics}).
    \item Description of the lower bound IoA and IoU computation (Section 5.1--5.2 / ~\ref{sec:pruningLB}).
    \item Per-dataset results for varying the keep rate $r$ and backbone capacity (Section~5.1--5.2 / \ref{sec:consistencySupp}). 
    \item Additional saliency metrics for the cross-dataset reduction pattern comparison (Section~5.3 / \ref{sec:saliencyMetrics}).
    \item Results of the $\ell_p$ reduction pattern comparison (Section~5.4 / \ref{sec:lpSupp}). 
    \item Results using CKA and PWCCA as proxies of model performance (Section~5.5 / \ref{sec:proxySupp}).
    \item Per-dataset scatter plots of the proxy measures and model performance (Section~5.5 / \ref{sec:proxyScatter}). 
    \item Per-dataset visualization of the averaged reduction patterns (Section~5.3 / \ref{sec:global}). 
    \item Per-dataset example visualization of the reduction patterns (Section~5 / \ref{sec:qualitativeViz}).
\end{itemize}

\begin{table}[]
    \centering
    \caption{\textbf{Hyperparameter grid search.} We conduct a gird search over a subset of the hyperparameters. For ImageNet the search is conducted over the token reduction methods (restricted to the warmup epochs, backbone scale, and how many the backbone weights are frozen), whereas for NABirds, COCO, and NUS-WIDE it is conducted for the DeiT-S baseline. We note that for the ImageNet dataset we restrict the backbone LR scale factor to only 1 or $0.01$, following Rao \etal \cite{DyViT_2021}.}
    \resizebox{\linewidth}{!}{%
    \begin{tabular}{ll}
        \toprule
        Hyperparameter & Grid Values \\
        \midrule
        Learning Rate (LR) & $[0.01, 0.001, 0.0001]$ \\
        LR Normalization Factor (LR-Norm) & $[512, 1024]$\\
        Warmup Epochs (W-E) & $[5, 20]$\\
        Backbone LR Scale  (B-LR) & $[1, 0.1, 0.01]$\\
        Backbone Freeze Epochs (B-FE) & $[0, 5]$\\
        \bottomrule
    \end{tabular}%
    }
    \label{tab:hyperMethod}
    \vspace{-3.5mm}
\end{table}
\section{Hyperparameters}\label{sec:hypers}
In this section we further elaborate on the training details in Section~3.3 and describe the hyperparameters used during training in detail. The hyperparameters can be split into two groups: 1) the static hyperparameters per dataset and 2) the hyperparameters which we conducted a search on per method. The static hyperparameters were selected based on what have been used in prior methods applied on each dataset \cite{DyViT_2021, ASL, EViT_2022, TransFG_2022}, as well as training guidelines from the DeiT paper \cite{DeiT_2021}. For all datasets we used the AdamW optimizer \cite{AdamW} with a momentum of 0.9 weight decay of 0.05, a Cosine learning rate schedule \cite{CosineLR} with a decay rate of 0.1, and stochastic depth of 0.1 \cite{StochasticDepth}. We train all methods on 2 V100 GPUs with mixed precision, repeated augmentations (x3) \cite{RepeatedAug, MultiGrain}, and gradient accumulation if the batch cannot fit onto the GPUs. For the K-Medoids and Sinkhorn methods we perform three iterations for the clustering, set the entropy regularization $\epsilon$ in the Sinkhorn method to 1, and set the number of neighbours $k=5$ for the DPC-KNN method. The remaining  hyperparameters are shown in Table~\ref{tab:hypersDataset}.

\begin{table*}[]
\centering
\caption{\textbf{Dataset-specific hyperparameters.} We fix a large set of the hyperparameters based on prior work. For ImageNet we are inspired by the DynamicViT and DeiT papers \cite{DyViT_2021, DeiT_2021}, NABirds is based on the hyperparameters used in the TransFG work \cite{TransFG_2022}, and COCO and NUS-WIDE are based on the hyperparameters from the ASL work \cite{ASL}.}
\label{tab:hypersDataset}
\begin{tabular}{@{}lllll@{}}
\toprule
Dataset & ImageNet & NABirds & COCO & NUS-WIDE \\ \midrule
Epochs & 30 & 50 & 40 & 40 \\
Batch size & 1024 & 1024 & 512 & 512 \\
Loss & Cross-Entropy & Cross-Entropy & ASL \cite{ASL} & ASL \cite{ASL} \\
Label Smoothing & 0.1 & 0 & 0 & 0 \\
ASL $\gamma_-$ & - & - & 4 & 4 \\
ASL Clip & - & - & 0.05 & 0.05 \\
Model EMA & 0.9999 & 0.9999 & 0.9997 & 0.9997 \\
Augmentations & \begin{tabular}[t]{@{}l@{}}Random Resize and Crop\\ Horizontal Flip (50\%)\\ RandAugment \cite{RandAugment}\\ Normalization\\ Random Erasing (25\%) \cite{RandomErasing}\\ Mixup/CutMix \cite{MixUp, CutMix}\end{tabular} & \begin{tabular}[t]{@{}l@{}}Random Resize and Crop\\ Horizontal Flip (50\%)\\ Normalization\end{tabular} & \begin{tabular}[t]{@{}l@{}}Resize\\ Cutout (50\%) \cite{CutOut}\\ RandAugment \cite{RandAugment}\\ Normalization\end{tabular} & \begin{tabular}[t]{@{}l@{}}Resize\\ Cutout (50\%) \cite{CutOut}\\ RandAugment \cite{RandAugment}\\ Normalization\end{tabular} \\ \bottomrule
\end{tabular}
\end{table*}

\begin{table*}[!htp]
\caption{\textbf{Selected token reduction method hyperparameters - ImageNet.} We present the selected hyperparameters when searching on ImageNet for each token reduction method.}
\label{tab:imagenetHyper}
\centering
\begin{tabular}{@{}lccclccclccclccc@{}}
\toprule
 $r$ (\%) & \multicolumn{3}{c}{25} &  & \multicolumn{3}{c}{50} &  & \multicolumn{3}{c}{70} &  & \multicolumn{3}{c}{90} \\\cmidrule(lr){2-4} \cmidrule(lr){6-8} \cmidrule(lr){10-12} \cmidrule(lr){14-16}
 & W-E & B-LR & B-FE & & W-E & B-LR & B-FE & & W-E & B-LR & B-FE & & W-E & B-LR & B-FE \\ \midrule
$\ell_1$ & 5 & 1 & 0 & & 5 & 0.01 & 5 & & 5 & 0.01 & 5 & & 5 & 0.01 & 5 \\
$\ell_2$ & 5 & 1 & 0 & & 5 & 0.01 & 5 & & 5 & 0.01 & 5 & & 5 & 0.01 & 5 \\
$\ell_\infty$ & 5 & 1 & 0 & & 5 & 0.01 & 5 & & 5 & 0.01 & 5 & & 5 & 0.01 & 5 \\ \midrule
Top-K & 5 & 1 & 0 & & 5 & 0.01 & 5 & & 5 & 0.01 & 5 & & 20 & 1 & 0 \\
EViT & 5 & 1 & 0 & & 5 & 0.01 & 5 & & 5 & 0.01 & 5 & & 20 & 1 & 0 \\
DynamicViT & 20 & 0.01 & 5 & & 5 & 0.01 & 5 & & 20 & 0.01 & 5 & & 20 & 0.01 & 5 \\
ATS & 5 & 1 & 0 & & 5 & 0.01 & 5 & & 20 & 0.01 & 5 & & 5 & 0.01 & 5 \\ \midrule
ToMe & - & - & - & & 5 & 0.01 & 5 & & 5 & 0.01 & 5 & & 5 & 0.01 & 5 \\
K-Medoids & 5 & 1 & 0 & & 5 & 0.01 & 5 & & 20 & 0.01 & 5 & & 20 & 1 & 0 \\
DPC-KNN & 5 & 0.01 & 5 & & 20 & 0.01 & 5 & & 5 & 0.01 & 5 & & 5 & 0.01 & 5 \\ \midrule
SiT & 5 & 0.01 & 5 & & 5 & 0.01 & 5 & & 5 & 0.01 & 5 & & 20 & 0.01 & 5 \\
PatchMerger & 5 & 0.01 & 5 & & 5 & 0.01 & 5 & & 5 & 0.01 & 5 & & 5 & 0.01 & 5 \\
Sinkhorn & 5 & 1 & 0 & & 5 & 1 & 0 & & 5 & 1 & 0 & & 5 & 1 & 0 \\ \bottomrule
\end{tabular}
\end{table*}

\begin{table}[!htp]
\caption{\textbf{Selected DeiT baseline hyperparameters.} We present the selected hyperparameters for the DeiT baselines on NABirds, COCO, and NUS-WIDE.}
\label{tab:deitHypers}
\centering
\begin{tabular}{@{}lcccccc@{}}
\toprule
Dataset & LR & LR-Norm & W-E & B-LR & B-FE \\ \midrule
NABirds & 0.001 & 1024 & 5 & 0.1 & 5 \\
COCO & 0.0001 & 512 & 5 & 1 & 0 \\
NUS-WIDE & 0.0001 & 512 & 5 & 1 & 0 \\ \bottomrule
\end{tabular}
\end{table}

\begin{table*}[!htp]
\caption{\textbf{Hyperparameter indicator matrix.} We illustrate below for each method and keep rate $r$ whether the hyperparameter settings from the ImageNet dataset ($\mathcal{I}$) or the dataset specific DeiT-S baseline ($\mathcal{D}$) are used.}
\label{tab:hyperAssign}
\centering
\begin{tabular}{@{}lcccclcccclcccc@{}}
\toprule
 & \multicolumn{4}{c}{NABirds} &  & \multicolumn{4}{c}{COCO} &  & \multicolumn{4}{c}{NUS-WIDE} \\\cmidrule(lr){2-5} \cmidrule(lr){7-10} \cmidrule(lr){12-15}
$r$  (\%) & 25 & 50 & 70 & 90 &  & 25 & 50 & 70 & 90 &  & 25 & 50 & 70 & 90 \\ \midrule
$\ell_1$ & $\mathcal{I}$  & $\mathcal{D}$ & $\mathcal{D}$ & $\mathcal{D}$ & & $\mathcal{D}$ & $\mathcal{D}$ & $\mathcal{D}$ & $\mathcal{D}$ & & $\mathcal{D}$ & $\mathcal{D}$ & $\mathcal{D}$ & $\mathcal{D}$ \\
$\ell_2$ & $\mathcal{I}$  & $\mathcal{D}$ & $\mathcal{D}$ & $\mathcal{D}$ & & $\mathcal{D}$ & $\mathcal{D}$ & $\mathcal{D}$ & $\mathcal{D}$ & & $\mathcal{D}$ & $\mathcal{D}$ & $\mathcal{D}$ & $\mathcal{D}$ \\
$\ell_\infty$ & $\mathcal{I}$  & $\mathcal{D}$ & $\mathcal{D}$ & $\mathcal{D}$ & & $\mathcal{D}$ & $\mathcal{D}$ & $\mathcal{D}$ & $\mathcal{D}$ & & $\mathcal{D}$ & $\mathcal{D}$ & $\mathcal{D}$ & $\mathcal{D}$ \\ \midrule
Top-K & $\mathcal{D}$  & $\mathcal{D}$ & $\mathcal{D}$ & $\mathcal{D}$ & & $\mathcal{D}$ & $\mathcal{D}$ & $\mathcal{D}$ & $\mathcal{D}$ & & $\mathcal{D}$ & $\mathcal{D}$ & $\mathcal{D}$ & $\mathcal{D}$ \\
EViT &$\mathcal{D}$  & $\mathcal{D}$ & $\mathcal{D}$ & $\mathcal{D}$ & & $\mathcal{D}$ & $\mathcal{D}$ & $\mathcal{D}$ & $\mathcal{D}$ & & $\mathcal{D}$ & $\mathcal{D}$ & $\mathcal{D}$ & $\mathcal{D}$ \\
DynamicViT & $\mathcal{I}$ & $\mathcal{D}$ & $\mathcal{D}$ & $\mathcal{D}$ & & $\mathcal{I}$ & $\mathcal{I}$  & $\mathcal{D}$ & $\mathcal{D}$ & & $\mathcal{I}$ & $\mathcal{D}$ & $\mathcal{D}$ & $\mathcal{D}$ \\
ATS & $\mathcal{I}$  & $\mathcal{D}$ & $\mathcal{D}$ & $\mathcal{D}$ & & $\mathcal{D}$ & $\mathcal{D}$ & $\mathcal{D}$ & $\mathcal{D}$ & & $\mathcal{D}$ & $\mathcal{D}$ & $\mathcal{D}$ & $\mathcal{D}$ \\ \midrule
ToMe & - & $\mathcal{D}$ & $\mathcal{D}$ & $\mathcal{D}$ & & - & $\mathcal{D}$ & $\mathcal{D}$ & $\mathcal{D}$ & & - & $\mathcal{D}$ & $\mathcal{D}$ & $\mathcal{D}$ \\
K-Medoids &$\mathcal{D}$  & $\mathcal{D}$ & $\mathcal{D}$ & $\mathcal{D}$ & & $\mathcal{D}$ & $\mathcal{D}$ & $\mathcal{D}$ & $\mathcal{D}$ & & $\mathcal{D}$ & $\mathcal{D}$ & $\mathcal{D}$ & $\mathcal{D}$ \\
DPC-KNN &$\mathcal{D}$  & $\mathcal{D}$ & $\mathcal{D}$ & $\mathcal{D}$ & & $\mathcal{D}$ & $\mathcal{D}$ & $\mathcal{D}$ & $\mathcal{D}$ & & $\mathcal{D}$ & $\mathcal{D}$ & $\mathcal{D}$ & $\mathcal{D}$ \\ \midrule
SiT &$\mathcal{D}$  & $\mathcal{D}$ & $\mathcal{D}$ & $\mathcal{D}$ & & $\mathcal{D}$ & $\mathcal{D}$ & $\mathcal{D}$ & $\mathcal{D}$& & $\mathcal{I}$ & $\mathcal{I}$ & $\mathcal{I}$ & $\mathcal{I}$  \\
PatchMerger &$\mathcal{D}$  & $\mathcal{D}$ & $\mathcal{D}$ & $\mathcal{D}$ & & $\mathcal{D}$ & $\mathcal{D}$ & $\mathcal{D}$ & $\mathcal{D}$ & & $\mathcal{D}$ & $\mathcal{D}$ & $\mathcal{D}$ & $\mathcal{D}$ \\
Sinkhorn & $\mathcal{I}$ & $\mathcal{I}$ & $\mathcal{I}$ & $\mathcal{I}$ & & $\mathcal{I}$ & $\mathcal{I}$ & $\mathcal{I}$ & $\mathcal{I}$ &  & $\mathcal{I}$ & $\mathcal{I}$ & $\mathcal{I}$ & $\mathcal{I}$  \\ \bottomrule
\end{tabular}
\end{table*}

For a subset of the hyperparameters we perform a grid search per token reduction method with the DeiT-S backbone on the ImageNet dataset, and for the DeiT-S baseline on the NABirds, COCO, and NUS-WIDE datasets. The grid searched hyperparameters are: the learning rate, the number of warmup epochs in the cosine scheduler, the number of epochs where the backbone weights should be fixed, the backbone weights learning rate scaling factor, and a normalization factor of the learning rate \cite{LRNormFactor}. The hyperparameter value ranges are shown in Table~\ref{tab:hyperMethod}.  On ImageNet we fix the learning rate to $0.001$ and the normalization factor to 1024 (\ie the batch size). On the NABirds, COCO, and NUS-WIDE datasets we determine the final per-method hyperparameters by comparing models trained with the method-specific hyperparameters obtained on ImageNet and the dataset-specific DeiT-S baseline hyperparameters.
The best hyperparameters per token reduction method and keep rate $r$ on the ImageNet dataset is shown in Table~\ref{tab:imagenetHyper}. For NABirds, COCO, and NUS-WIDE we show the best hyperparameters for the DeiT-S baseline in Table~\ref{tab:deitHypers}, and an indicator matrix in Table~\ref{tab:hyperAssign} indicating whether the dataset fine-tuned DeiT-S hyperparameters or the ImageNet hyperparameters are used per token reduction method and $r$.

\section{Token Reduction Performance using DeiT-T and DeiT-B Backbones}\label{sec:resultsSupp}
In this section we present the results with the DeiT-T and DeiT-B baselines as mentioned in Section~4; see Table~\ref{tab:perfAll}.

\begin{table*}[!htp]
\caption{\textbf{Performance of Token Reduction methods.} Measured across varying keep rates, $r$, and backbone capacities. Scores exceeding the DeiT baseline are noted in \textbf{bold}, measured as Top-1 accuracy for ImageNet \& NABirds and mean Average Precision for COCO and NUS-WIDE. The three best performing methods per keep rate are denoted in descending order with \colorbox{red!25}{red}, \colorbox{orange!25}{orange}, and \colorbox{yellow!25}{yellow},  respectively. Similarly, the three worst performing methods are denoted in descending order with \colorbox{blue!10}{light blue}, \colorbox{blue!20}{blue}, and \colorbox{blue!30}{dark blue}}
\label{tab:perfAll}
\begin{subtable}{\textwidth}
\centering
\caption{Performance comparison of token reduction methods trained with a DeiT-Base backbone.}
\label{tab:perfBase}
\resizebox{\textwidth}{!}{%
\begin{tabular}{@{}lcccclcccclcccclcccc@{}}
\toprule
 & \multicolumn{4}{c}{ImageNet} &  & \multicolumn{4}{c}{NABirds} &  & \multicolumn{4}{c}{COCO} &  & \multicolumn{4}{c}{NUS-WIDE} \\ \cmidrule(lr){2-5} \cmidrule(lr){7-10} \cmidrule(lr){12-15} \cmidrule(lr){17-20}  
DeiT-B & \multicolumn{4}{c}{81.85} &  & \multicolumn{4}{c}{83.32} &  & \multicolumn{4}{c}{80.93} &  & \multicolumn{4}{c}{64.37} \\ \midrule\midrule
$r$  (\%) & 25 & 50 & 70 & 90 &  & 25 & 50 & 70 & 90 &  & 25 & 50 & 70 & 90 &  & 25 & 50 & 70 & 90 \\ \midrule

$\ell_1$ & 71.23 & 74.96 & 78.94 & 81.04 &  & 59.79 & 71.57 & 78.92 & 82.42 &  & 58.28 & 69.27 & 76.23 & 79.65 &  & 53.01 & 60.10 & 63.25 & 64.14 \\
$\ell_2$ & 71.41 & 75.40 & 79.07 & 81.18 &  & 61.55 & 73.24 & 79.52 & 82.55 &  & 59.69 & 70.33 & 76.56 & 79.75 &  & 54.00 & 60.37 & 63.29 & 64.28 \\
$\ell_\infty$ & 71.67 & \cellcolor{blue!10}74.40 & 78.95 & 81.20 &  & 59.96 & 70.51 & 79.73 & 82.59 &  & 58.48 & 68.50 & 76.54 & 79.89 &  & 53.00 & 59.59 & 63.12 & 64.25 \\ \midrule
Top-K & \cellcolor{yellow!25}73.63 & \cellcolor{orange!25}78.97 & \cellcolor{yellow!25}80.91 & \cellcolor{red!25}\textbf{82.03} &  & \cellcolor{orange!25}74.71 & \cellcolor{orange!25}82.22 & \cellcolor{red!25}83.20 & \cellcolor{red!25}\textbf{83.40} &  & \cellcolor{yellow!25}67.63 & \cellcolor{orange!25}76.91 & \cellcolor{red!25}79.95 & \cellcolor{red!25}\textbf{80.97} &  & 58.51 & \cellcolor{yellow!25}62.78 & \cellcolor{yellow!25}63.92 & \cellcolor{yellow!25}\textbf{64.40} \\
EViT & \cellcolor{red!25}75.26 & \cellcolor{red!25}79.22 & \cellcolor{orange!25}80.99 & \cellcolor{orange!25}\textbf{82.00} &  & \cellcolor{red!25}74.73 & \cellcolor{yellow!25}82.00 & \cellcolor{orange!25}83.19 & \cellcolor{orange!25}\textbf{83.33} &  & \cellcolor{red!25}68.93 & \cellcolor{red!25}76.92 & \cellcolor{orange!25}79.87 & \cellcolor{orange!25}80.92 &  & \cellcolor{orange!25}59.00 & \cellcolor{orange!25}62.88 & 63.90 & \cellcolor{orange!25}\textbf{64.43} \\
DynamicViT & \cellcolor{blue!30}27.94 & 74.58 & 80.68 & \cellcolor{yellow!25}81.76 &  & \cellcolor{blue!10}49.23 & \cellcolor{red!25}82.30 & \cellcolor{yellow!25}83.16 & 83.23 &  & \cellcolor{blue!30}24.88 & \cellcolor{blue!10}62.79 & 76.54 & \cellcolor{yellow!25}80.64 &  & \cellcolor{blue!30}28.56 & \cellcolor{blue!20}55.51 & \cellcolor{blue!10}60.73 & 63.83 \\
ATS & \cellcolor{orange!25}73.89 & \cellcolor{yellow!25}78.94 & 80.78 & 81.57 &  & \cellcolor{yellow!25}71.00 & 80.10 & 82.58 & \cellcolor{yellow!25}83.26 &  & \cellcolor{orange!25}68.17 & \cellcolor{yellow!25}76.38 & \cellcolor{yellow!25}79.35 & 80.50 &  & \cellcolor{red!25}59.49 & \cellcolor{red!25}63.17 & \cellcolor{red!25}64.21 & \cellcolor{red!25}\textbf{64.48} \\ \midrule
ToMe & - & 78.89 & \cellcolor{red!25}81.05 & \cellcolor{orange!25}\textbf{82.00} &  & - & 73.67 & 81.59 & 82.98 &  & - & 74.11 & 78.82 & 80.48 &  & - & 62.38 & \cellcolor{orange!25}64.06 & 64.35 \\
K-Medoids & 69.12 & 76.86 & 79.98 & \cellcolor{yellow!25}81.76 &  & 57.54 & 75.29 & 80.62 & 82.57 &  & 61.79 & 73.60 & 77.58 & 80.32 &  & 56.67 & 62.18 & 63.53 & 64.35 \\
DPC-KNN & 69.40 & 75.87 & 79.06 & 81.05 &  & 58.16 & \cellcolor{blue!10}67.36 & 72.83 & 78.29 &  & 65.99 & 73.32 & 77.03 & 79.76 &  & \cellcolor{yellow!25}58.58 & 61.39 & 62.96 & 63.87 \\ \midrule
SiT & 68.39 & 75.53 & \cellcolor{blue!10}76.63 & \cellcolor{blue!10}77.26 &  & 65.09 & 70.75 & \cellcolor{blue!10}70.36 & \cellcolor{blue!10}68.96 &  & 54.86 & \cellcolor{blue!20}53.27 & \cellcolor{blue!20}53.16 & \cellcolor{blue!20}52.73 &  & 56.12 & 59.76 & \cellcolor{blue!20}60.64 & \cellcolor{blue!20}61.08 \\
PatchMerger & \cellcolor{blue!20}58.78 & \cellcolor{blue!20}70.63 & \cellcolor{blue!20}74.52 & \cellcolor{blue!20}76.76 &  & \cellcolor{blue!30}40.38 & \cellcolor{blue!20}57.21 & \cellcolor{blue!20}62.20 & \cellcolor{blue!20}67.06 &  & \cellcolor{blue!10}54.25 & 66.22 & \cellcolor{blue!10}70.97 & \cellcolor{blue!10}73.72 &  & \cellcolor{blue!10}51.80 & \cellcolor{blue!10}58.83 & 60.79 & \cellcolor{blue!10}62.09 \\
Sinkhorn & \cellcolor{blue!10}63.37 & \cellcolor{blue!30}63.33 &\cellcolor{blue!30}63.36 & \cellcolor{blue!30}63.50 &  & \cellcolor{blue!20}42.89 &\cellcolor{blue!30}42.33 & \cellcolor{blue!30}41.72 & \cellcolor{blue!30}42.86 &  & \cellcolor{blue!20}52.57 & \cellcolor{blue!30}52.33 & \cellcolor{blue!30}52.21 & \cellcolor{blue!30}52.12 &  & \cellcolor{blue!20}47.55 & \cellcolor{blue!30}47.41 & \cellcolor{blue!30}47.26 & \cellcolor{blue!30}47.48 \\ \bottomrule
\end{tabular}%
}
\end{subtable}

\bigskip
\begin{subtable}{\textwidth}
\centering
\label{tab:perfTiny}
\caption{Performance comparison of token reduction methods trained with a DeiT-Tiny backbone.}
\resizebox{\textwidth}{!}{%
\begin{tabular}{@{}lcccclcccclcccclcccc@{}}
\toprule
 & \multicolumn{4}{c}{ImageNet} &  & \multicolumn{4}{c}{NABirds} &  & \multicolumn{4}{c}{COCO} &  & \multicolumn{4}{c}{NUS-WIDE} \\ \cmidrule(lr){2-5} \cmidrule(lr){7-10} \cmidrule(lr){12-15} \cmidrule(lr){17-20} 
DeiT-T & \multicolumn{4}{c}{72.20} & & \multicolumn{4}{c}{74.16} & & \multicolumn{4}{c}{71.09} & & \multicolumn{4}{c}{59.27} \\ \midrule\midrule
$r$  (\%) & 25 & 50 & 70 & 90 &  & 25 & 50 & 70 & 90 &  & 25 & 50 & 70 & 90 &  & 25 & 50 & 70 & 90 \\ \midrule
$\ell_1$ & 58.58 & \cellcolor{blue!10}62.27 & 67.91 & 71.06 &  & 51.82 & 59.25 & 68.36 & 73.47 &  & \cellcolor{blue!10}49.09 & \cellcolor{blue!10}58.27 & 67.03 & 70.24 &  & \cellcolor{blue!10}44.81 & 52.64 & 57.30 & 58.73 \\
$\ell_2$ & 58.85 & 62.91 & 67.91 & 71.13 &  & 53.10 & 60.87 & 69.20 & 73.42 &  & 50.46 & 60.00 & 67.33 & 69.98 &  & 45.73 & 53.45 & 57.34 & 58.54 \\
$\ell_\infty$ & 59.08 & \cellcolor{blue!20}61.92 & \cellcolor{blue!10}67.79 & 71.38 &  & 52.60 & 57.70 & 69.25 & 73.34 &  & 50.08 & 57.30 & 67.22 & 69.89 &  & 45.32 & 51.91 & 57.41 & 58.30 \\ \midrule
Top-K & 62.19 & 68.55 & 70.96 & 71.85 &  & 62.14 & \cellcolor{orange!25}73.19 & \cellcolor{red!25}\textbf{74.57} & \cellcolor{red!25}\textbf{74.64} &  & 60.31 & \cellcolor{yellow!25}67.47 & \cellcolor{orange!25}70.20 & \cellcolor{red!25}\textbf{71.65} &  & 52.20 & 57.02 & \cellcolor{orange!25}58.60 & \cellcolor{red!25}\textbf{59.50} \\
EViT & \cellcolor{orange!25}64.11 & \cellcolor{yellow!25}68.69 & \cellcolor{yellow!25}71.06 & 71.83 &  & \cellcolor{yellow!25}64.13 & \cellcolor{red!25}73.24 & \cellcolor{orange!25}\textbf{74.49} & \cellcolor{orange!25}\textbf{74.53} &  & \cellcolor{orange!25}61.44 & \cellcolor{orange!25}67.62 & \cellcolor{red!25}70.25 & \cellcolor{orange!25}\textbf{71.63} &  & \cellcolor{orange!25}53.09 & \cellcolor{yellow!25}57.26 & \cellcolor{red!25}58.64 & \cellcolor{orange!25}\textbf{59.49} \\
DynamicViT & \cellcolor{blue!30}36.93 & 67.40 & 70.94 & \cellcolor{orange!25}72.14 &  & 57.38 & \cellcolor{yellow!25}72.54 & \cellcolor{yellow!25}73.97 & \textbf{74.30} &  & \cellcolor{blue!30}24.67 & 61.70 & 68.83 & \cellcolor{yellow!25}\textbf{71.30} &  & \cellcolor{blue!30}28.09 & \cellcolor{blue!10}49.36 & 56.79 & 58.95 \\
ATS & 62.63 & 68.61 & 70.77 & 71.71 &  & \cellcolor{red!25}64.53 & 71.07 & 73.71 & \textbf{74.43} &  & \cellcolor{yellow!25}60.97 &67.37 & \cellcolor{yellow!25}69.88 & \textbf{71.10} &  & \cellcolor{yellow!25}52.85 & \cellcolor{orange!25}57.30 & \cellcolor{yellow!25}58.55 & \cellcolor{yellow!25}59.20 \\ \midrule
ToMe & - & \cellcolor{red!25}69.72 & \cellcolor{red!25}71.74 & \cellcolor{red!25}72.16 &  & - & 66.61 & 73.65 & \cellcolor{yellow!25}\textbf{74.50} &  & - & 65.66 & 69.70 & \textbf{71.16} &  & - & 55.32 & 57.78 & 58.98 \\
K-Medoids & \cellcolor{blue!10}57.50 & 65.82 & 69.90 & 71.50 &  & 44.62 & 66.52 & 72.09 & 74.04 &  & 54.08 & 64.83 & 69.09 & 71.05 &  & 49.13 & 55.92 & 58.38 & 59.07 \\
DPC-KNN & \cellcolor{red!25}64.56 & \cellcolor{orange!25}69.68 & \cellcolor{orange!25}71.10 & \cellcolor{yellow!25}71.88 &  & \cellcolor{orange!25}64.23 & 71.05 & 73.02 & 74.05 &  & \cellcolor{red!25}63.32 & \cellcolor{red!25}68.03 & 69.55 & 70.84 &  & \cellcolor{red!25}55.37 & \cellcolor{red!25}57.33 & 58.08 & 58.88 \\ \midrule
SiT & \cellcolor{yellow!25}63.43 & 67.98 & 68.99 & \cellcolor{blue!10}68.90 &  & \cellcolor{blue!30}36.35 & \cellcolor{blue!20}36.65 & \cellcolor{blue!30}34.00 & \cellcolor{blue!30}35.07 &  & \cellcolor{blue!20}48.01 & \cellcolor{blue!30}47.50 & \cellcolor{blue!30}46.98 & \cellcolor{blue!30}46.48 &  & \cellcolor{blue!20}36.67 & \cellcolor{blue!30}38.15 & \cellcolor{blue!30}36.98 & \cellcolor{blue!30}37.70 \\
PatchMerger & 60.38 & 64.80 & \cellcolor{blue!20}66.81 & \cellcolor{blue!20}68.09 &  & \cellcolor{blue!10}38.83 & \cellcolor{blue!10}54.20 & \cellcolor{blue!10}59.94 & \cellcolor{blue!10}62.60 &  & 52.49 & 59.69 & \cellcolor{blue!10}62.63 & \cellcolor{blue!10}64.30 &  & 47.56 & 52.69 & \cellcolor{blue!10}54.33 & \cellcolor{blue!10}55.37 \\
Sinkhorn & \cellcolor{blue!20}53.61 & \cellcolor{blue!30}53.49 & \cellcolor{blue!30}53.19 & \cellcolor{blue!30}53.51 &  & \cellcolor{blue!20}36.94 & \cellcolor{blue!30}35.98 & \cellcolor{blue!20}37.29 & \cellcolor{blue!20}36.19 &  & 50.47 & \cellcolor{blue!20}49.52 & \cellcolor{blue!20}49.12 & 4\cellcolor{blue!20}9.01 &  & 45.77 & \cellcolor{blue!20}44.81 & \cellcolor{blue!20}44.52 & \cellcolor{blue!20}44.20 \\ \bottomrule
\end{tabular}%
}
\end{subtable}
\end{table*}

\section{Analysis of ATS Keep Rates}\label{sec:atsKeepRate}
\begin{figure*}[!htp]
    \centering\includegraphics[width=\linewidth]{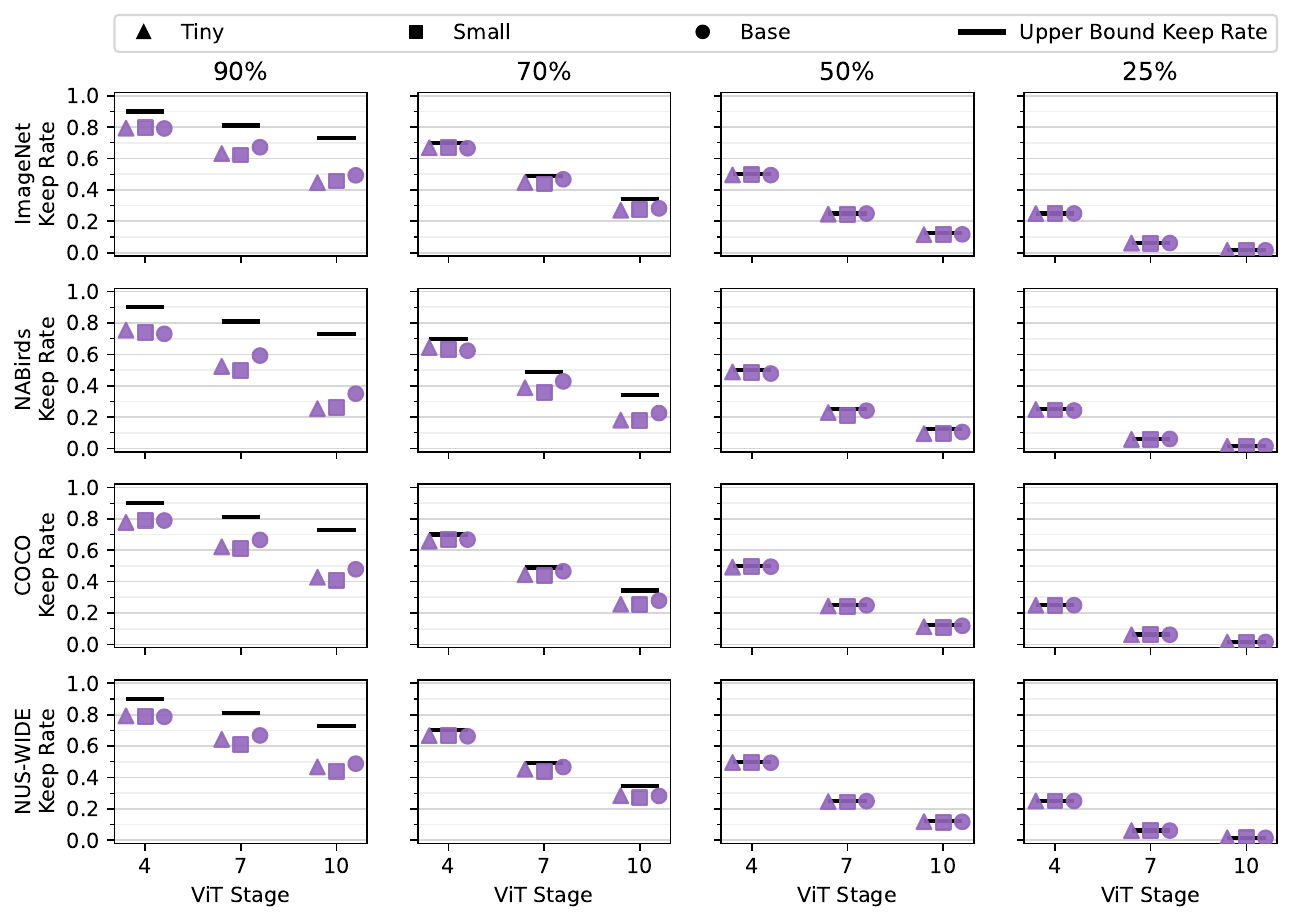}
    \caption{\textbf{Actual ATS keep rates (Section~\ref{sec:atsKeepRate}).} The ATS method is a dynamic keep rate pruning method, meaning the amount of tokens kept at each reduction stage can be variable. This means the keep rate $r$ is interpreted as an upper bound. We find that with an  $r$ value of 90\% and 70\% the actual keep rates are dramatically lower at the later reduction stages.}
    \label{fig:atsKeepRate}
\end{figure*}

As discussed in Section~3.1.3, the ATS \cite{ATS_2022} method is a dynamic keep rate pruning method, and therefore the meaning of the keep rate $r$ makes a subtle but important change. Instead of being the ratio of kept tokens, it instead represents the upper bound of the ratio of tokens to be kept, which the ATS method cannot exceed. In order to better understand the ATS method we plot the per-dataset average keep rate at each ViT stage for the different values of $r$; see Figure~\ref{fig:atsKeepRate}. We observe that when 90\% and 70\% of the tokens may be kept the actual keep rate is much lower, especially during the later stages of the ViT.

\section{Reduction Pattern Similarity Metrics}\label{sec:patternMetrics}
In this section we describe in more detail the metrics used to compare reduction patterns in Sections~5.1--5.2. 
For the Intersection over Area (IoA) and Intersection over Union (IoU) metrics used to compare pruning-based methods, each method produces a reduction pattern $M$ with keep rate $r$, where $M$ consists of the kept tokens after applying the reduction method. Using set notation the IoA and IoU can then be defined as in Equations~\ref{eq:IoA}-\ref{eq:IoU}.
\begin{equation}\label{eq:IoA}
    \text{IoA} = \frac{M_1\cap M_2}{M_2} \ \text{s.t.}\ r_1 \geq r_2
\end{equation}
\begin{equation}\label{eq:IoU}
    \text{IoU} = \frac{M_1\cap M_s}{M_1\cup M_2}
\end{equation}

For clustering-based methods, we utilize two information theoretic metrics: Homogeneity \cite{Homogeneity} and Normalized Mutual Information (NMI) \cite{NMI}. Homogeneity measures the class distribution within the constructed clusters, where the optimal value is obtained if all data points from the same class are assigned to the same cluster. This can be expressed using entropy as in Equations~\ref{eq:Homogeneity}-\ref{eq:clusterEntropy}.
\begin{equation}\label{eq:Homogeneity}
    h = \left\{ 
  \begin{array}{ c l }
    1 & \text{if } H(C,K) = 0 \\
    1-\frac{H(C|K)}{H(C)}                 & \quad \textrm{else}
  \end{array}
\right.
\end{equation}
\begin{equation}
    H(C|K) = -\sum_{k=1}^{|K|}\sum_{c=1}^{|C|}\frac{n_{c,k}}{N}\log\frac{n_{c,k}}{n_k}
\end{equation}
\begin{equation}\label{eq:clusterEntropy}
    H(C) = -\sum_{c=1}^{|C|}\frac{n_c}{N}\log\frac{n_c}{N}
\end{equation}
where $K$ is the set of generated clusters, $C$ is the set of ground truth classes, $n_{c,k}$ is the number of data points from class $c$ in cluster $k$, $n_{c}$ is the number of data points in class $c$, $n_{k}$ is the number of data points in cluster $k$, and $N$ is the total number of data points. 

In our analysis we define $C$ to be the constructed clusters in $M_1$ and $K$ to be the clusters in $M_2$, given $r_1 \geq r_2$. Thereby $|K|\leq|C|$, and the Homogeneity measures how well each cluster in $M_1$ maps to the reduced amount of clusters in $M_2$.\\

Similarly the NMI can be expressed using the Mutual Information (MI) between the constructed clusters in $M_1$ and $M_2$ (denoted as $C$ and $K$ to keep consistency with the Homogeneity notation), normalized by the averaged entropy of C and K, see Equations~\ref{eq:NMI}-\ref{eq:MI}.\\
\begin{equation}\label{eq:NMI}
    \text{NMI}(C,K) = \frac{I(C,K)}{(H(C)+H(K))/2}
\end{equation}
\begin{equation}\label{eq:MI}
    I(C,K) = \frac{n_{c,k}}{N}\log\frac{Nn_{c,k}}{n_c n_k}
\end{equation}\\

Furthermore, using the Homogeneity $h$, and the symmetric metric ``Completeness'', $c$, a combined metric called the ``V-Measure'' can be defined as the harmonic mean of $h$ and $c$ \cite{Homogeneity}. It has been shown by Becker \cite{Becker} that the V-Measure is equivalent to the Normalized Mutual Information, when the arithmetic mean is used for normalizing the MI.

\section{Lower Bound of IoA and IoU}\label{sec:pruningLB}
When comparing pruning-based token reduction methods in Section~5.1--5.2 the keep rates, $r_1$ and $r_2$, may be selected such that a subset of tokens will be selected by both models. This can skew the interpretation of the IoA and IoU metrics, as the metrics may have high values but in fact only due to the inherently overlapping subset. In order to account for this we determine the minimum IoA and IoU for the reduction stage given $r_1$, $r_2$, and number of spatial tokens in the input image, $P$, using the Algorithms~\ref{alg:ioaLB}-\ref{alg:iouLB}. These lower bounds are only true for pruning-based methods with a static keep rate and may therefore be broken by the ATS method.

It is not necessary to derive similar lower bounds for the clustering-based Homogeneity and NMI metrics, as both metrics can reach a value of 0. Homogeneity reaches 0 when the clustering provides no new information, \ie when the class distribution in each cluster is equal to the overall class distribution \cite{Homogeneity}. Similarly, it can be inferred the same is true for the NMI, since NMI can be expressed in terms of Homogeneity and Completeness. 

\begin{algorithm}
\caption{Lower bound of IoA }\label{alg:ioaLB}
\begin{algorithmic}
\Require $P$, $r_1$, $r_2$, s.t. $r_1 \geq r_2$
\Ensure $LB$
\State $LB \gets \varnothing$
\For{$s\in\{1,2,3\}$}
\State $P_{s,r_1} \gets \lfloor Pr_1^s\rfloor$
\State $P_{s,r_2} \gets \lfloor Pr_2^s\rfloor$
\State $P_{s,r_1,r_2} \gets P_{s,r_1}+P_{s,r_2}$
\If{$P_{s,r_1,r_2} \geq P$}
    \State $LB_s \gets \frac{P_{s,r_1,r_2}-P}{P_{s,r_2}}$
\Else
    \State $LB_s \gets 0$
\EndIf
\State $LB \gets LB\cup LB_s$
\EndFor
\end{algorithmic}
\end{algorithm}

\begin{algorithm}
\caption{Lower bound of IoU}\label{alg:iouLB}
\begin{algorithmic}
\Require $P$, $r_1$, $r_2$
\Ensure $LB$
\State $LB \gets \varnothing$
\For{$s\in\{1,2,3\}$}
\State $P_{s,r_1} \gets \lfloor Pr_1^s\rfloor$
\State $P_{s,r_2} \gets \lfloor Pr_2^s\rfloor$
\State $P_{s,r_1,r_2} \gets P_{s,r_1}+P_{s,r_2}$
\If{$P_{s,r_1,r_2} \geq P$}
    \State $LB_s \gets \frac{P_{s,r_1,r_2}-P}{P}$
\Else
    \State $LB_s \gets 0$
\EndIf
\State $LB \gets LB\cup LB_s$
\EndFor
\end{algorithmic}
\end{algorithm}

\section{Per-Dataset Results when Varying $r$ and backbone capacity}\label{sec:consistencySupp}
\begin{figure*}
    \centering
    \includegraphics[width=0.85\textwidth]{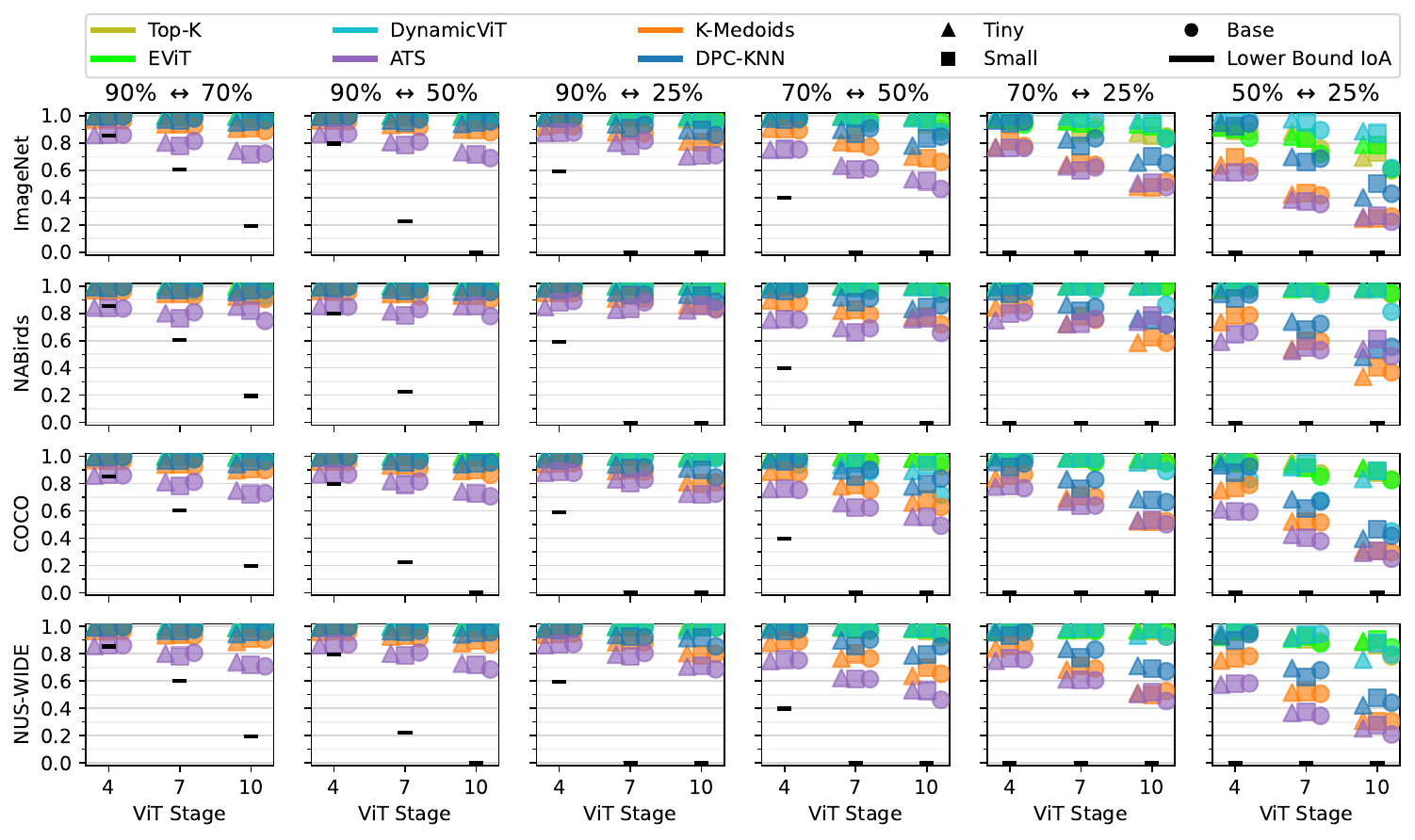}
    \caption{\textbf{Per-Dataset IoA results (Section~\ref{sec:consistencySupp}).} We observe that across all dataset the fixed rate pruning-based methods achieves high IoA scores across all keep rates $r$, indicating the reduction patterns are consistent. On the contrary, the ATS method and the cluster centers of the K-Medoids and DPC-KNN methods have low IoAs, indicating more inconsistent reduction patterns.}
    \label{fig:datasetRatePruning}
\end{figure*}

\begin{figure*}
    \centering
    \includegraphics[width=0.85\textwidth]{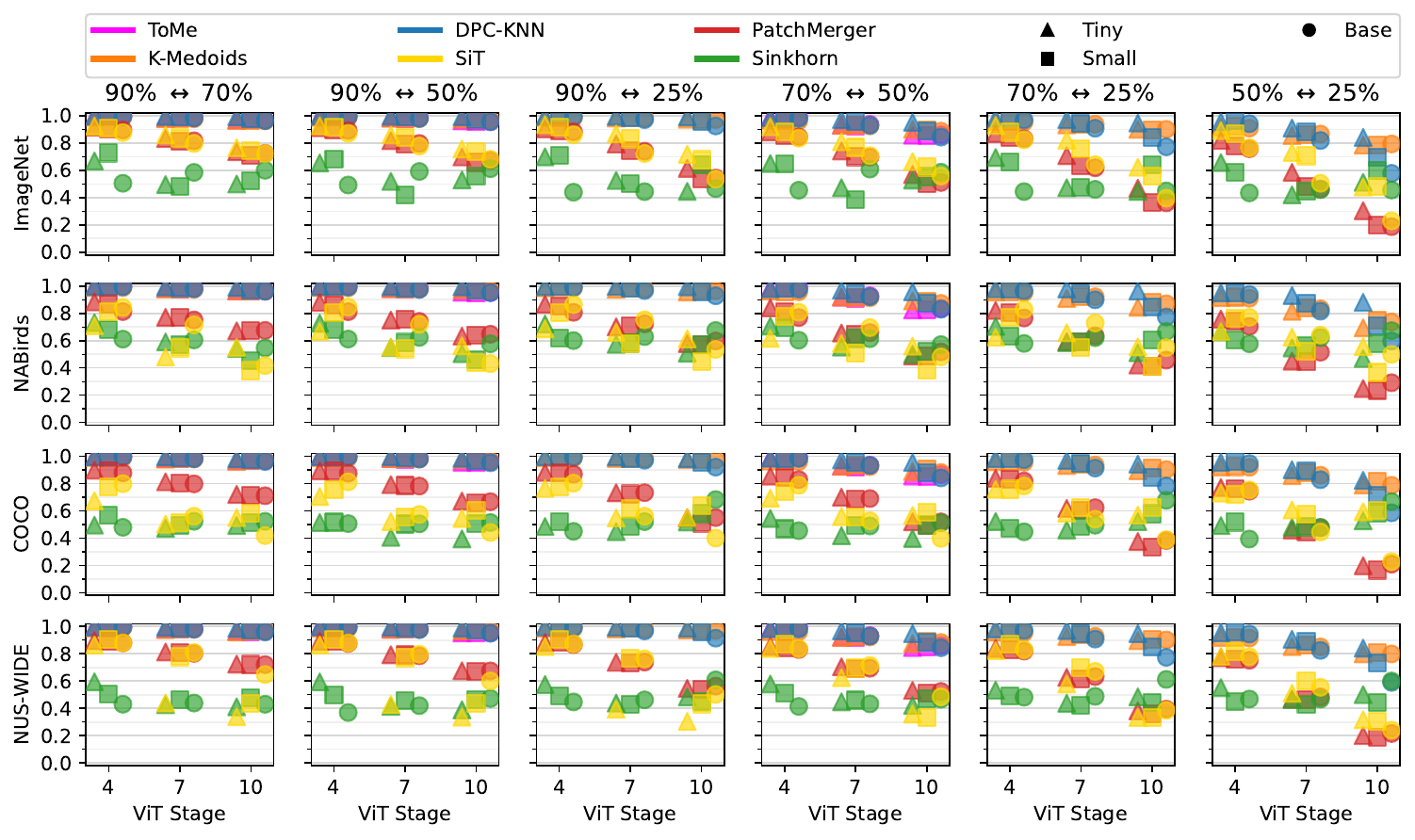}
    \caption{\textbf{Per-Dataset Homogeneity results (Section~\ref{sec:consistencySupp}).} We observe across all datasets that the hard-merging methods achieve a high Homogeneity score, indicating a high consistency of the constructed clusters when varying $r$. We also observe that soft-merging methods generally have lower Homogeneity scores, indicating less consistent clusters. We note that the PatchMerger and SiT methods have high scores at the earlier reduction stages, but that the scores reduces dramatically at later stages.}
    \label{fig:datasetRateClustering}
\end{figure*}

\begin{figure*}
    \centering
    \includegraphics[width=0.93\textwidth]{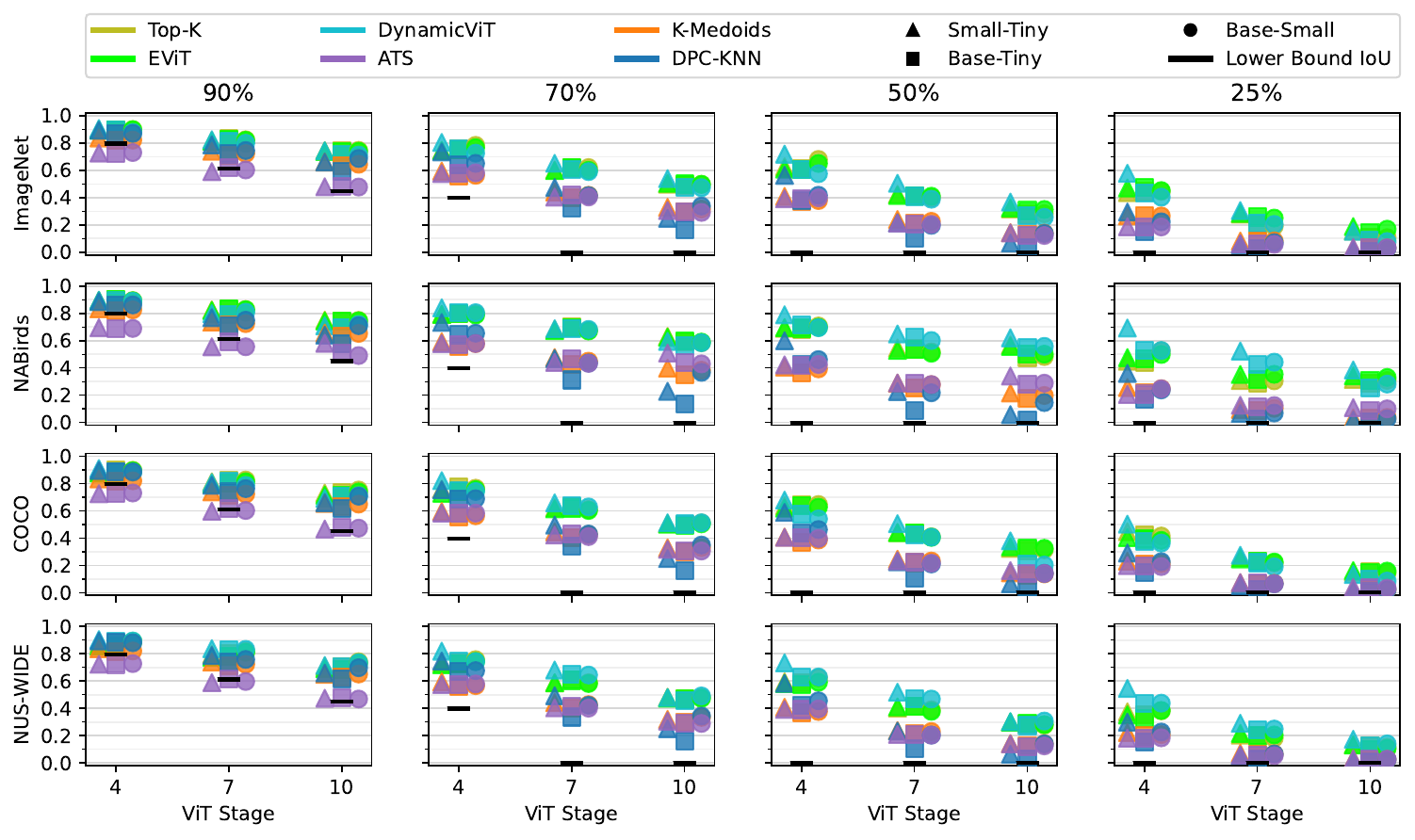}
    \caption{\textbf{Per-Dataset IoU results (Section~\ref{sec:consistencySupp}).} For all datasets we observe that the IoU score is very low for all tested methods. This indicates that the reduction patterns are not consistent when varying the backbone capacity.}
    \label{fig:datasetCapacityPruning}
\end{figure*}

\begin{figure*}
    \centering
    \includegraphics[width=0.93\textwidth]{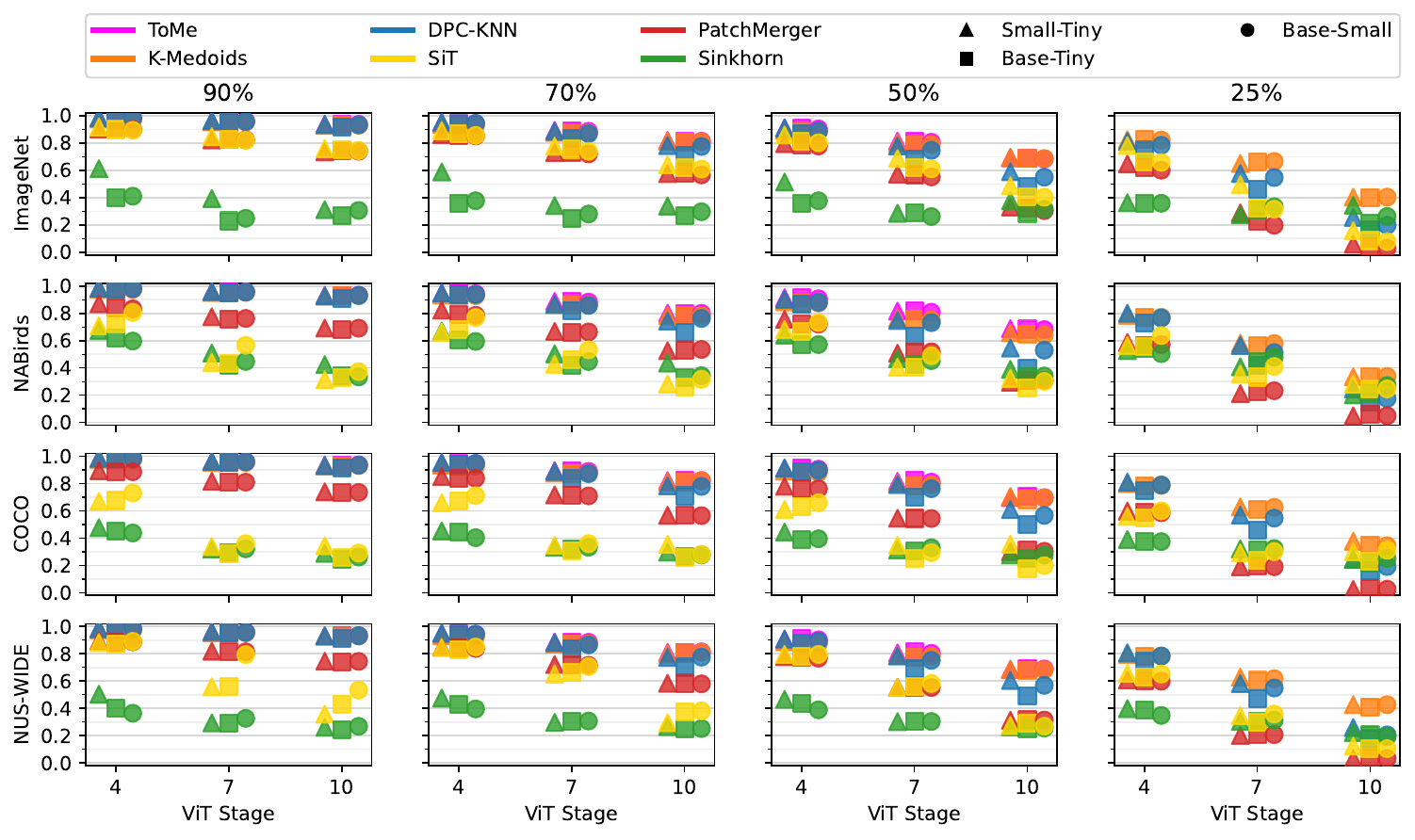}
    \caption{\textbf{Per-Dataset NMI results (Section~\ref{sec:consistencySupp}).} We observe that the hard-merging methods achieve high NMI scores when varying the backbone capacity as long as $r$ is above 25\%, where after it lower dramatically. We observe similar behaviour for the PatchMerger method as long as $r$ is above 50\%.}
    \label{fig:datasetCapacityClustering}
\end{figure*}

In this section, we extend the analysis conducted in Sections~5.1--5.2 by presenting per-dataset results when testing the consistency of reduction patterns under varying keep rate $r$ and backbone capacity; see Figures~\ref{fig:datasetRatePruning}-\ref{fig:datasetCapacityClustering}. For all datasets we find that fixed rate pruning-based reduction patterns are consistent when varying $r$, but inconsistent when varying the backbone capacity. We also observe that the hard-merging methods have a high Homogeneity when varying $r$ indicating the constructed clusters are very consistent, while DPC-KNN and K-Medoids have a low IoU indicating varying cluster centers, similar to the observations made in Section~5.1--5.2. We also observe the Homogeneity to be lower for soft-merging methods for all datasets. Similar to the observations made in Section~5.2, we found that the hard-merging method have consistent reduction patterns when varying the backbone as long as $r$ is above 25\% and 50\% for PatchMerger, while the constructed clusters are inconsistent for the Sinkhorn and SiT methods. These findings match the findings made when analyzing the data aggregated across datasets in Sections~5.1--5.2.

\section{Expanded Cross-Dataset Reduction Pattern Metric Suite}\label{sec:saliencyMetrics}
\begin{figure*}
    \centering
    \includegraphics[width=\linewidth]{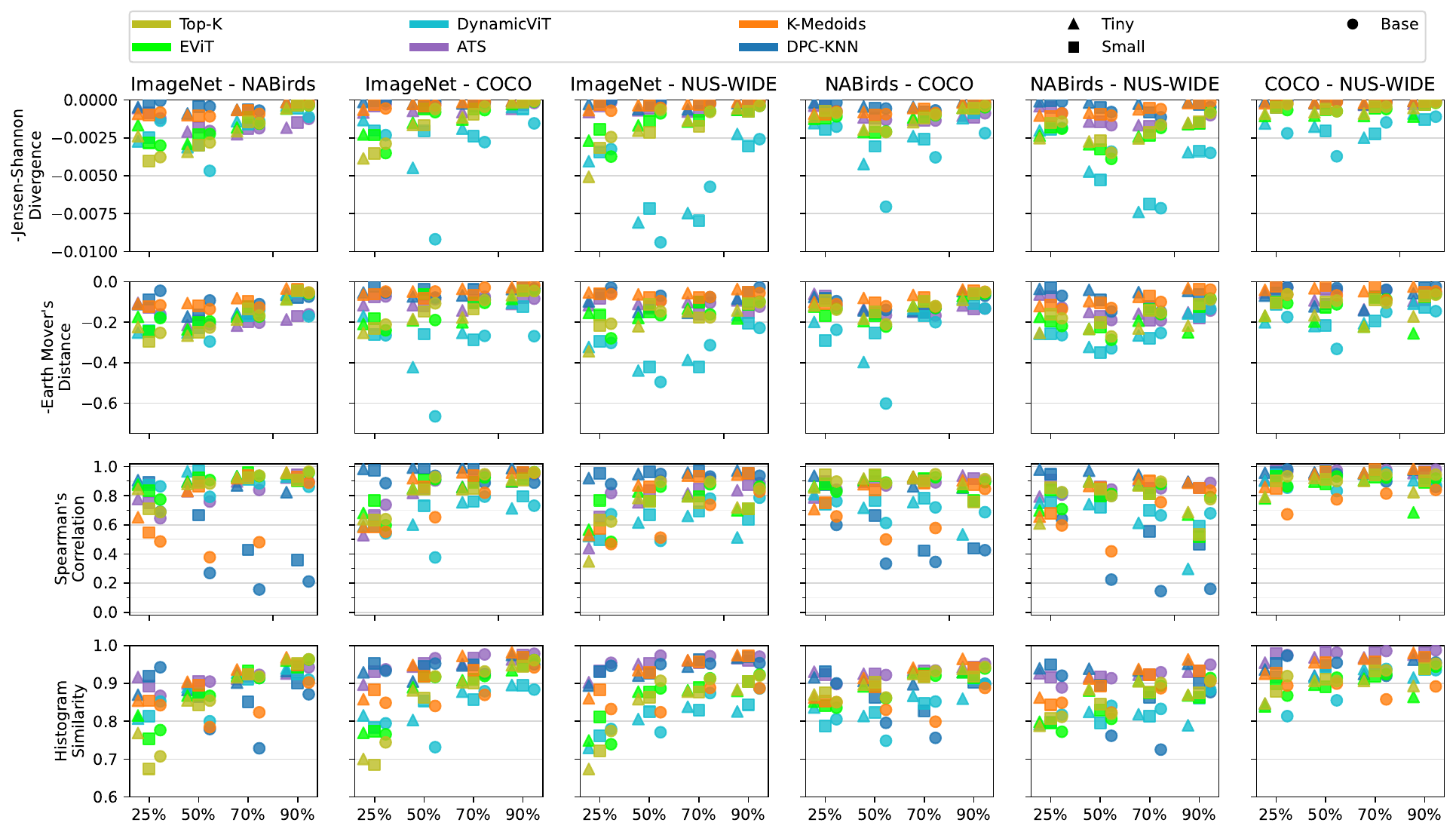}
    \caption{\textbf{Results with expanded averaged reduction pattern metric suite (Section~\ref{sec:saliencyMetrics}).} We compare with an expanded set of saliency metrics \cite{SaliencyMetrics}. Across all metrics we observe high similarity when comparing the dataset-averaged reduction patterns.}
    \label{fig:saliencyMetrics}
\end{figure*}

We extend our analysis of the cross-dataset pruning-based reduction patterns in Section~5.3, by reporting results when using additional metrics from the saliency domain \cite{SaliencyMetrics}. Specifically, we report results using the Spearman's ranked correlation coefficient, Jensen-Shannon Divergence, Earth Mover's Distance, and histogram similarity; see Figure~\ref{fig:saliencyMetrics}. We observe that for all metrics there is a high similarity between reduction patterns from different datasets. Specifically, we note that the results observed when using the Earth's Mover Distance are similar to results obtained with all other metrics. This is noteworthy, as the Earth Mover's Distance is the only metric which incorporates the spatial distance between the tokens, whereas the other metrics interpret the reduction patterns as 1D distributions.

\section{Comparison of Pruning-based and $\ell_p$ Reduction Patterns}\label{sec:lpSupp}
\begin{figure*}
     \begin{subfigure}[b]{\textwidth}
         \centering
        \includegraphics[width=\linewidth]{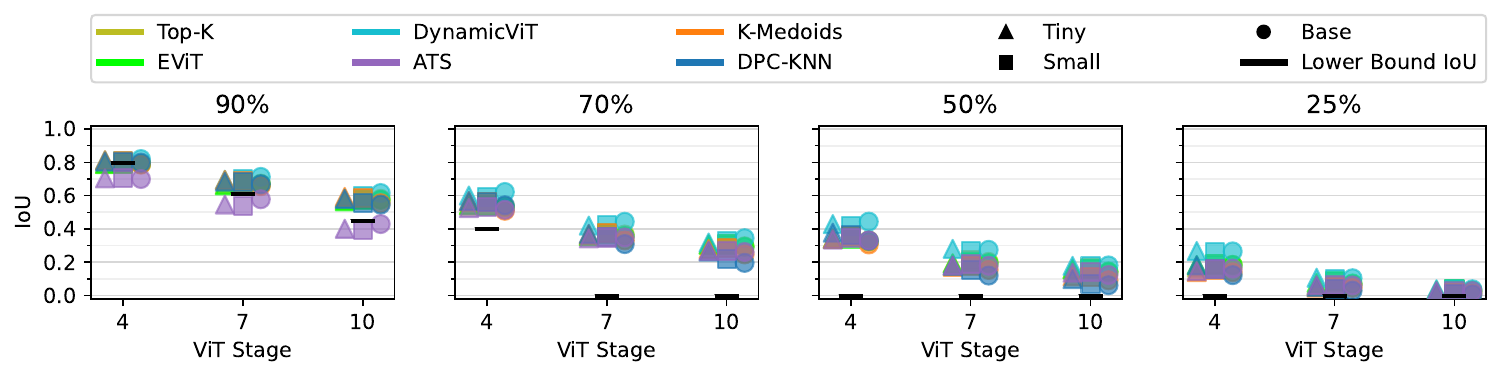}
         \caption{Comparison to $\ell_1$ fixed radial pattern.}
         \label{fig:l1Fig}
     \end{subfigure}
     \hfill
     \begin{subfigure}[b]{\textwidth}
         \centering
        \includegraphics[width=\linewidth]{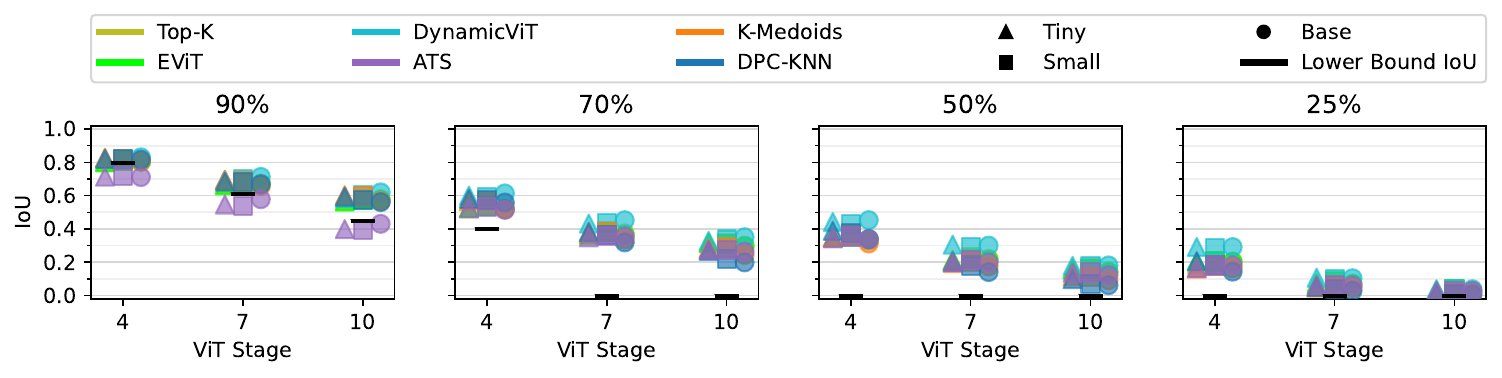}
         \caption{Comparison to $\ell_2$ fixed radial pattern.}
         \label{fig:l2Fig}
     \end{subfigure}
     \hfill
     \begin{subfigure}[b]{\textwidth}
         \centering
        \includegraphics[width=\linewidth]{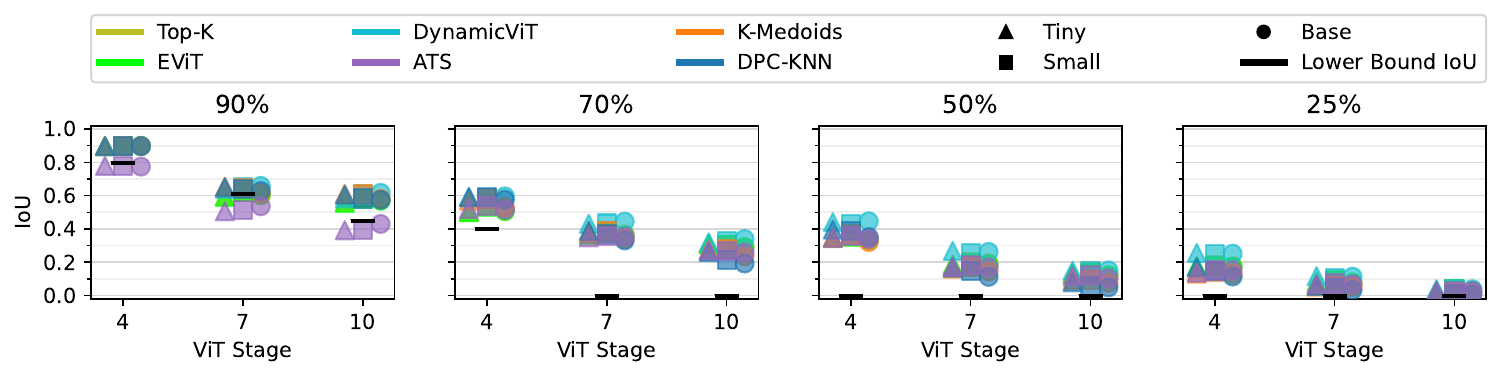}
         \caption{Comparison to $\ell_\infty$ fixed radial pattern.}
         \label{fig:linfFig}
     \end{subfigure}
        \caption{\textbf{Comparing pruning-based reduction patterns to $\ell_p$ fixed reduction patterns (Section~\ref{sec:lpSupp}).} We find that across all reduction stages the IoU between token reduction methods and the $\ell_p$ pruning baselines is very low. This indicates the learned reduction patterns are very different from the fixed radial patterns.} 
        \label{fig:lpNorms}
\end{figure*}

In this section, we present more detailed results of the comparison of learned reduction patterns and the $\ell_p$ reduction patterns in Section~5.4. We report the IoU between the different $\ell_p$ fixed pattern reduction methods and the learned pruning-based reduction methods as well as the DPC-KNN and K-Medoids cluster centers; see Figure~\ref{fig:lpNorms}. It is clear that all methods have a low IoU score across all reduction stages for all three $\ell_p$ methods, indicating that the learned reduction patterns are very different from the fixed image-centered radial patterns applied by the  $\ell_p$.

\section{Extended feature alignment metric suite - CKA and PWCCA analysis}\label{sec:proxySupp}
\begin{figure*}
    \centering
    \includegraphics[width=\linewidth]{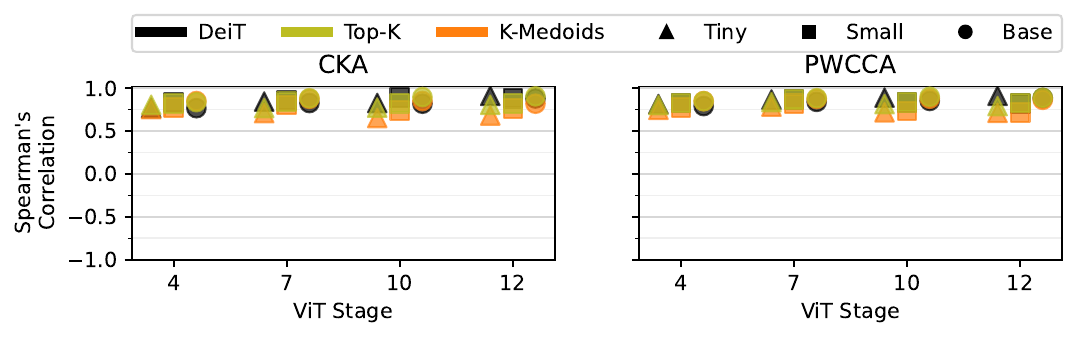}
    \caption{\textbf{Results with extended feature alignment metric suite (Section~\ref{sec:proxySupp}).} We present results when comparing feature alignment using CKA and PWCCA with the difference in model performance. We find a high correlation, similar to what was observed using the orthogonal Procrustes distance.}
    \label{fig:CKAPWCCA}
\end{figure*}

We extend the analysis of whether feature alignment is a good proxy for model performance conducted in Section~5.5, by considering the commonly used metrics: Centered Kernel Alignment (CKA) \cite{CKA_2019} and Projection-Corrected Canonical Correlation Analysis (PWCCA) \cite{CCA_2018}. We follow the procedure laid out by Ding \etal \cite{Procrustes_2021} and make pairwise comparisons between all methods to the three anchor methods: Top-K, K-Medoids, and the baseline DeiT. The results are presented in Figure~\ref{fig:CKAPWCCA}, and we observe that the CKA and PWCCA metrics are as good proxies for model performance as the orthogonal Procrustes Distance, with no noticeable differences in the results.

\section{Model Performance Proxies - Scatter Plots}\label{sec:proxyScatter}
\begin{figure*}
    \centering
    \includegraphics[width=\linewidth]{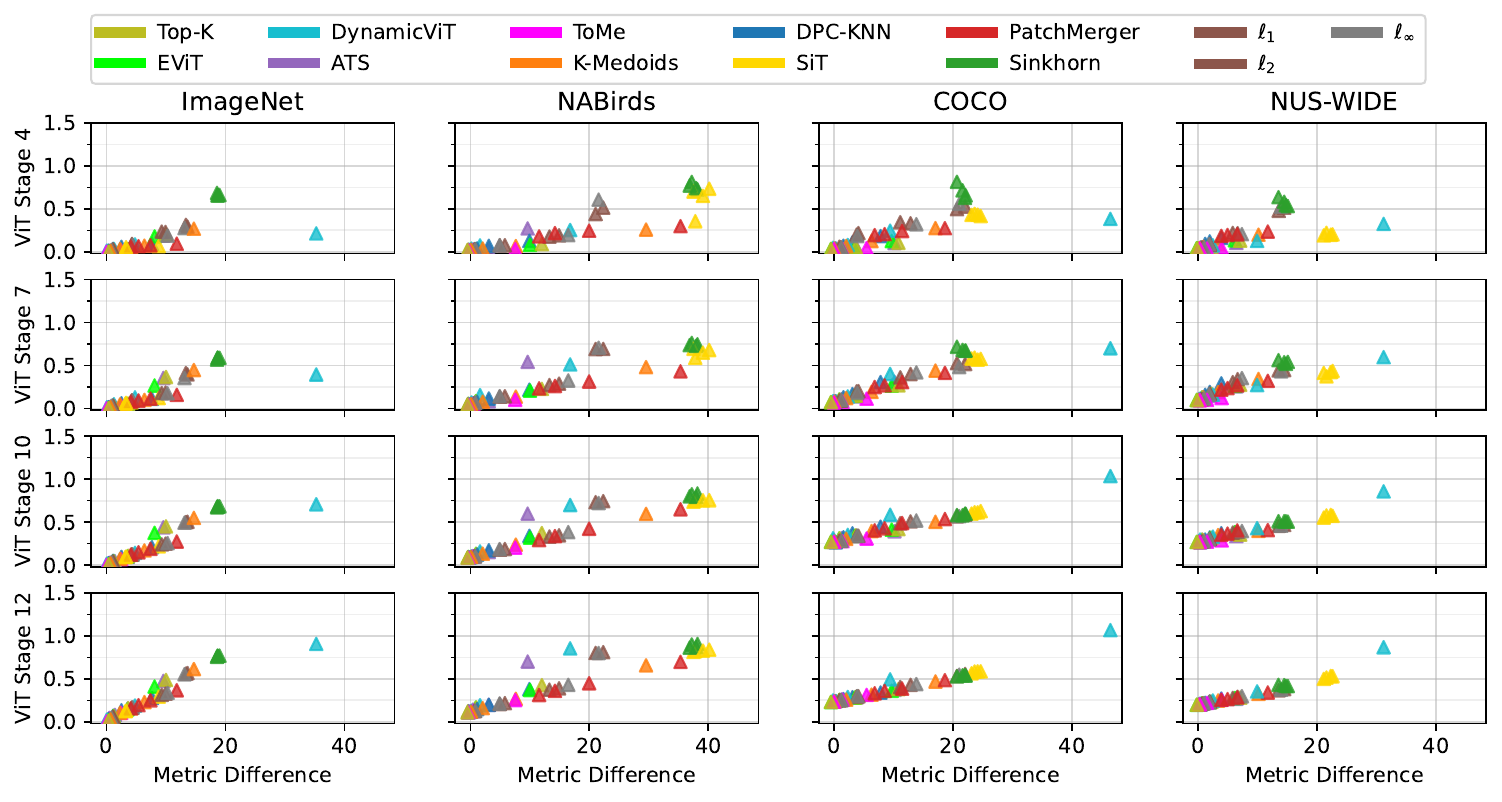}
    \caption{\textbf{Procrustes and DeiT-T anchor model as model performance proxy (Section~\ref{sec:proxyScatter}).} Scatter plot between difference in model performance and the orthogonal Procrustes distance with DeiT-T as anchor model.}
    \label{fig:deitScatterTiny}
\end{figure*}

\begin{figure*}
    \centering
    \includegraphics[width=\linewidth]{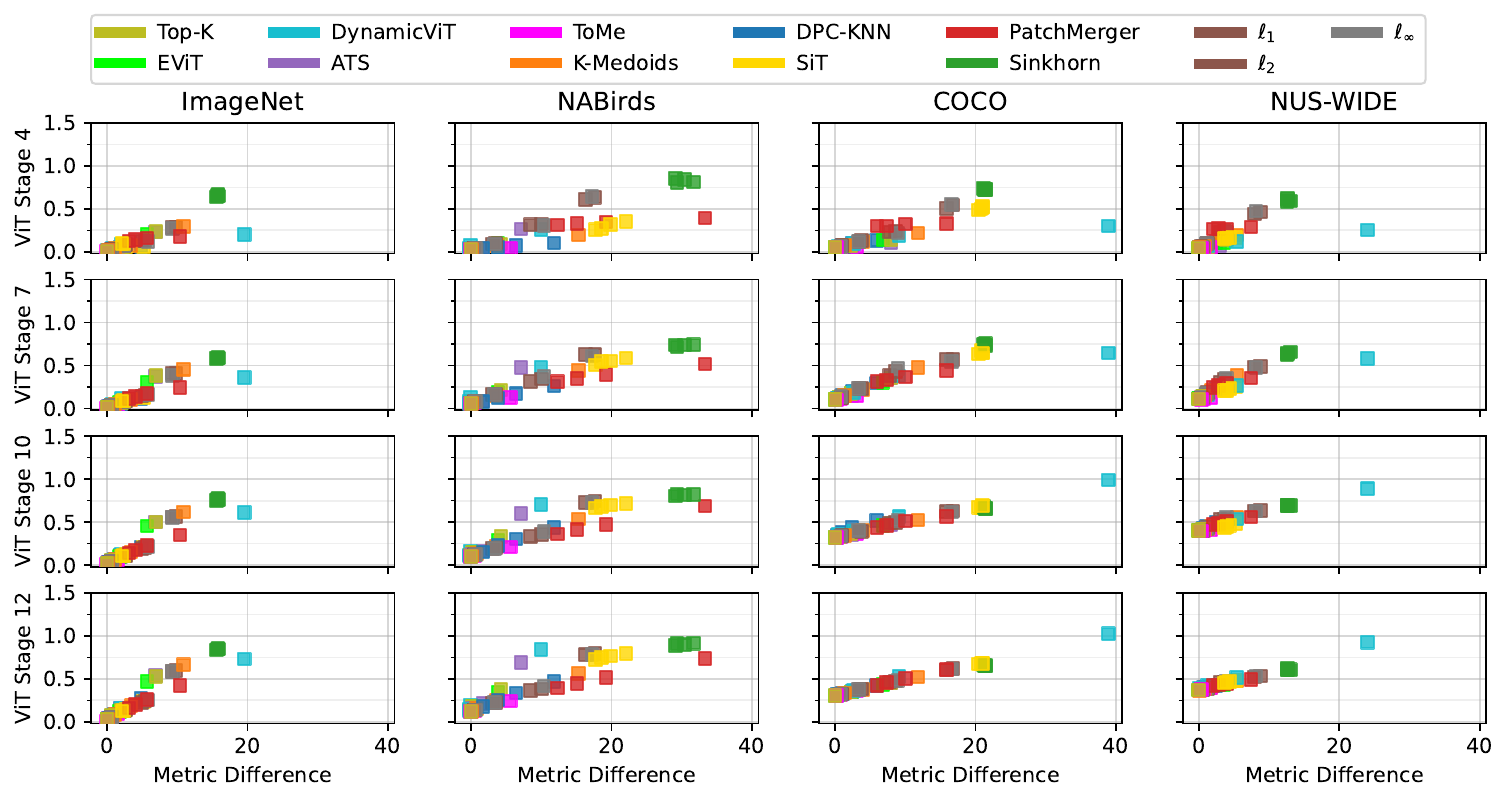}
    \caption{\textbf{Procrustes and DeiT-S anchor model as model performance proxy (Section~\ref{sec:proxyScatter}).} Scatter plot between difference in model performance and the orthogonal Procrustes distance with DeiT-S as anchor model.}
    \label{fig:deitScatterSmall}
\end{figure*}

\begin{figure*}
    \centering
    \includegraphics[width=\linewidth]{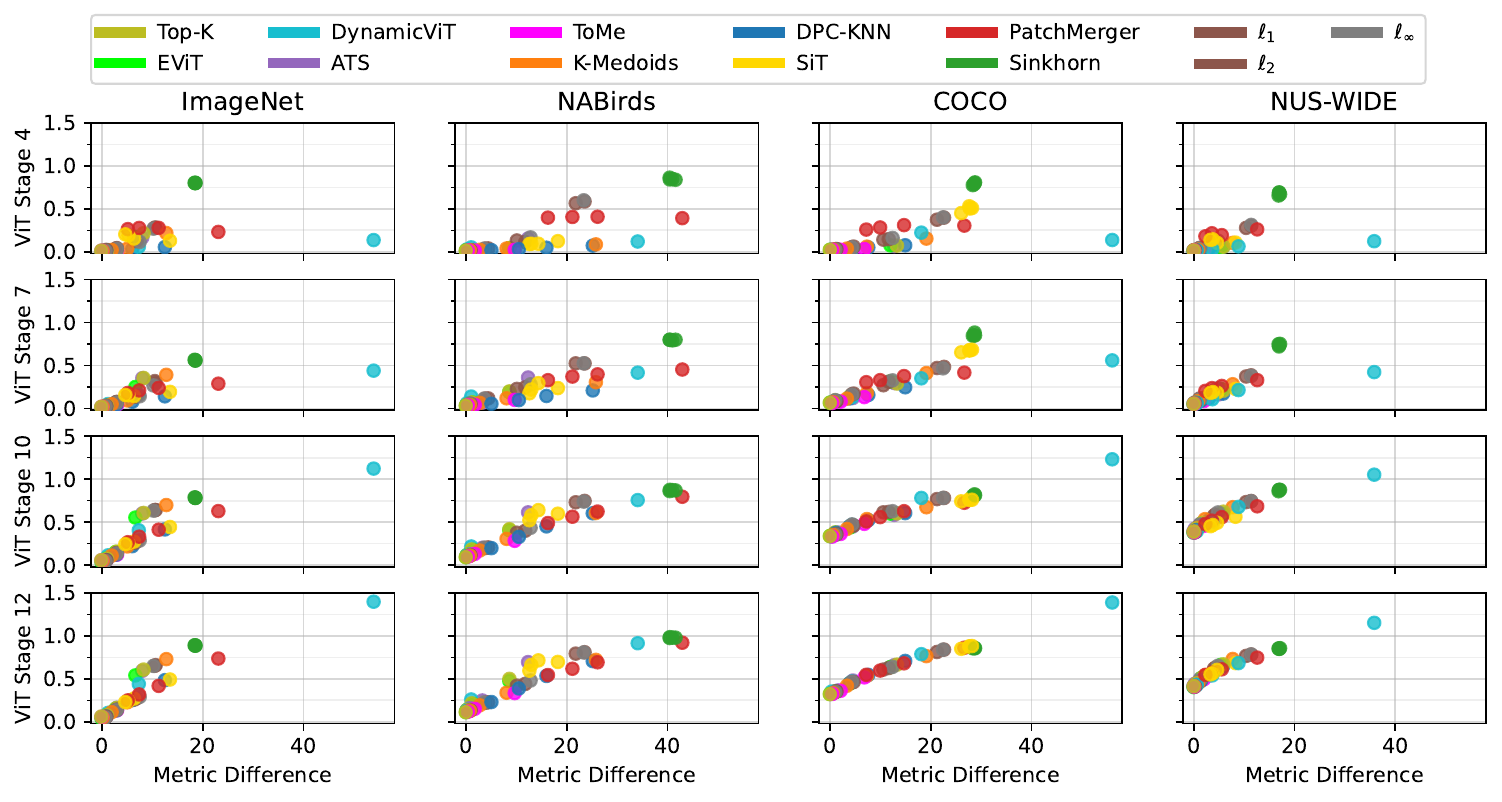}
    \caption{\textbf{Procrustes and DeiT-B anchor model as model performance proxy (Section~\ref{sec:proxyScatter}).} Scatter plot between difference in model performance and the orthogonal Procrustes distance with DeiT-B as anchor model.}
    \label{fig:deitScatterBase}
\end{figure*}

\begin{figure*}
    \centering
    \includegraphics[width=\linewidth]{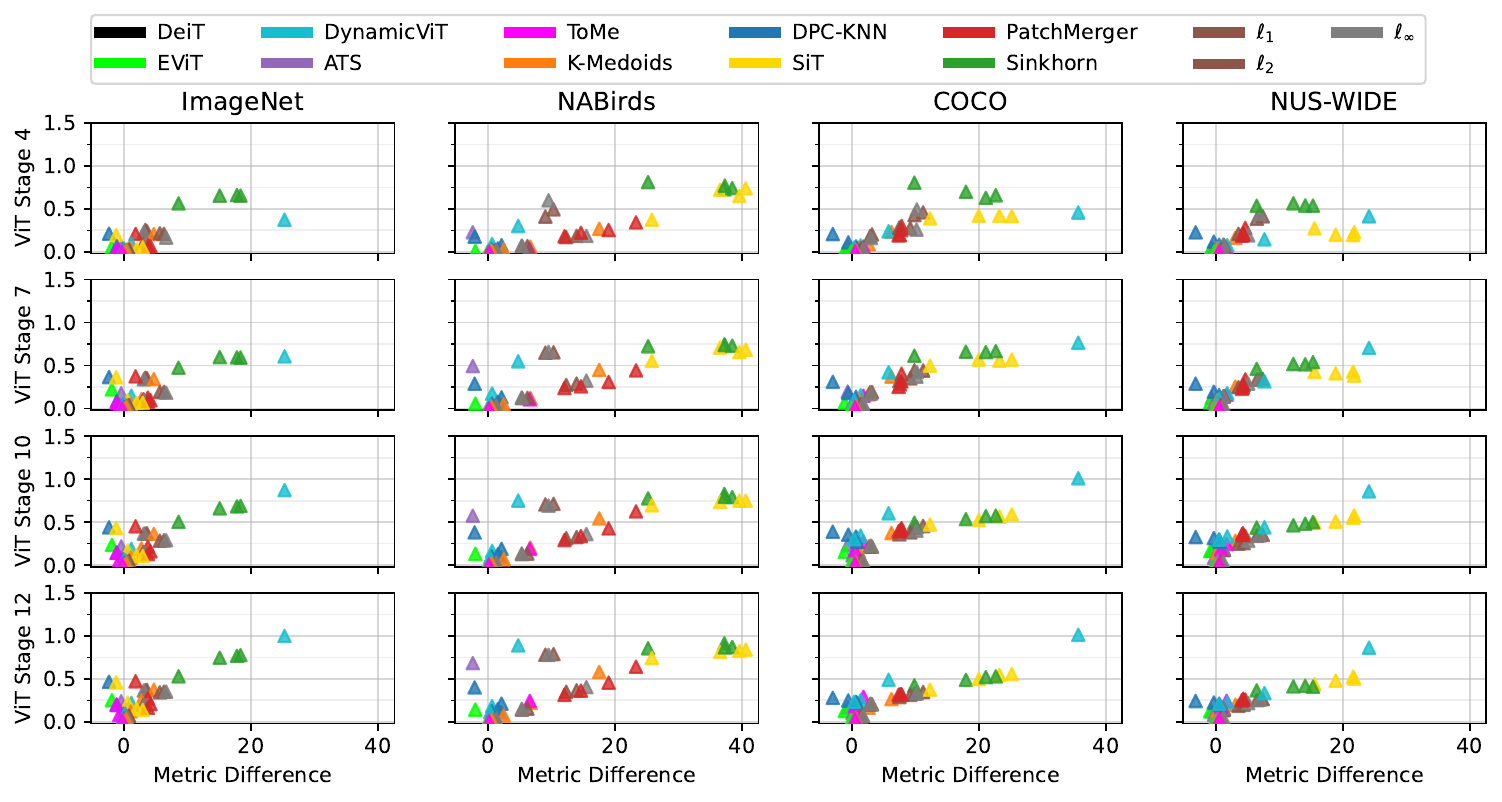}
    \caption{\textbf{Procrustes and Top-K anchor model with a DeiT-T backbone as model performance proxy (Section~\ref{sec:proxyScatter}).} Scatter plot between difference in model performance and the orthogonal Procrustes distance with Top-K as anchor model.}
    \label{fig:topkScatterTiny}
\end{figure*}

\begin{figure*}
    \centering
    \includegraphics[width=\linewidth]{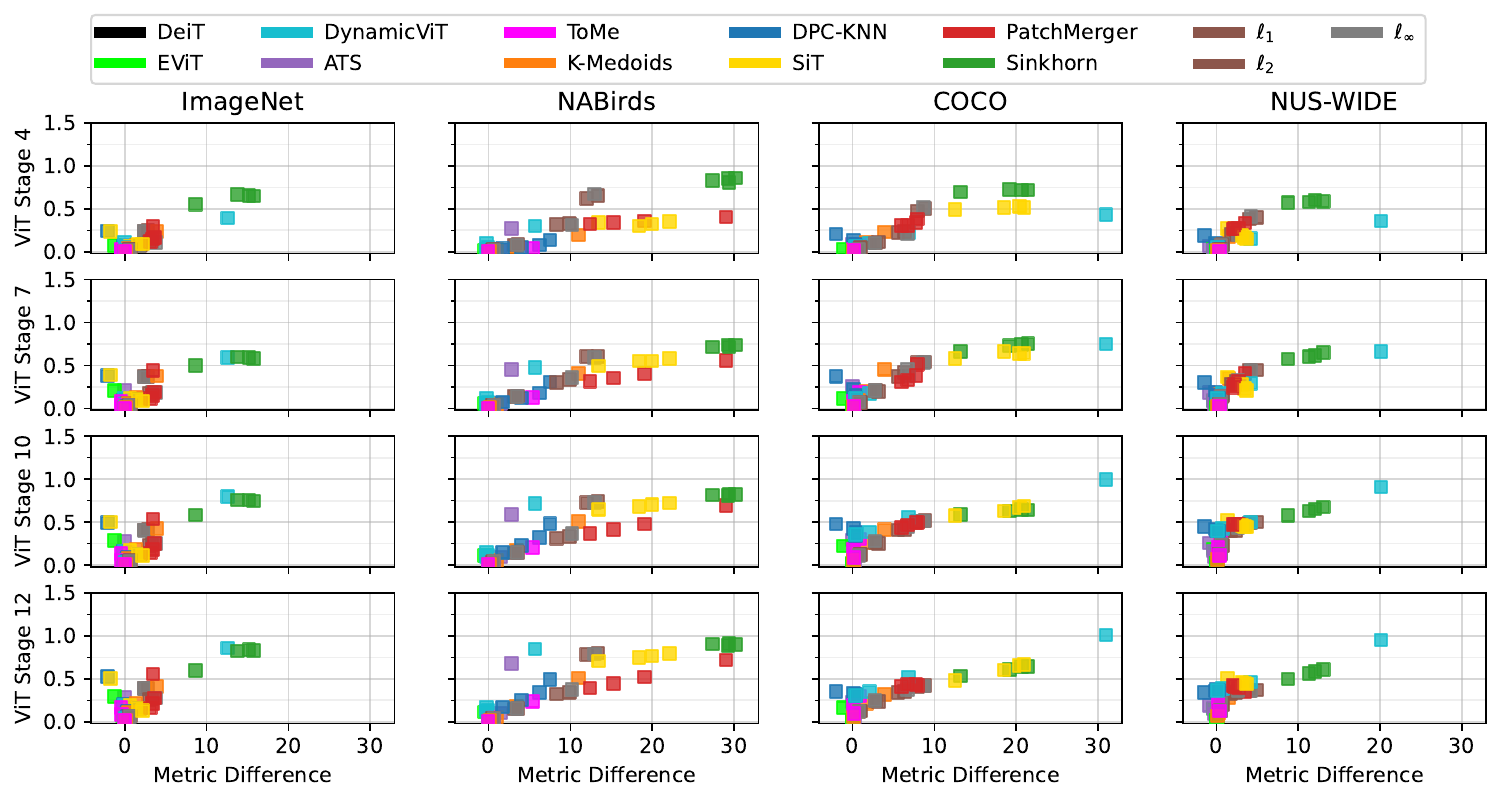}
    \caption{\textbf{Procrustes and Top-K anchor model with a DeiT-S backbone as model performance proxy (Section~\ref{sec:proxyScatter}).} Scatter plot between difference in model performance and the orthogonal Procrustes distance with Top-K as anchor model.}
    \label{fig:topkScatterSmall}
\end{figure*}

\begin{figure*}
    \centering
    \includegraphics[width=\linewidth]{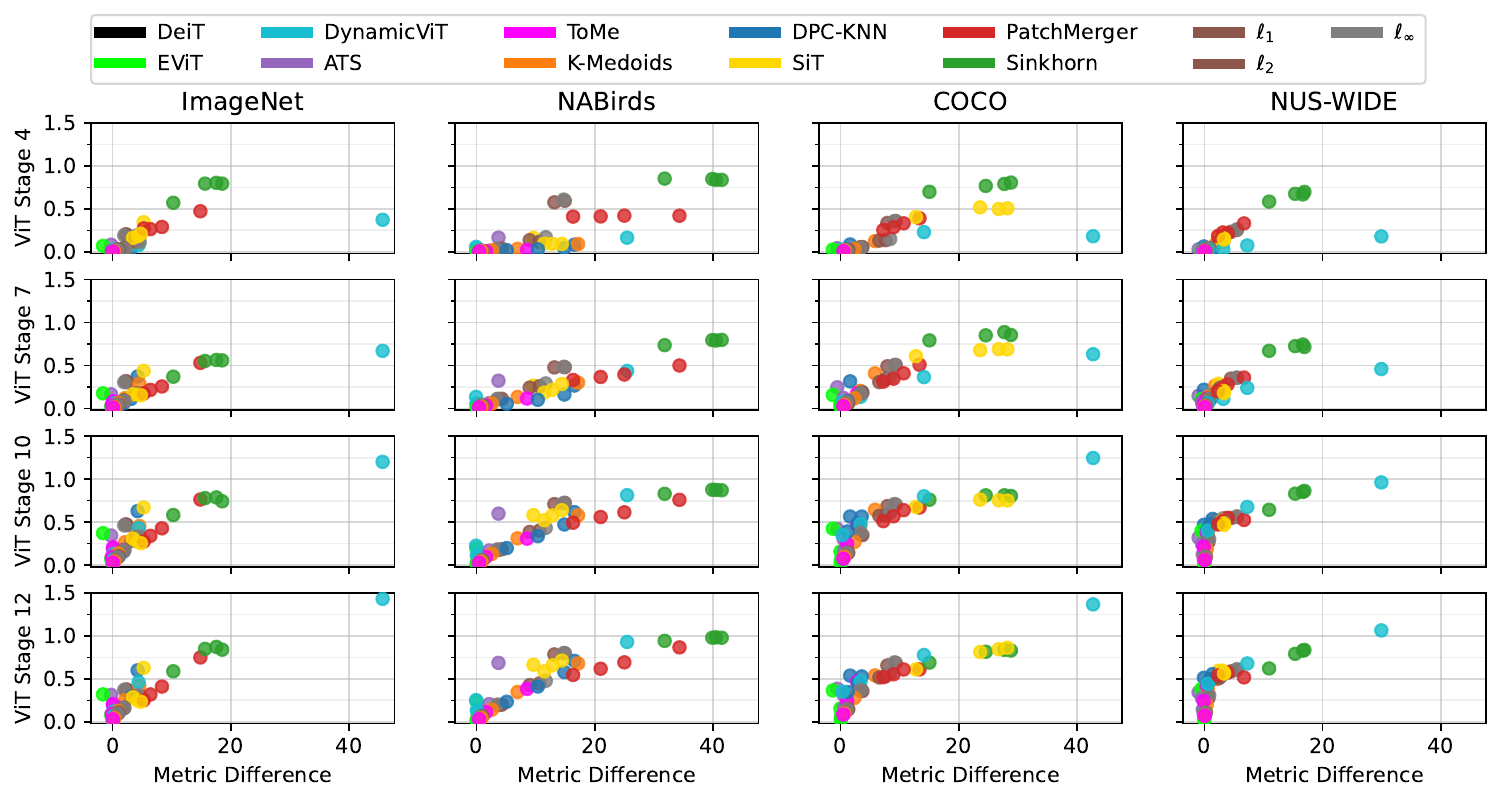}
    \caption{\textbf{Procrustes and Top-K anchor model with a DeiT-B backbone as model performance proxy (Section~\ref{sec:proxyScatter}).} Scatter plot between difference in model performance and the orthogonal Procrustes distance with Top-K as anchor model.}
    \label{fig:topkScatterBase}
\end{figure*}

\begin{figure*}
    \centering
    \includegraphics[width=\linewidth]{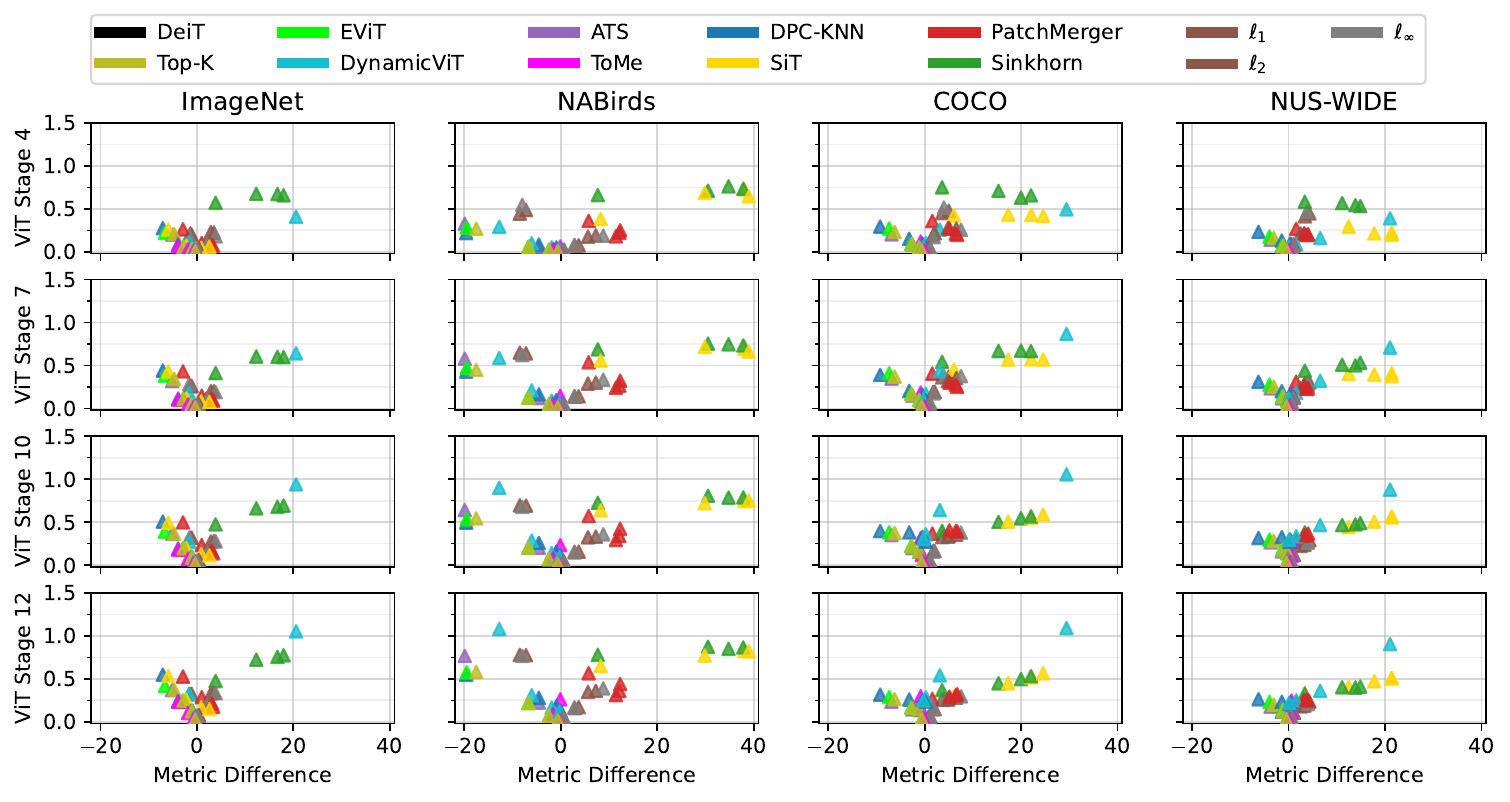}
    \caption{\textbf{Procrustes and K-Medoids anchor model with a DeiT-T backbone as model performance proxy (Section~\ref{sec:proxyScatter}).} Scatter plot between difference in model performance and the orthogonal Procrustes distance with K-Medoids as anchor model.}
    \label{fig:kmedoidsScatterTiny}
\end{figure*}

\begin{figure*}
    \centering
    \includegraphics[width=\linewidth]{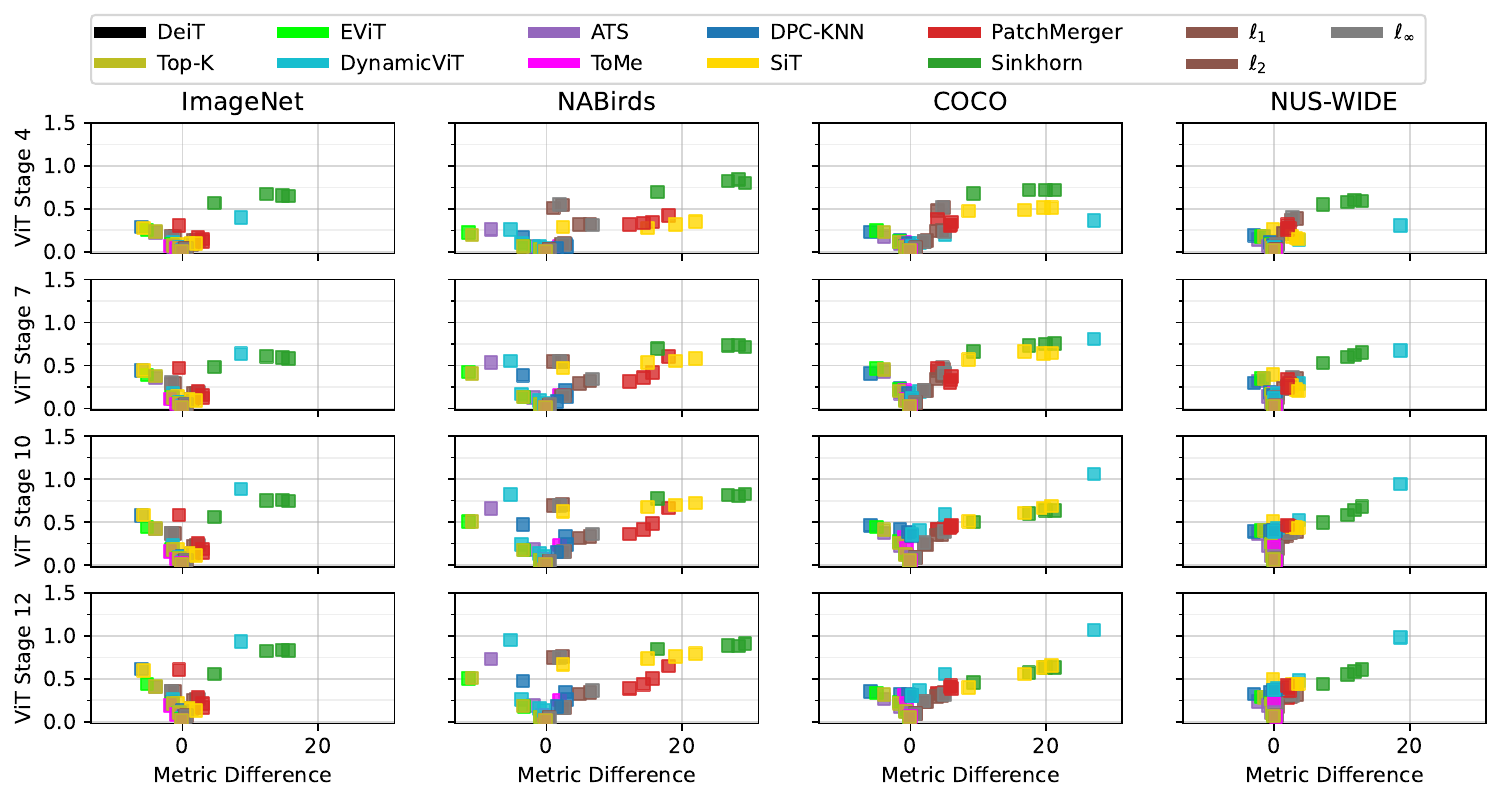}
    \caption{\textbf{Procrustes and K-Medoids anchor model with a DeiT-S backbone as model performance proxy (Section~\ref{sec:proxyScatter}).} Scatter plot between difference in model performance and the orthogonal Procrustes distance with K-Medoids as anchor model.}
    \label{fig:kmedoidsScatterSmall}
\end{figure*}

\begin{figure*}
    \centering
    \includegraphics[width=\linewidth]{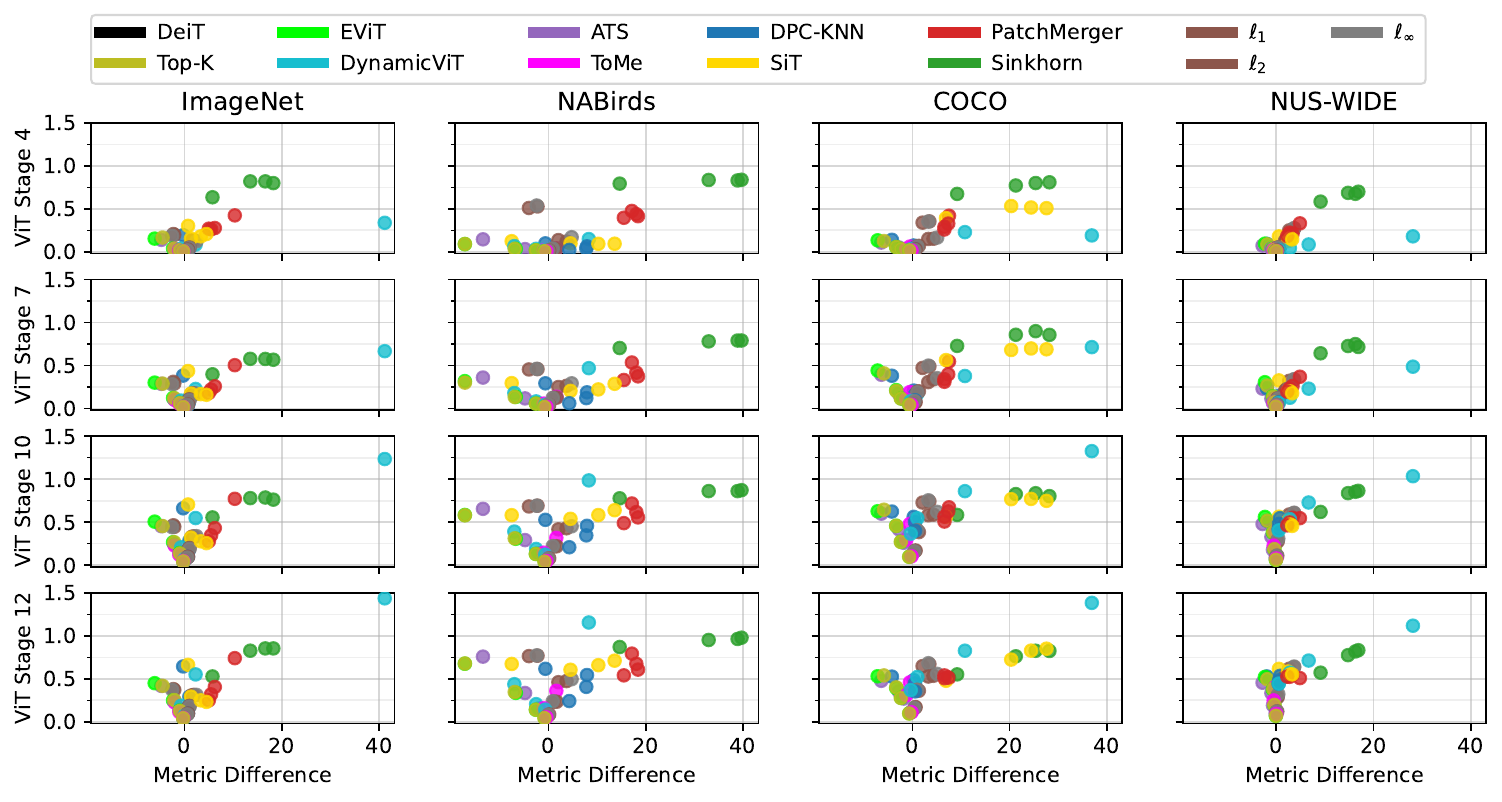}
    \caption{\textbf{Procrustes and K-Medoids anchor model with a DeiT-B backbone as model performance proxy (Section~\ref{sec:proxyScatter}).} Scatter plot between difference in model performance and the orthogonal Procrustes distance with K-Medoids as anchor model.}
    \label{fig:kmedoidsScatterBase}
\end{figure*}

\begin{figure*}
    \centering
    \includegraphics[width=\linewidth]{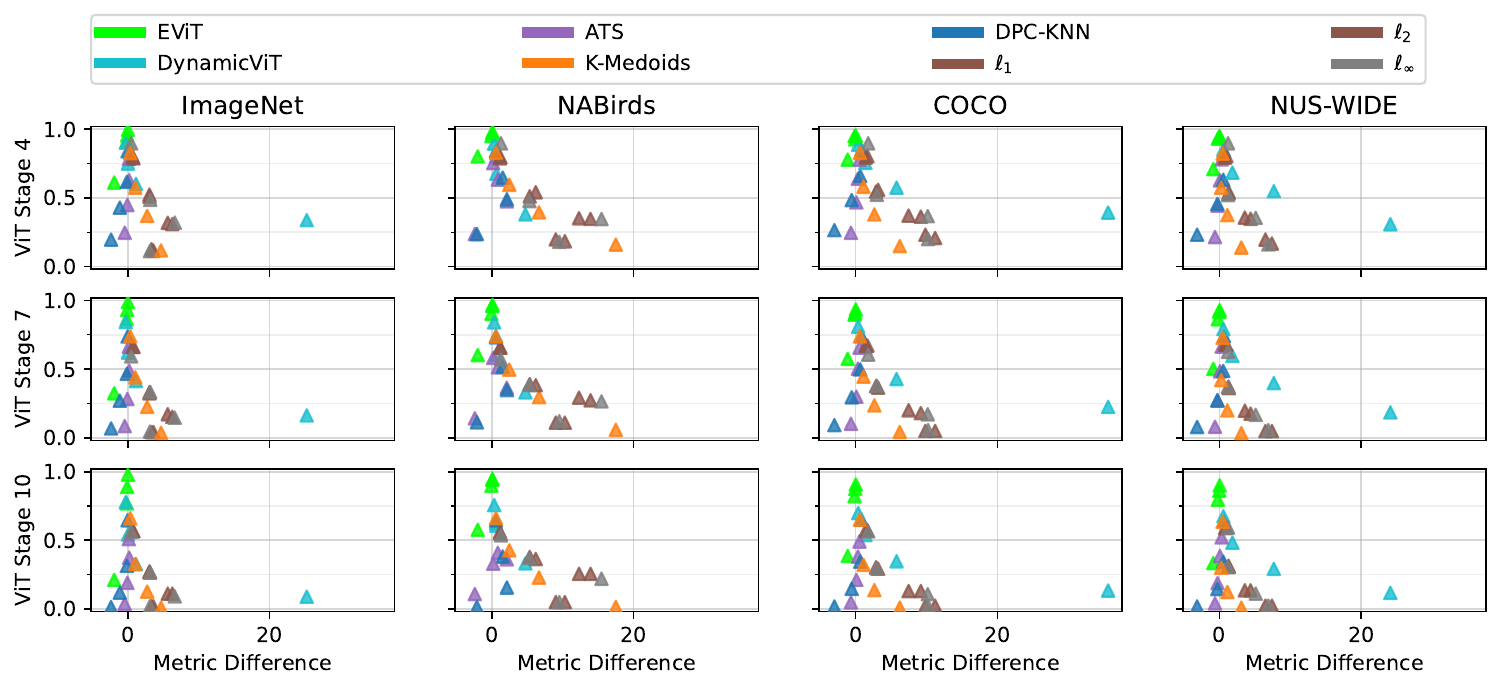}
    \caption{\textbf{IoU and Top-K anchor model with a DeiT-T backbone as model performance proxy (Section~\ref{sec:proxyScatter}).} Scatter plot between difference in performance and the IoU with Top-K as anchor model.}
    \label{fig:topkIoUScatterTiny}
\end{figure*}

\begin{figure*}
    \centering
    \includegraphics[width=\linewidth]{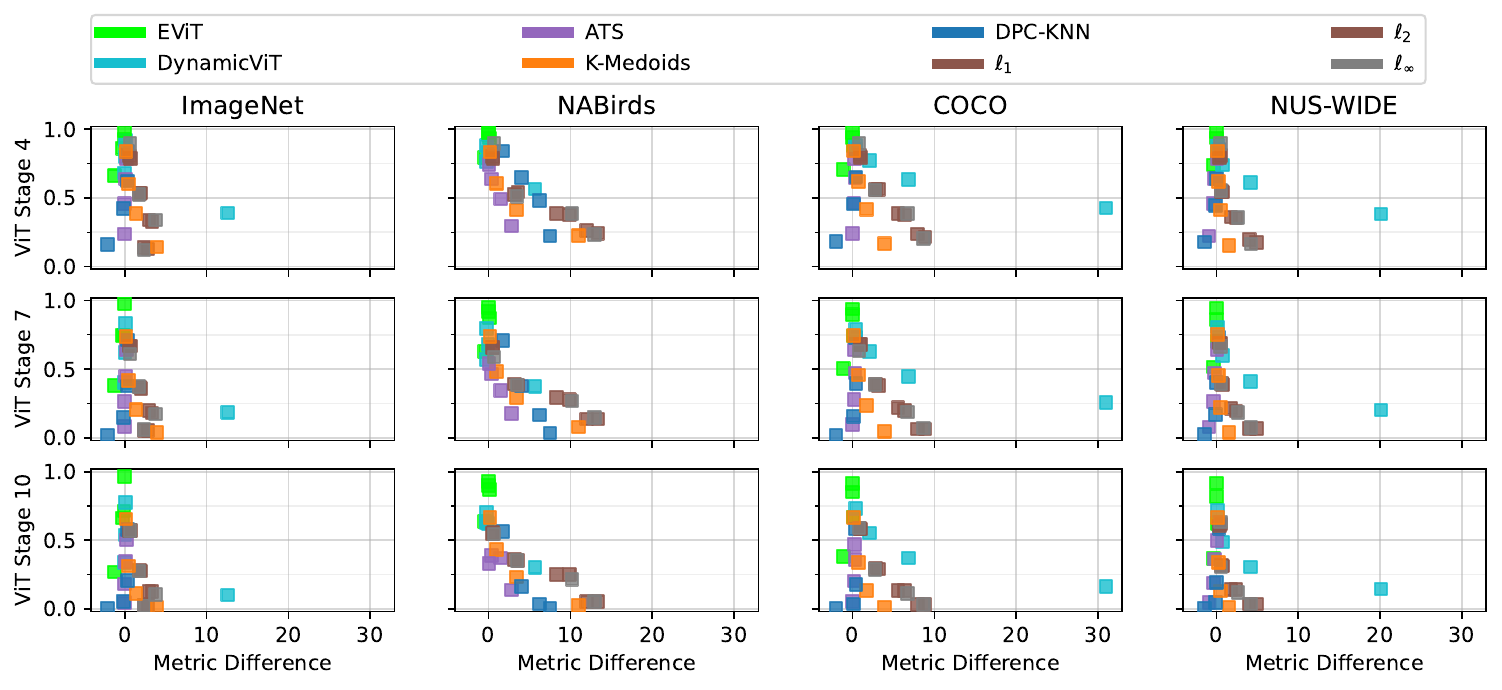}
    \caption{\textbf{IoU and Top-K anchor model with a DeiT-S backbone as model performance proxy (Section~\ref{sec:proxyScatter}).} Scatter plot between difference in performance and the IoU with Top-K as anchor model.}
    \label{fig:topkIoUScatterSmall}
\end{figure*}

\begin{figure*}
    \centering
    \includegraphics[width=\linewidth]{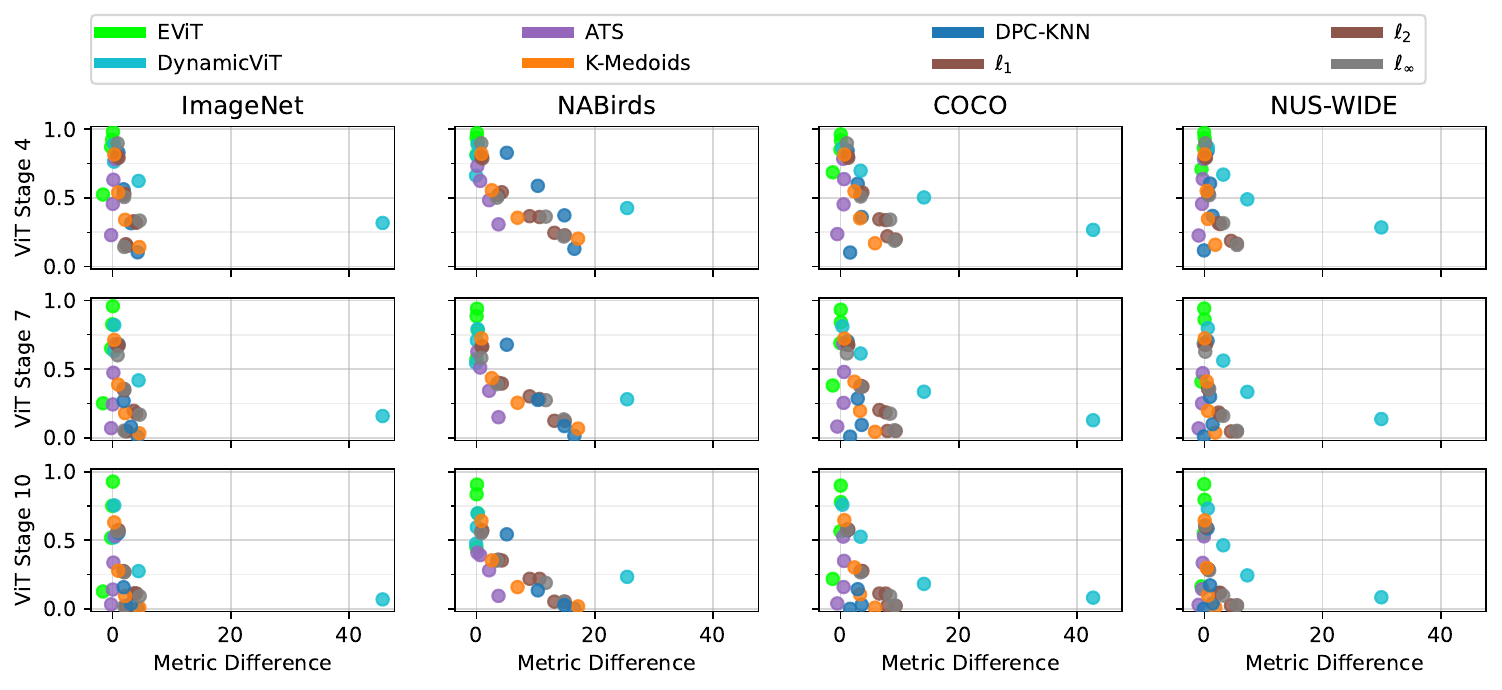}
    \caption{\textbf{IoU and Top-K anchor model with a DeiT-B backbone as model performance proxy (Section~\ref{sec:proxyScatter}).} Scatter plot between difference in performance and the IoU with Top-K as anchor model.}
    \label{fig:topkIoUScatterBase}
\end{figure*}

\begin{figure*}
    \centering
    \includegraphics[width=\linewidth]{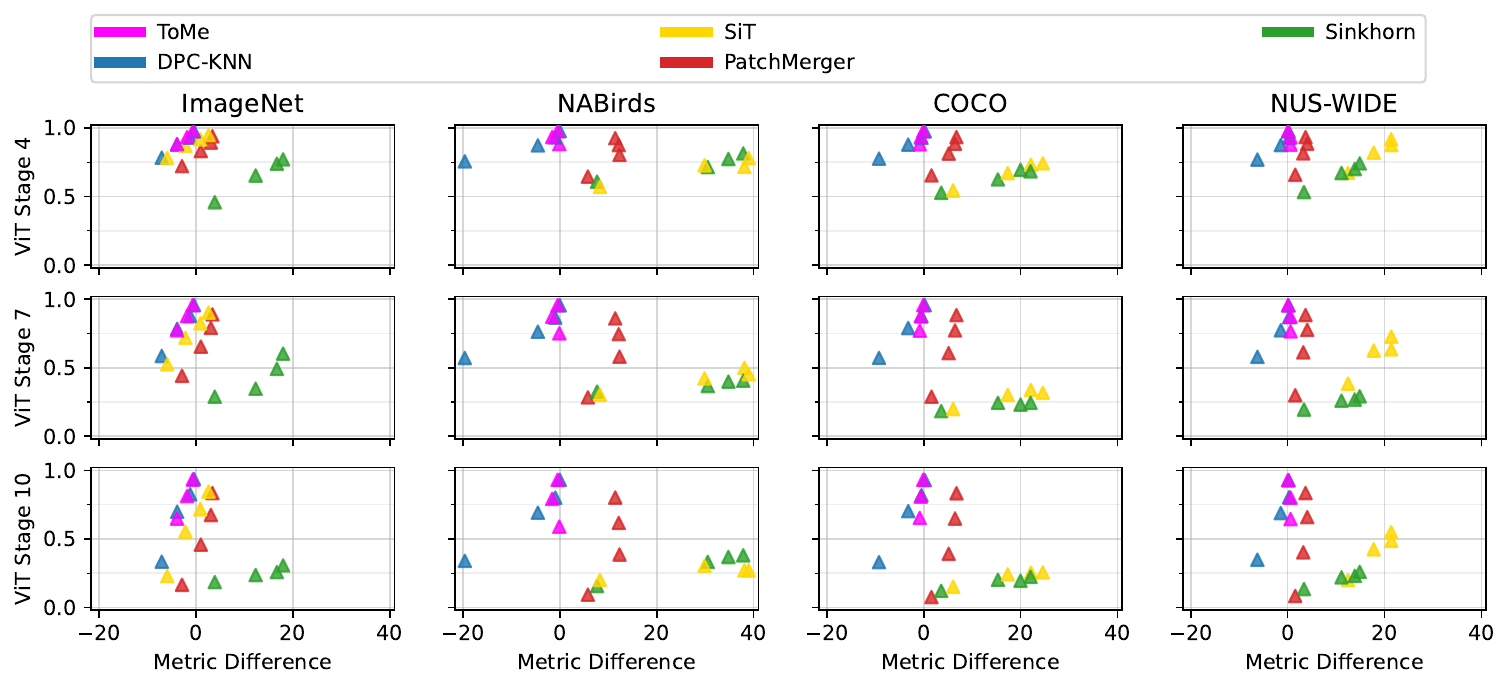}
    \caption{\textbf{NMI and K-Medoids anchor model with a DeiT-T backbone as model performance proxy (Section~\ref{sec:proxyScatter}).} Scatter plot between difference in model performance and the NMI with K-Medoids as anchor model.}
    \label{fig:kmedoidsNMIScatterTiny}
\end{figure*}

\begin{figure*}
    \centering
    \includegraphics[width=\linewidth]{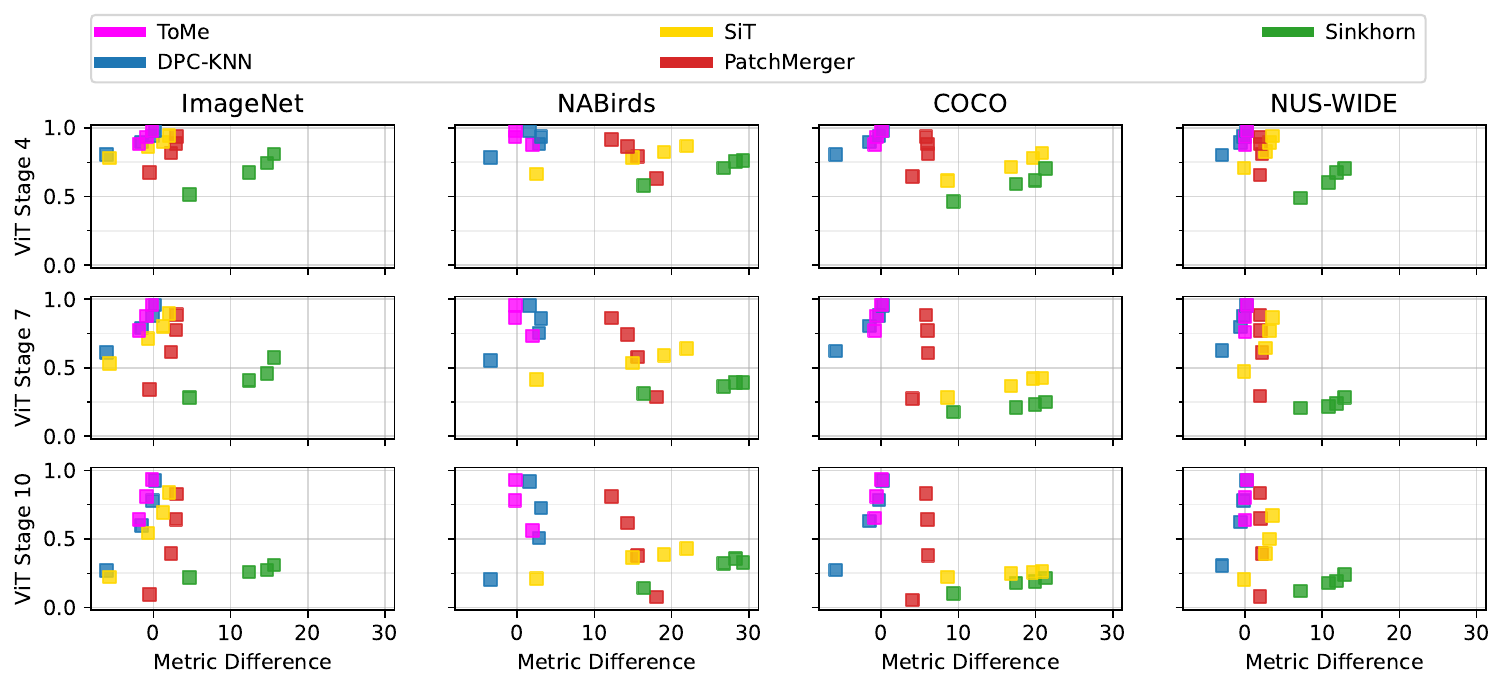}
    \caption{\textbf{NMI and K-Medoids anchor model with a DeiT-S backbone as model performance proxy (Section~\ref{sec:proxyScatter}).} Scatter plot between difference in model performance and the NMI with K-Medoids as anchor model.}
    \label{fig:kmedoidsNMIScatterSmall}
\end{figure*}

\begin{figure*}
    \centering
    \includegraphics[width=\linewidth]{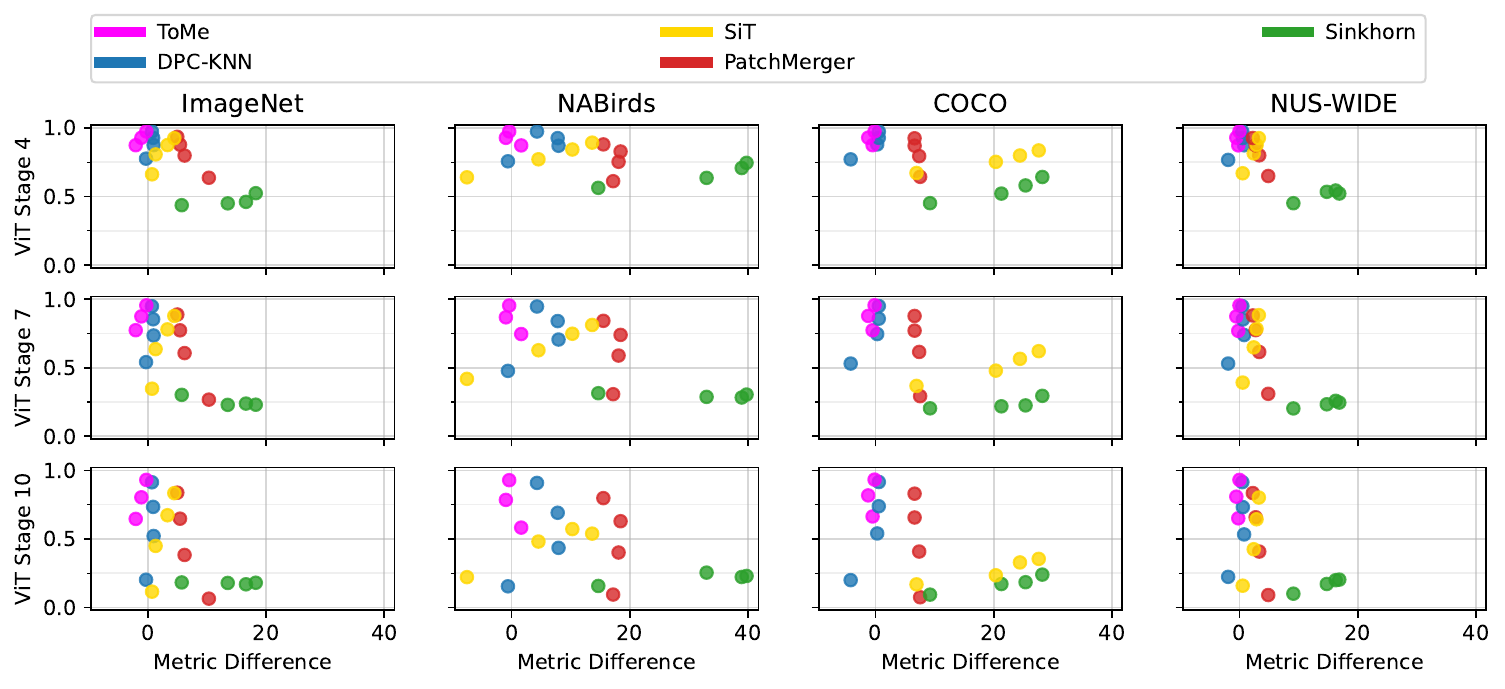}
    \caption{\textbf{NMI and K-Medoids anchor model with a DeiT-B backbone as model performance proxy (Section~\ref{sec:proxyScatter}).} Scatter plot between difference in model performance and the NMI with K-Medoids as anchor model.}
    \label{fig:kmedoidsNMIScatterBase}
\end{figure*}

As described in Section~5.5, we find that reduction pattern similarity and CLS token feature alignment are moderate-to-strong proxies of model performance. We present scatter plots comparing the metric difference between the anchor model and all other models against the orthogonal Procrustes distance, IoU, and NMI; see Figures~\ref{fig:deitScatterTiny}-\ref{fig:kmedoidsNMIScatterBase}. Note that for the sake of brevity the results from different keep rates are plotted in the same plot, but do report separate results per backbone model capacities.

\section{Visualization of Dataset Averaged Reduction Patterns}\label{sec:global}
\begin{figure*}
     \begin{subfigure}[b]{0.33\textwidth}
         \centering
        \includegraphics[width=\linewidth]{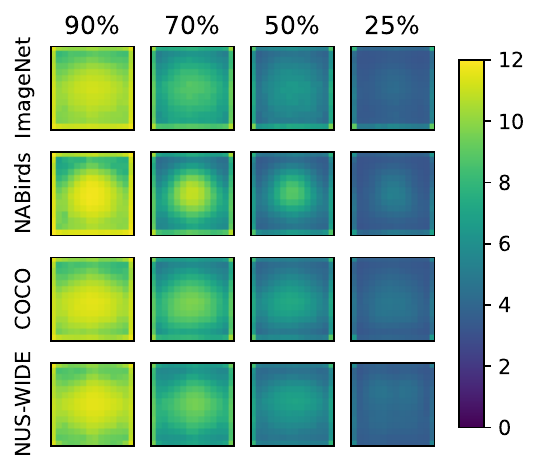}
         \caption{DeiT-T Backbone}
         \label{fig:topkHeatTiny}
     \end{subfigure}
     \hfill
     \begin{subfigure}[b]{0.33\textwidth}
         \centering
        \includegraphics[width=\linewidth]{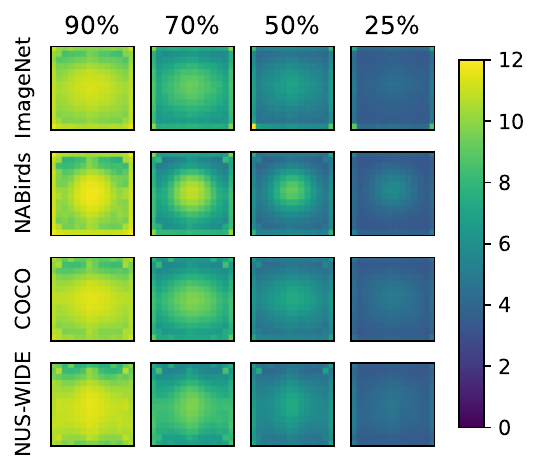}
         \caption{DeiT-S Backbone}
         \label{fig:topkHeatSmall}
     \end{subfigure}
     \hfill
     \begin{subfigure}[b]{0.33\textwidth}
         \centering
        \includegraphics[width=\linewidth]{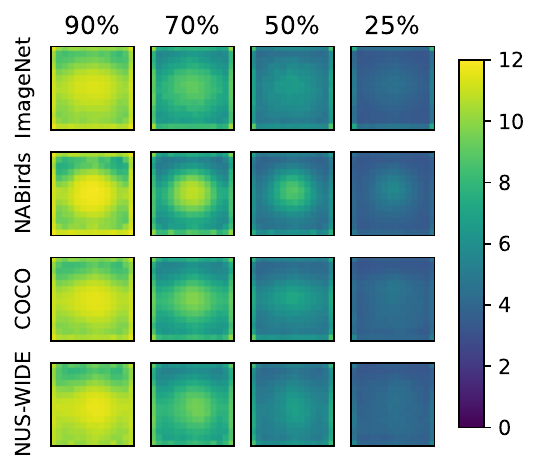}
         \caption{DeiT-B Backbone}
         \label{fig:TopKHeatBase}
     \end{subfigure}
        \caption{\textbf{Top-K Dataset-averaged reduction patterns (Section~\ref{sec:global}).} We find that the Top-K method on average select tokens from both the image center and border.}
        \label{fig:topkHeatGlobal}
\end{figure*}

\begin{figure*}
     \begin{subfigure}[b]{0.33\textwidth}
         \centering
        \includegraphics[width=\linewidth]{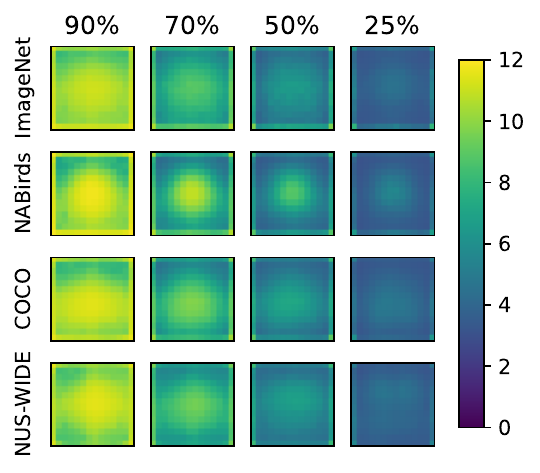}
         \caption{DeiT-T Backbone}
         \label{fig:evitHeatTiny}
     \end{subfigure}
     \hfill
     \begin{subfigure}[b]{0.33\textwidth}
         \centering
        \includegraphics[width=\linewidth]{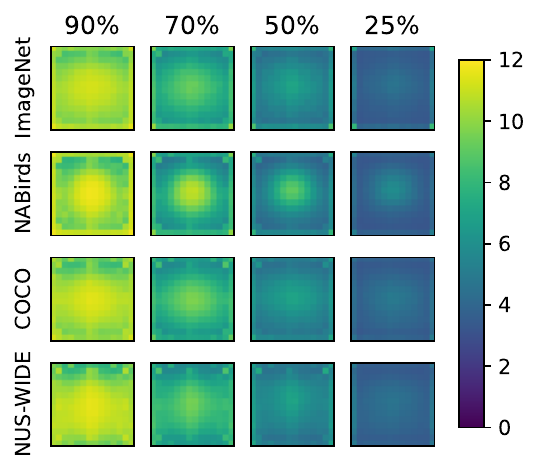}
         \caption{DeiT-S Backbone}
         \label{fig:evitHeatSmall}
     \end{subfigure}
     \hfill
     \begin{subfigure}[b]{0.33\textwidth}
         \centering
        \includegraphics[width=\linewidth]{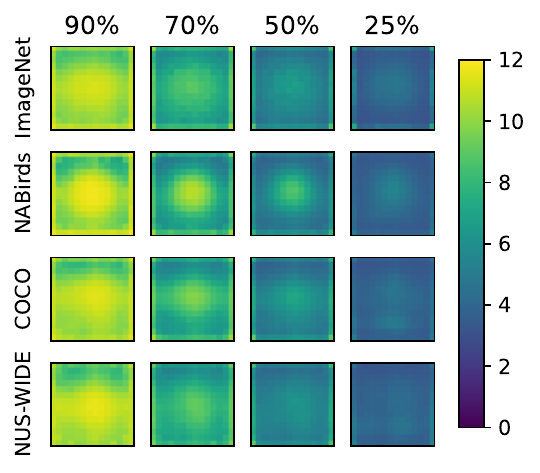}
         \caption{DeiT-B Backbone}
         \label{fig:evitHeatBase}
     \end{subfigure}
        \caption{\textbf{EViT Dataset-averaged reduction patterns (Section~\ref{sec:global}).} We find that the EViT method on average select tokens from both the image center and border.}
        \label{fig:evitHeatGlobal}
\end{figure*}

\begin{figure*}
     \begin{subfigure}[b]{0.33\textwidth}
         \centering
        \includegraphics[width=\linewidth]{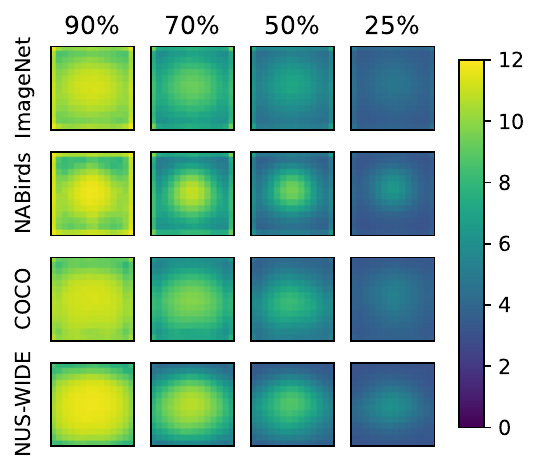}
         \caption{DeiT-T Backbone}
         \label{fig:dyvitHeatTiny}
     \end{subfigure}
     \hfill
     \begin{subfigure}[b]{0.33\textwidth}
         \centering
        \includegraphics[width=\linewidth]{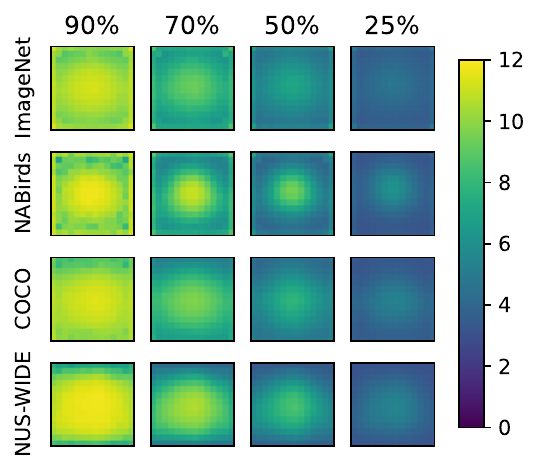}
         \caption{DeiT-S Backbone}
         \label{fig:dyvitHeatSmall}
     \end{subfigure}
     \hfill
     \begin{subfigure}[b]{0.33\textwidth}
         \centering
        \includegraphics[width=\linewidth]{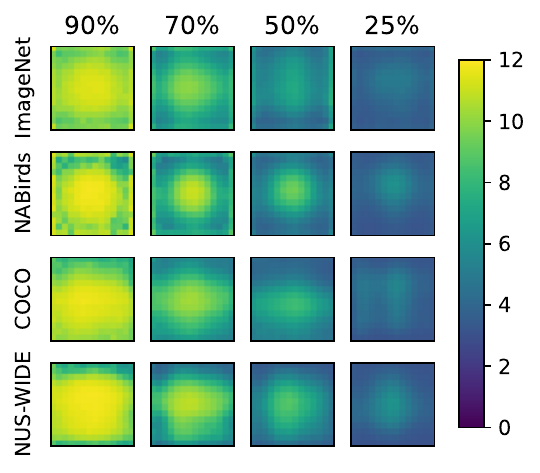}
         \caption{DeiT-B Backbone}
         \label{fig:dyvitHeatBase}
     \end{subfigure}
        \caption{\textbf{DynamicViT Dataset-averaged reduction patterns (Section~\ref{sec:global}).} We find that the DyanmicViT method on average selects tokens primarily from the image center, ignoring the borders.}
        \label{fig:dyvitHeatGlobal}
\end{figure*}

\begin{figure*}
     \begin{subfigure}[b]{0.33\textwidth}
         \centering
        \includegraphics[width=\linewidth]{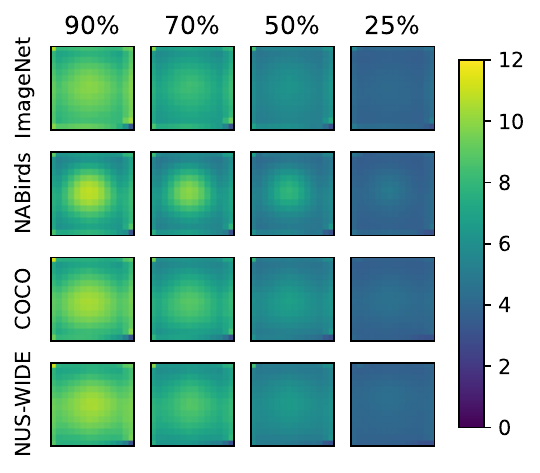}
         \caption{DeiT-T Backbone}
         \label{fig:atsHeatTiny}
     \end{subfigure}
     \hfill
     \begin{subfigure}[b]{0.33\textwidth}
         \centering
        \includegraphics[width=\linewidth]{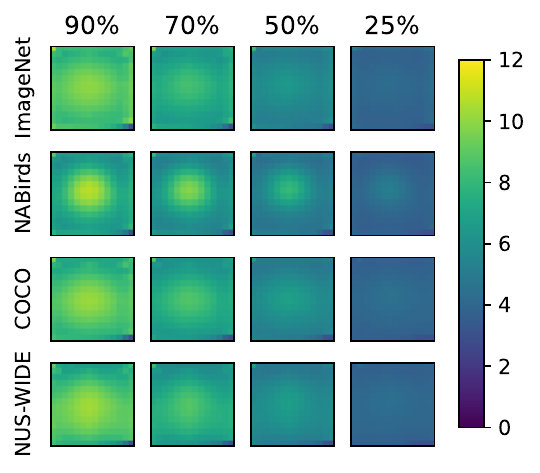}
         \caption{DeiT-S Backbone}
         \label{fig:atsHeatSmall}
     \end{subfigure}
     \hfill
     \begin{subfigure}[b]{0.33\textwidth}
         \centering
        \includegraphics[width=\linewidth]{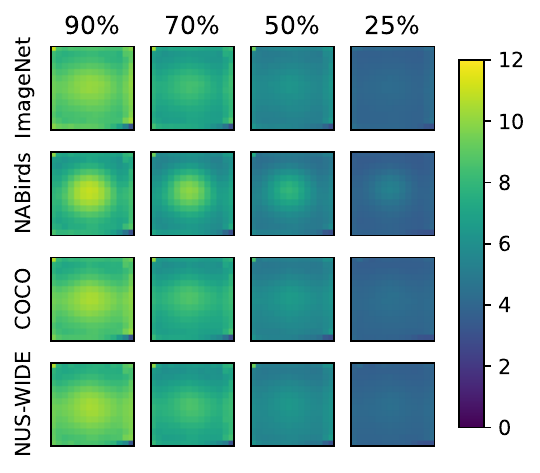}
         \caption{DeiT-B Backbone}
         \label{fig:atsHeatBase}
     \end{subfigure}
        \caption{\textbf{ATS Dataset-averaged reduction patterns (Section~\ref{sec:global}).} We find that the ATS method on average select tokens from both the image center and border, and that the tokens have a lower depth on average due to its dynamic keep rate.}
        \label{fig:atsHeatGlobal}
\end{figure*}

\begin{figure*}
     \begin{subfigure}[b]{0.33\textwidth}
         \centering
        \includegraphics[width=\linewidth]{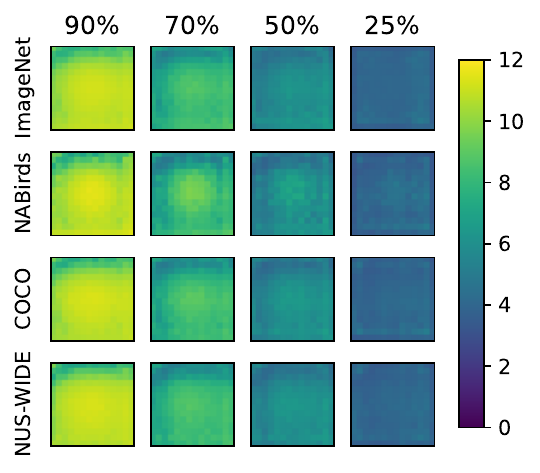}
         \caption{DeiT-T Backbone}
         \label{fig:kmedoidsHeatTiny}
     \end{subfigure}
     \hfill
     \begin{subfigure}[b]{0.33\textwidth}
         \centering
        \includegraphics[width=\linewidth]{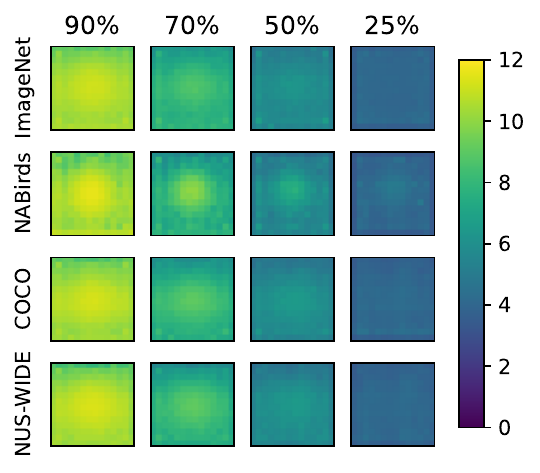}
         \caption{DeiT-S Backbone}
         \label{fig:kmedoidsHeatSmall}
     \end{subfigure}
     \hfill
     \begin{subfigure}[b]{0.33\textwidth}
         \centering
        \includegraphics[width=\linewidth]{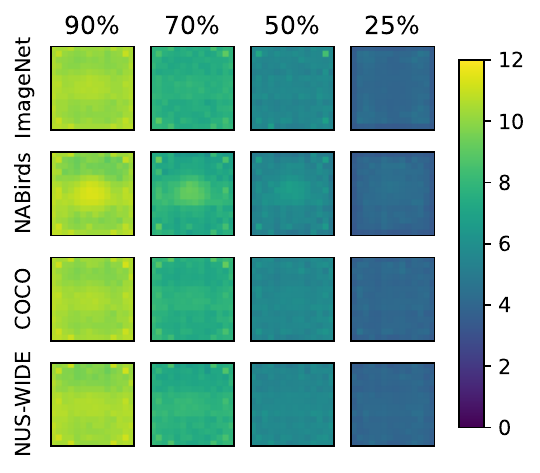}
         \caption{DeiT-B Backbone}
         \label{fig:kmedoidsHeatBase}
     \end{subfigure}
        \caption{\textbf{K-Medoids Dataset-averaged reduction patterns (Section~\ref{sec:global}).} We find that the K-Medoids method select tokens in a more uniform manner than the pruning-based methods.}
        \label{fig:kmedoidsHeatGlobal}
\end{figure*}

\begin{figure*}
     \begin{subfigure}[b]{0.33\textwidth}
         \centering
        \includegraphics[width=\linewidth]{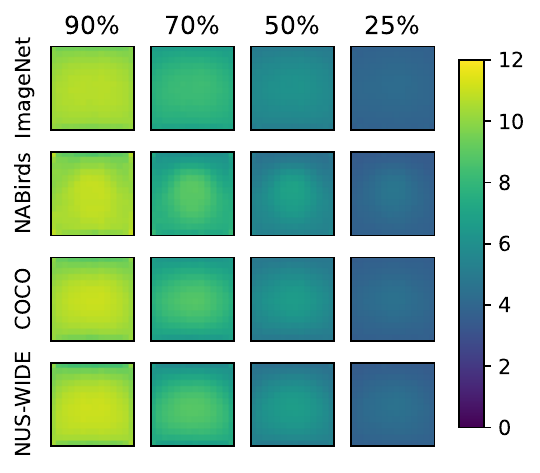}
         \caption{DeiT-T Backbone}
         \label{fig:dpcknnHeatTiny}
     \end{subfigure}
     \hfill
     \begin{subfigure}[b]{0.33\textwidth}
         \centering
        \includegraphics[width=\linewidth]{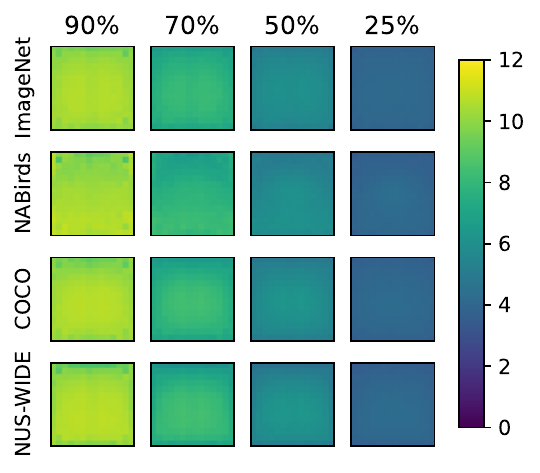}
         \caption{DeiT-S Backbone}
         \label{fig:dpcknnHeatSmall}
     \end{subfigure}
     \hfill
     \begin{subfigure}[b]{0.33\textwidth}
         \centering
        \includegraphics[width=\linewidth]{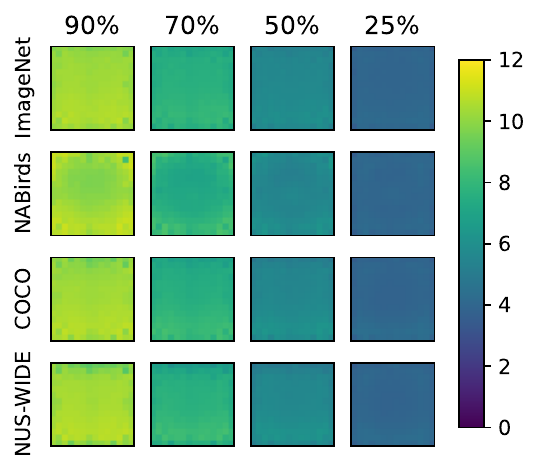}
         \caption{DeiT-B Backbone}
         \label{fig:dpcknnHeatBase}
     \end{subfigure}
        \caption{\textbf{DPC-KNN Dataset-averaged reduction patterns (Section~\ref{sec:global}).} We find that the DPC-KNN method select tokens in a more uniform manner than the pruning-based methods.}
        \label{fig:dpcknnHeatGlobal}
\end{figure*}

In this section, we present visualization of the dataset averaged reduction patterns used in Section~5.3 from the learned pruning-based methods as well as DPC-KNN and K-Medoids; see Figures~\ref{fig:topkHeatGlobal}-\ref{fig:dpcknnHeatGlobal}. We observe that the reduction patterns of the Top-K and EViT methods are visually very similar, which is intuitive given both methods use the same pruning technique. Comparatively, we observe that the DynamicViT tends to more often select tokens closer to the image center, whereas Top-K and EViT instead select tokens along the border. This behavior is also exhibited by the ATS method, where we can also observe that the tokens in general have a lower average depth (\ie the tokens are on average processed by fewer ViT layers) due to the dynamic keep rate nature of the ATS method. For the K-Medoids and DPC-KNN methods we see that the tokens are less centered on the image center. Instead the average depth of the tokens is much more uniform, corresponding to what we can visually observe (see Section~\ref{sec:qualitativeViz}) as well as determined when investigating the reduction pattern consistency.

\section{Per-Dataset Reduction Pattern Visualization of Randomly Selected Samples}\label{sec:qualitativeViz}
\begin{figure*}
     \begin{subfigure}[b]{\textwidth}
         \centering
        \includegraphics[width=0.7\linewidth]{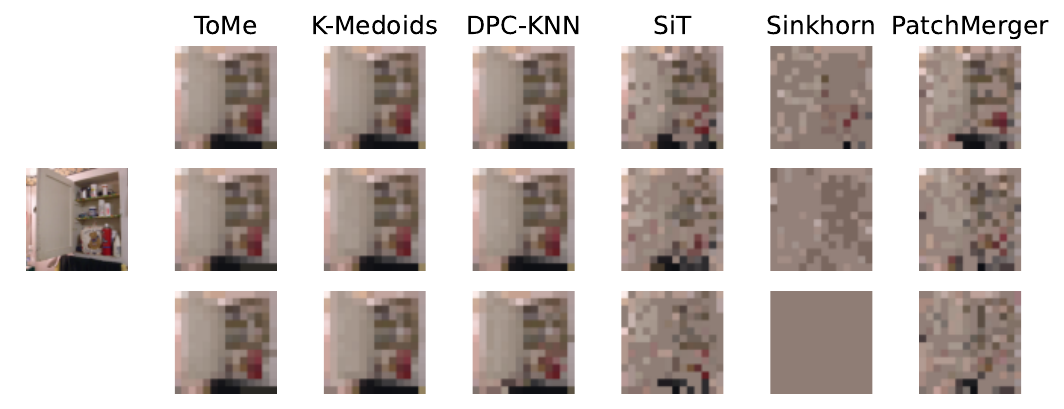}
         \caption{$r=0.90$}
         \label{fig:imQualClusterSmall90}
     \end{subfigure}
     \hfill
     \begin{subfigure}[b]{\textwidth}
         \centering
        \includegraphics[width=0.7\linewidth]{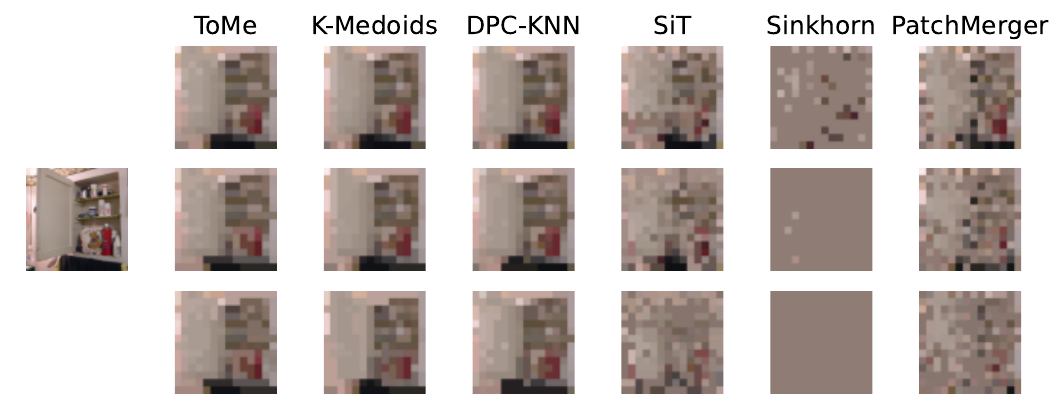}
         \caption{$r=0.70$}
         \label{fig:imQualClusterSmall70}
     \end{subfigure}
     \hfill
     \begin{subfigure}[b]{\textwidth}
         \centering
        \includegraphics[width=0.7\linewidth]{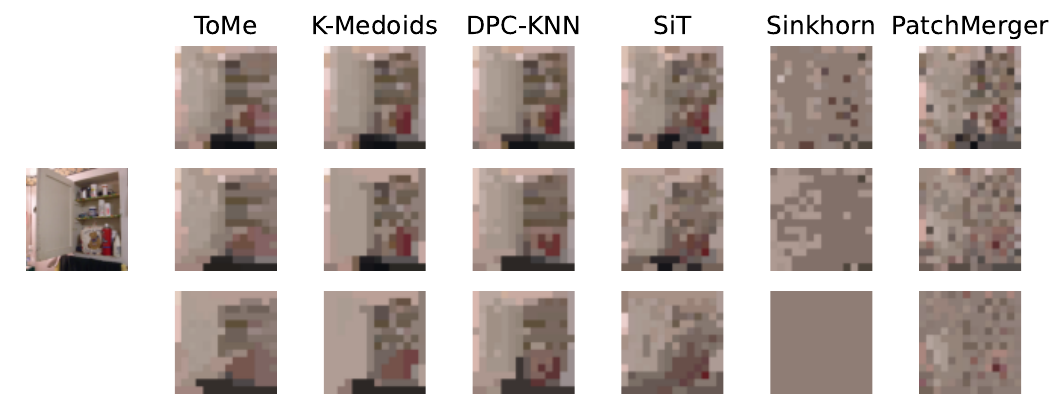}
         \caption{$r=0.50$}
         \label{fig:imQualClusterSmall50}
     \end{subfigure}
     \hfill
     \begin{subfigure}[b]{\textwidth}
         \centering
        \includegraphics[width=0.7\linewidth]{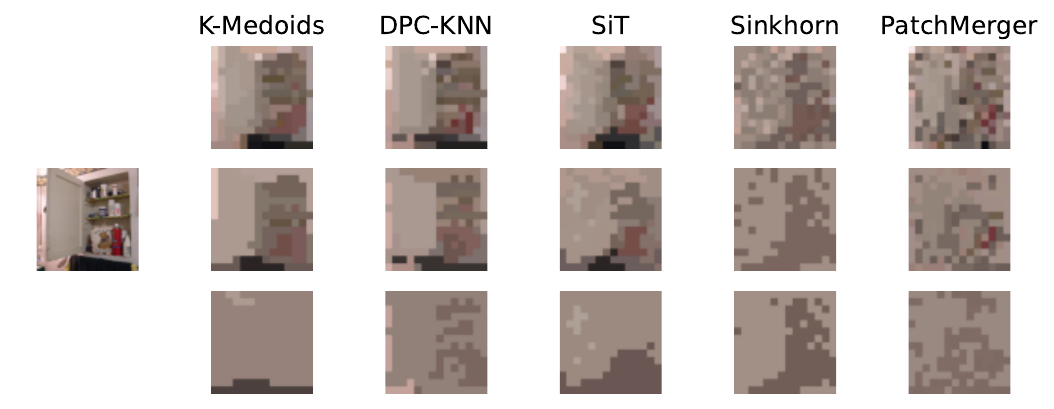}
         \caption{$r=0.25$}
         \label{fig:imQualClusterSmall25}
     \end{subfigure}
        \caption{\textbf{Cluster Reduction Patterns - ImageNet (Section~\ref{sec:qualitativeViz}).} Example of constructed clusters obtained at different keep rate $r$ values, on a random image from the ImageNet dataset.}
        \label{fig:imQualClusterSmall}
\end{figure*}

\begin{figure*}
     \begin{subfigure}[b]{\textwidth}
         \centering
        \includegraphics[width=0.7\linewidth]{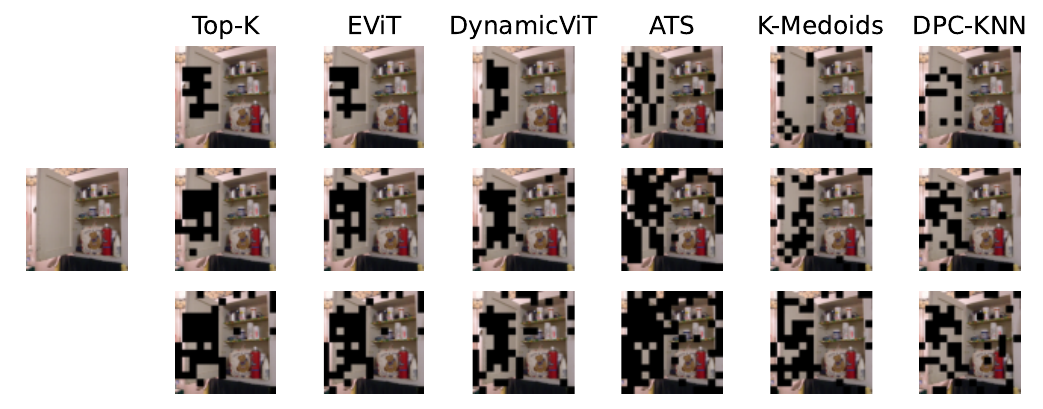}
         \caption{$r=0.90$}
         \label{fig:imQualPruneSmall90}
     \end{subfigure}
     \hfill
     \begin{subfigure}[b]{\textwidth}
         \centering
        \includegraphics[width=0.7\linewidth]{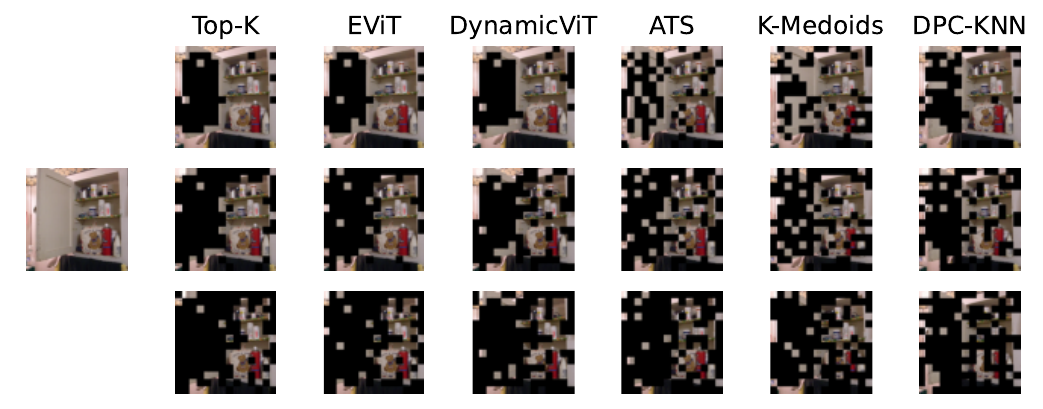}
         \caption{$r=0.70$}
         \label{fig:imQualPruneSmall70}
     \end{subfigure}
     \hfill
     \begin{subfigure}[b]{\textwidth}
         \centering
        \includegraphics[width=0.7\linewidth]{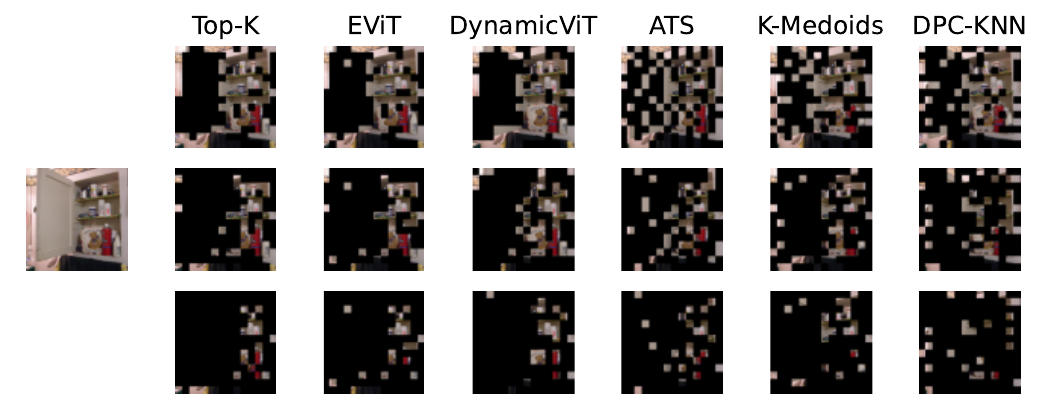}
         \caption{$r=0.50$}
         \label{fig:imQualPruneSmall50}
     \end{subfigure}
     \hfill
     \begin{subfigure}[b]{\textwidth}
         \centering
        \includegraphics[width=0.7\linewidth]{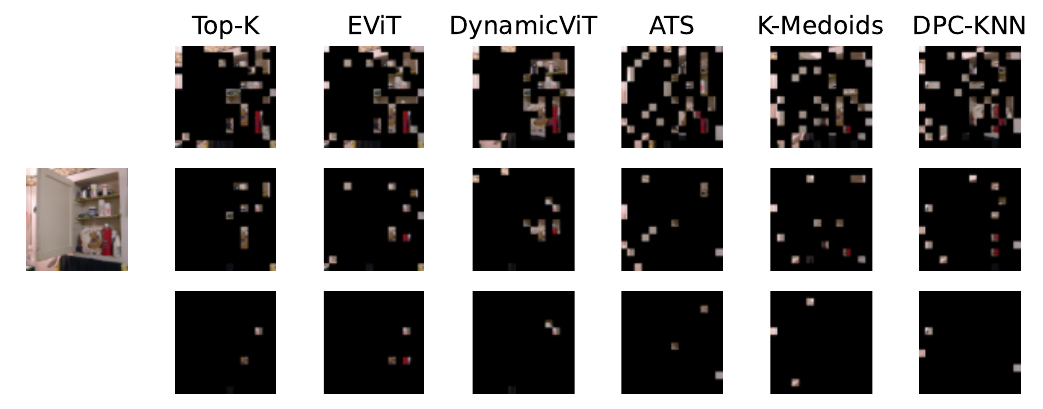}
         \caption{$r=0.25$}
         \label{fig:imQualPruneSmall25}
     \end{subfigure}
        \caption{\textbf{Pruning Reduction Patterns - ImageNet (Section~\ref{sec:qualitativeViz}).} Example of pruning reduction patterns obtained at different keep rate $r$ values, on a random image from the ImageNet dataset.}
        \label{fig:imQualPruneSmall}
\end{figure*}

\begin{figure*}
     \begin{subfigure}[b]{\textwidth}
         \centering
        \includegraphics[width=0.7\linewidth]{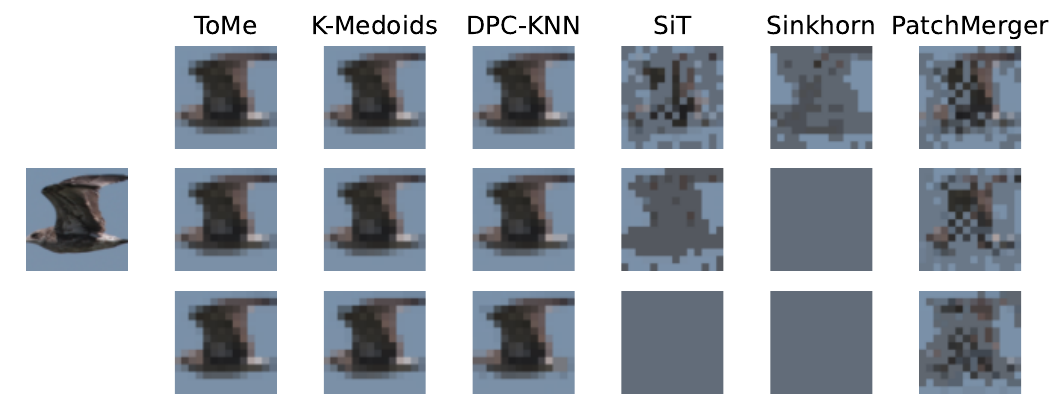}
         \caption{$r=0.90$}
         \label{fig:NABQualClusterSmall90}
     \end{subfigure}
     \hfill
     \begin{subfigure}[b]{\textwidth}
         \centering
        \includegraphics[width=0.7\linewidth]{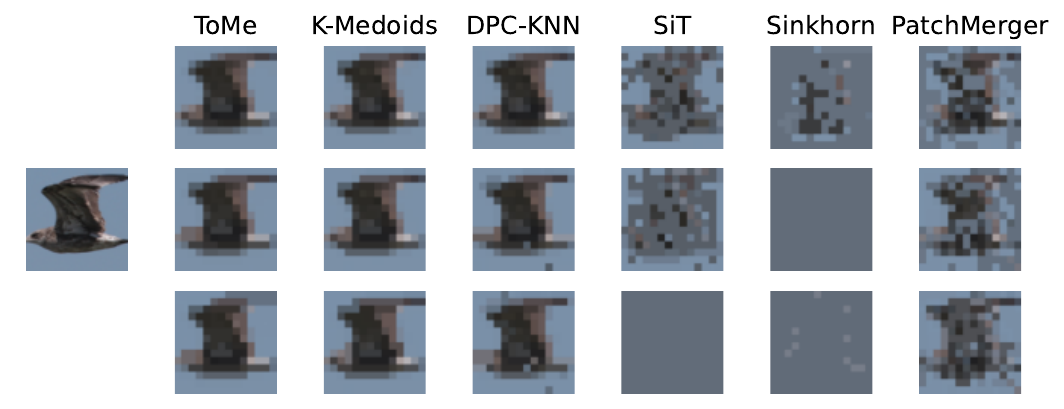}
         \caption{$r=0.70$}
         \label{fig:NABQualClusterSmall70}
     \end{subfigure}
     \hfill
     \begin{subfigure}[b]{\textwidth}
         \centering
        \includegraphics[width=0.7\linewidth]{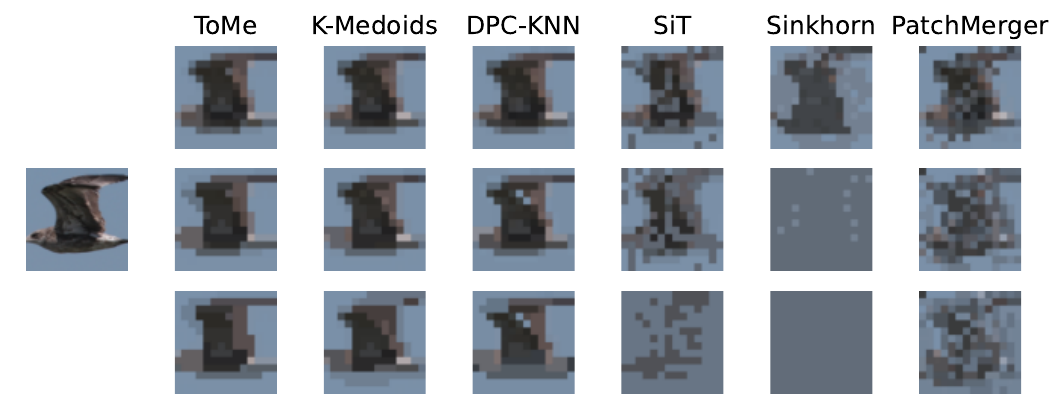}
         \caption{$r=0.50$}
         \label{fig:NABQualClusterSmall50}
     \end{subfigure}
     \hfill
     \begin{subfigure}[b]{\textwidth}
         \centering
        \includegraphics[width=0.7\linewidth]{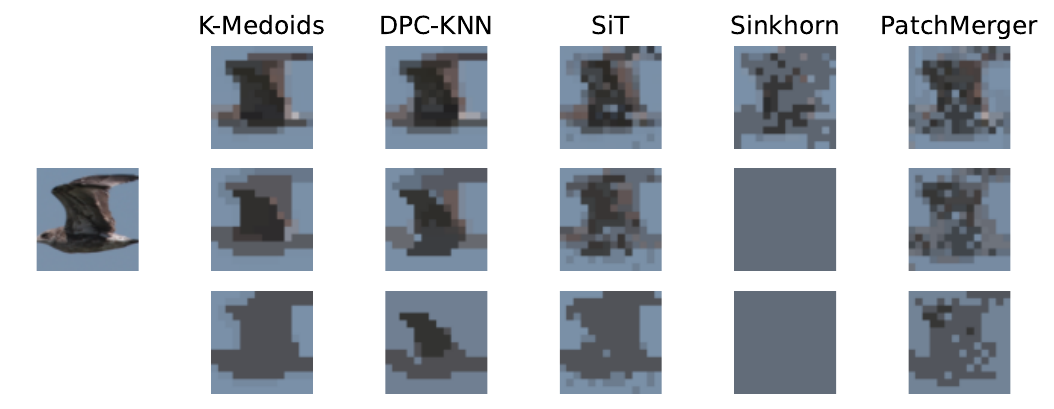}
         \caption{$r=0.25$}
         \label{fig:NABQualClusterSmall25}
     \end{subfigure}
        \caption{\textbf{Cluster Reduction Patterns - NABirds (Section~\ref{sec:qualitativeViz}).} Example of constructed clusters obtained at different keep rate $r$ values, on a random image from the NABirds dataset.}
        \label{fig:NABQualClusterSmall}
\end{figure*}

\begin{figure*}
     \begin{subfigure}[b]{\textwidth}
         \centering
        \includegraphics[width=0.7\linewidth]{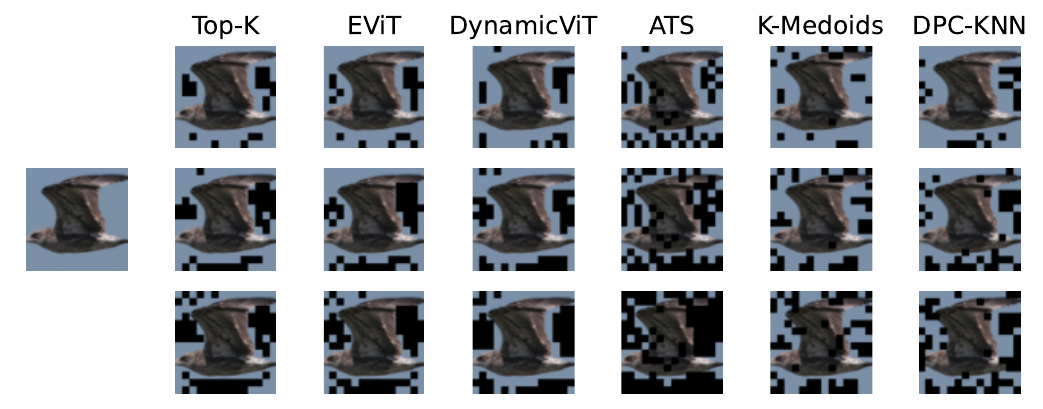}
         \caption{$r=0.90$}
         \label{fig:NABQualPruneSmall90}
     \end{subfigure}
     \hfill
     \begin{subfigure}[b]{\textwidth}
         \centering
        \includegraphics[width=0.7\linewidth]{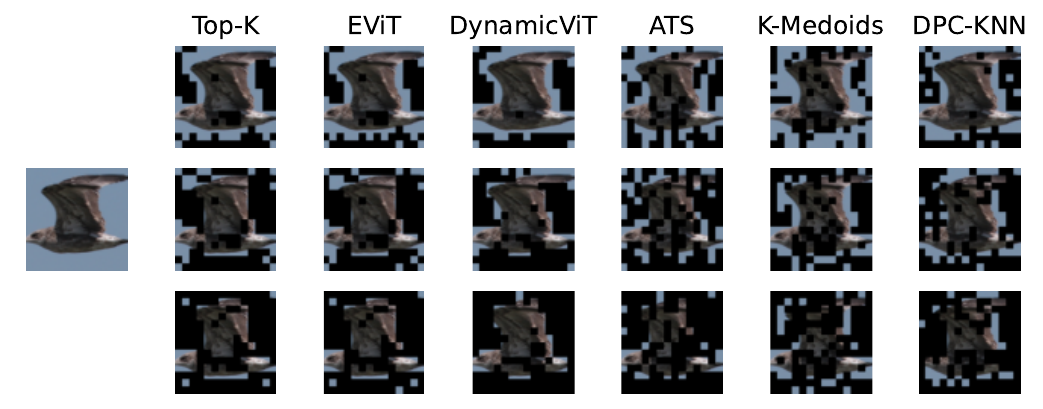}
         \caption{$r=0.70$}
         \label{fig:NABQualPruneSmall70}
     \end{subfigure}
     \hfill
     \begin{subfigure}[b]{\textwidth}
         \centering
        \includegraphics[width=0.7\linewidth]{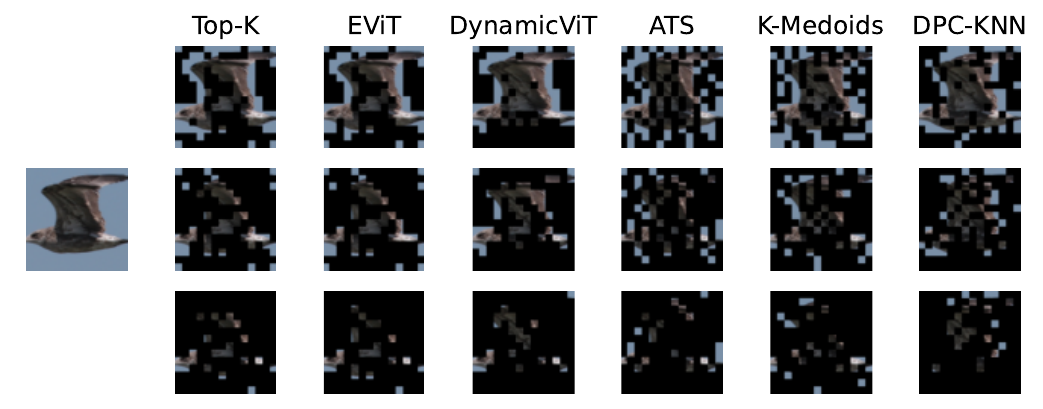}
         \caption{$r=0.50$}
         \label{fig:NABQualPruneSmall50}
     \end{subfigure}
     \hfill
     \begin{subfigure}[b]{\textwidth}
         \centering
        \includegraphics[width=0.7\linewidth]{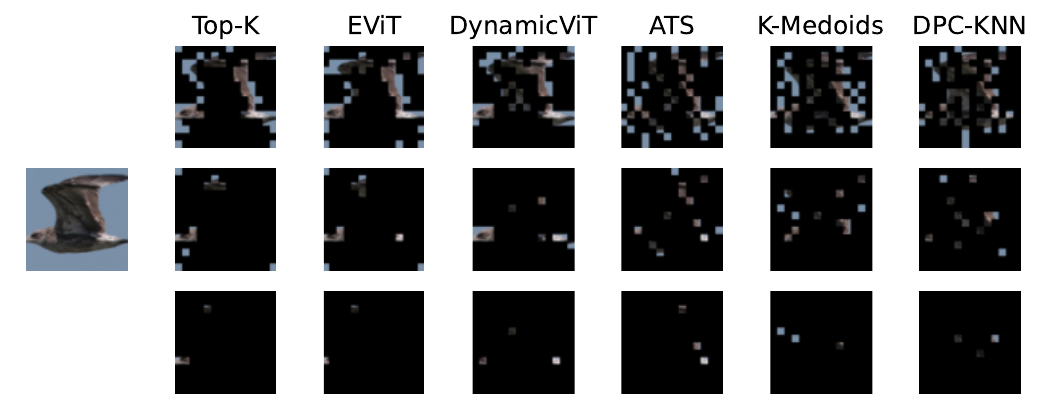}
         \caption{$r=0.25$}
         \label{fig:NABQualPruneSmall25}
     \end{subfigure}
        \caption{\textbf{Pruning Reduction Patterns - NABirds (Section~\ref{sec:qualitativeViz}).} Example of pruning reduction patterns obtained at different keep rate $r$ values, on a random image from the NABirds dataset.}
        \label{fig:NABQualPruneSmall}
\end{figure*}

\begin{figure*}
     \begin{subfigure}[b]{\textwidth}
         \centering
        \includegraphics[width=0.7\linewidth]{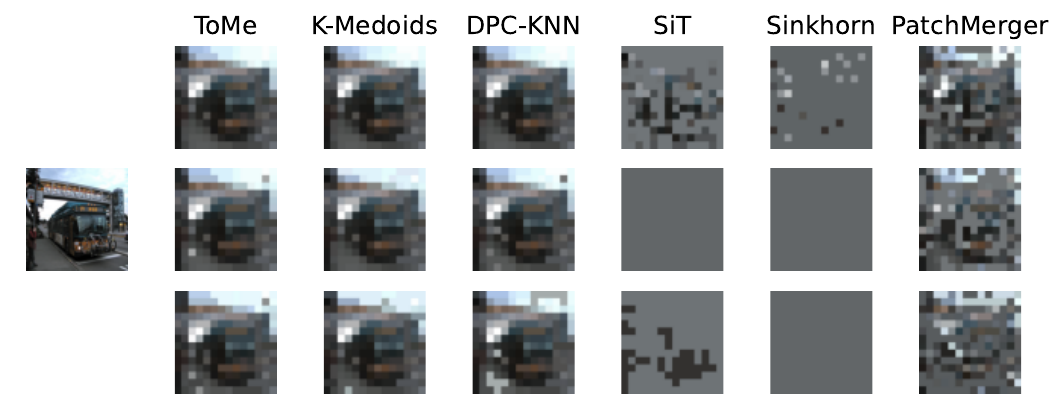}
         \caption{$r=0.90$}
         \label{fig:COCOQualClusterSmall90}
     \end{subfigure}
     \hfill
     \begin{subfigure}[b]{\textwidth}
         \centering
        \includegraphics[width=0.7\linewidth]{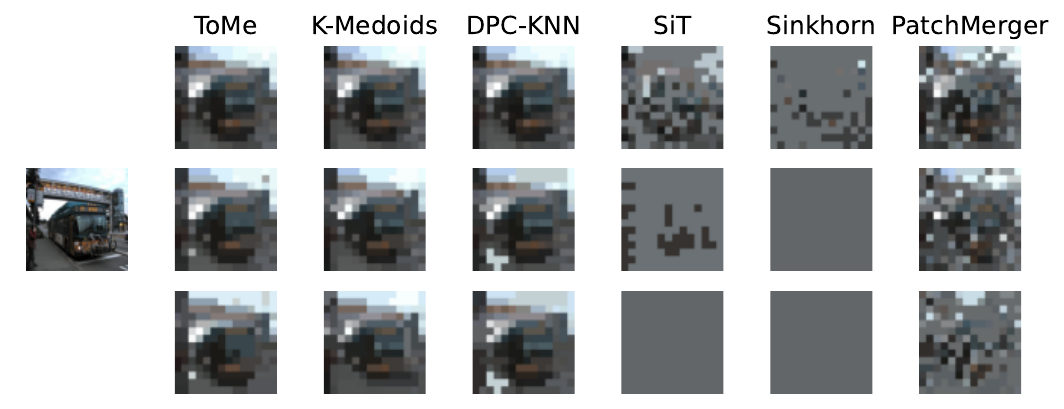}
         \caption{$r=0.70$}
         \label{fig:COCOQualClusterSmall70}
     \end{subfigure}
     \hfill
     \begin{subfigure}[b]{\textwidth}
         \centering
        \includegraphics[width=0.7\linewidth]{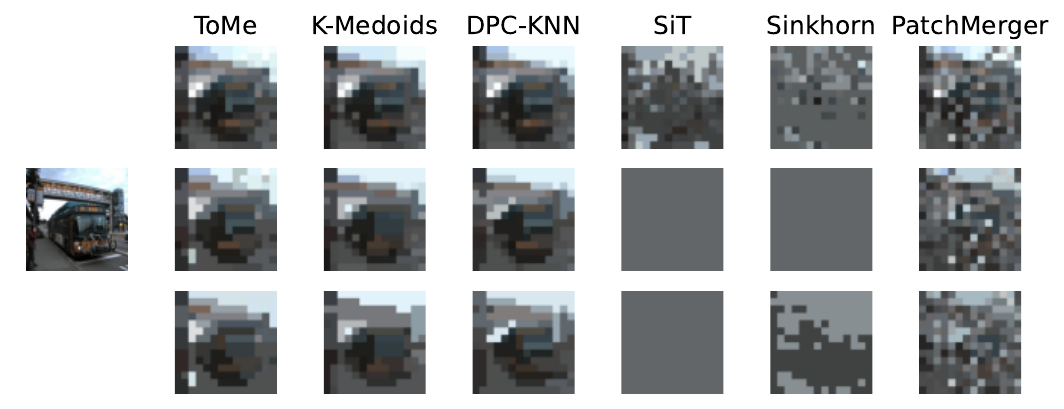}
         \caption{$r=0.50$}
         \label{fig:COCOQualClusterSmall50}
     \end{subfigure}
     \hfill
     \begin{subfigure}[b]{\textwidth}
         \centering
        \includegraphics[width=0.7\linewidth]{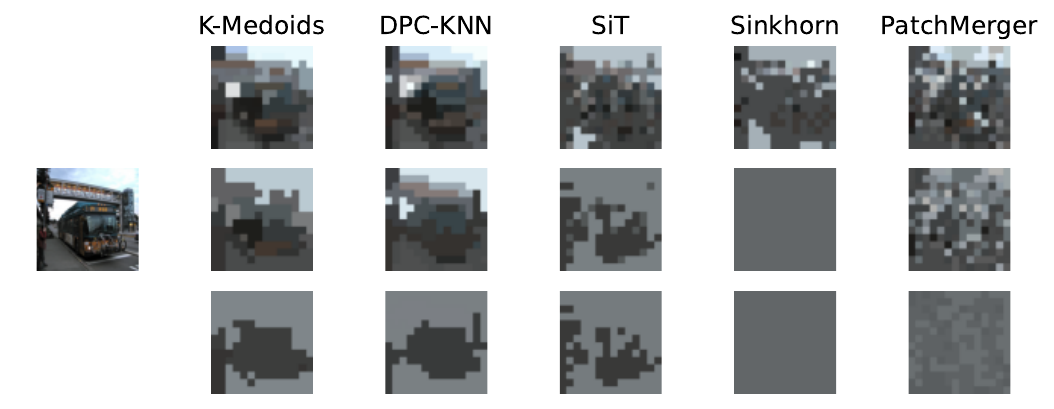}
         \caption{$r=0.25$}
         \label{fig:COCOQualClusterSmall25}
     \end{subfigure}
        \caption{\textbf{Cluster Reduction Patterns - COCO (Section~\ref{sec:qualitativeViz}).} Example of constructed clusters obtained at different keep rate $r$ values, on a random image from the COCO dataset.}
        \label{fig:COCOQualClusterSmall}
\end{figure*}

\begin{figure*}
     \begin{subfigure}[b]{\textwidth}
         \centering
        \includegraphics[width=0.7\linewidth]{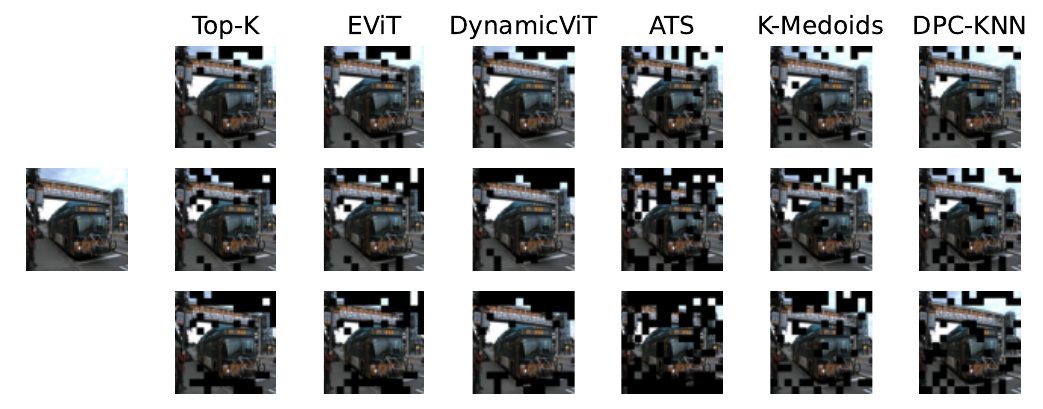}
         \caption{$r=0.90$}
         \label{fig:COCOQualPruneSmall90}
     \end{subfigure}
     \hfill
     \begin{subfigure}[b]{\textwidth}
         \centering
        \includegraphics[width=0.7\linewidth]{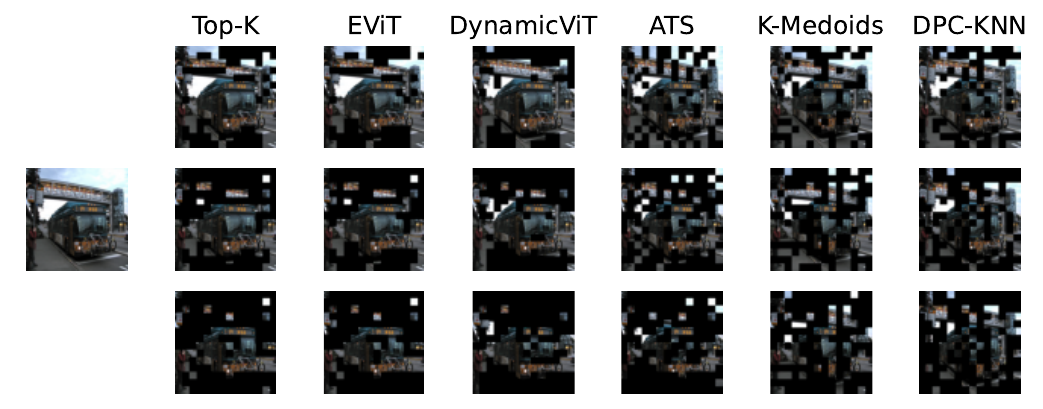}
         \caption{$r=0.70$}
         \label{fig:COCOQualPruneSmall70}
     \end{subfigure}
     \hfill
     \begin{subfigure}[b]{\textwidth}
         \centering
        \includegraphics[width=0.7\linewidth]{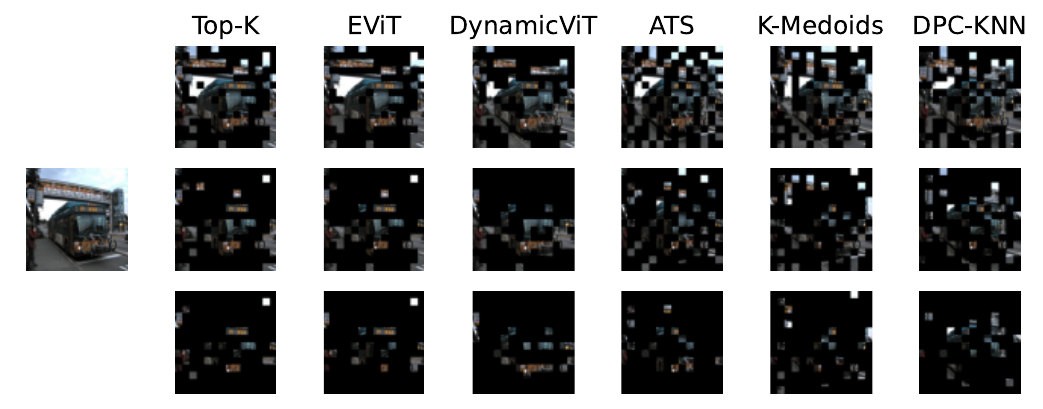}
         \caption{$r=0.50$}
         \label{fig:COCOQualPruneSmall50}
     \end{subfigure}
     \hfill
     \begin{subfigure}[b]{\textwidth}
         \centering
        \includegraphics[width=0.7\linewidth]{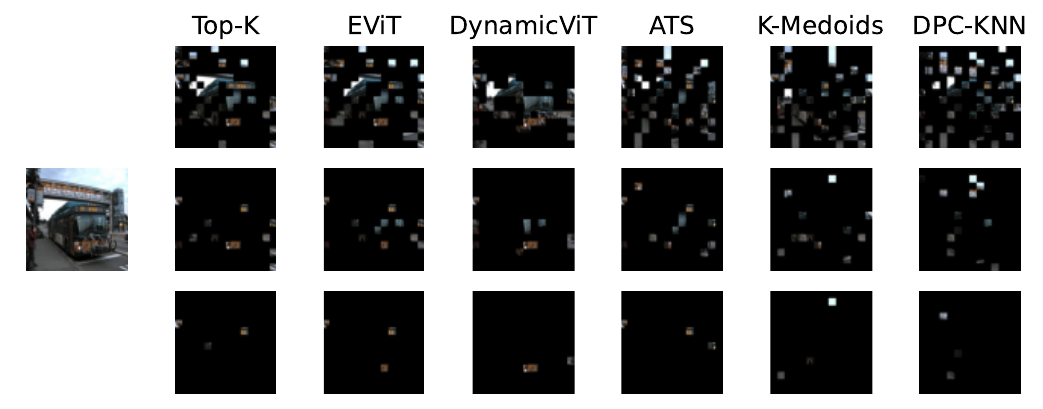}
         \caption{$r=0.25$}
         \label{fig:COCOQualPruneSmall25}
     \end{subfigure}
        \caption{\textbf{Pruning Reduction Patterns - COCO (Section~\ref{sec:qualitativeViz}).} Example of pruning reduction patterns obtained at different keep rate $r$ values, on a random image from the COCO dataset.}
        \label{fig:COCOQualPruneSmall}
\end{figure*}

\begin{figure*}
     \begin{subfigure}[b]{\textwidth}
         \centering
        \includegraphics[width=0.7\linewidth]{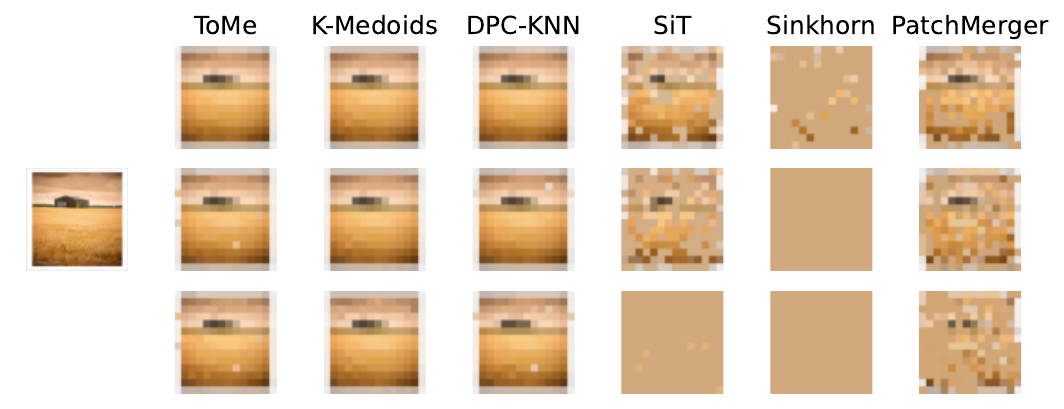}
         \caption{$r=0.90$}
         \label{fig:NUSQualClusterSmall90}
     \end{subfigure}
     \hfill
     \begin{subfigure}[b]{\textwidth}
         \centering
        \includegraphics[width=0.7\linewidth]{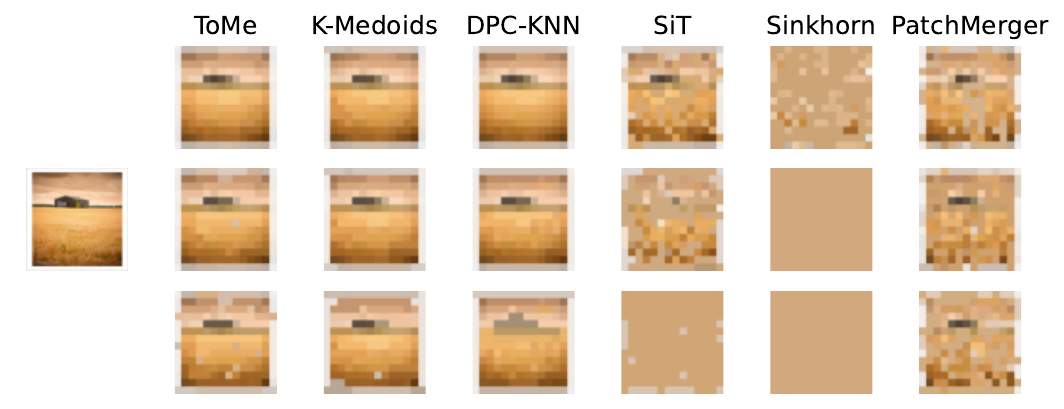}
         \caption{$r=0.70$}
         \label{fig:NUSQualClusterSmall70}
     \end{subfigure}
     \hfill
     \begin{subfigure}[b]{\textwidth}
         \centering
        \includegraphics[width=0.7\linewidth]{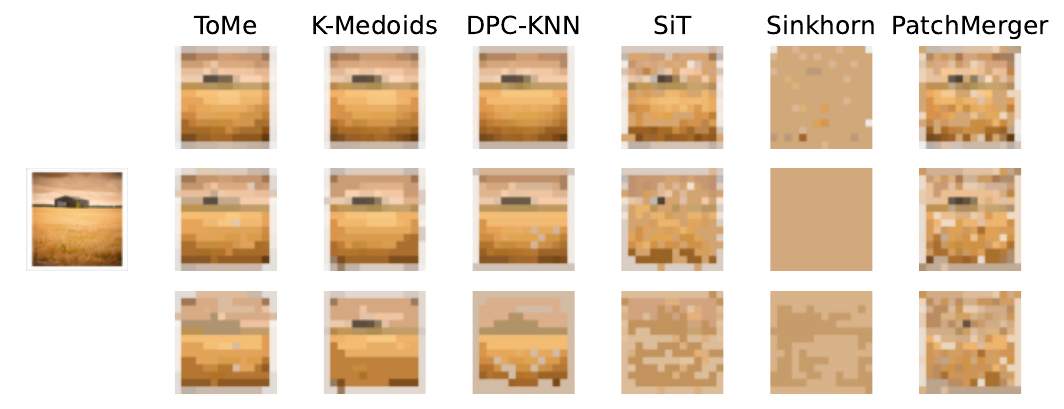}
         \caption{$r=0.50$}
         \label{fig:NUSQualClusterSmall50}
     \end{subfigure}
     \hfill
     \begin{subfigure}[b]{\textwidth}
         \centering
        \includegraphics[width=0.7\linewidth]{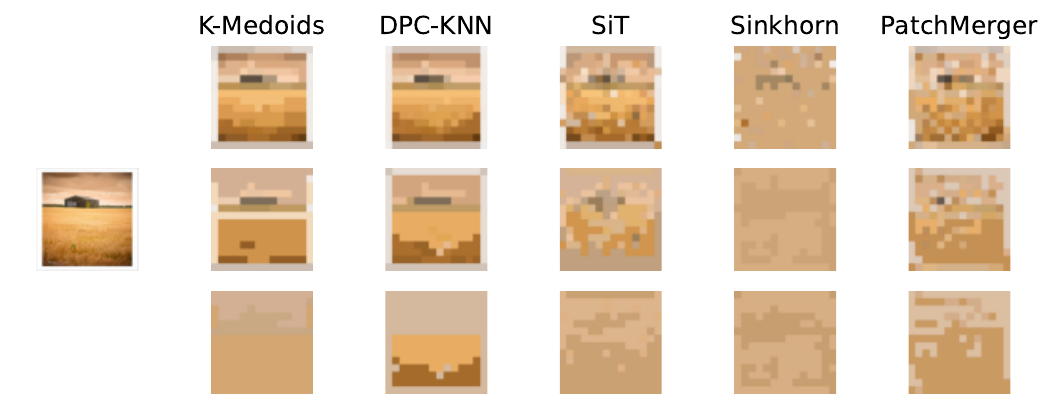}
         \caption{$r=0.25$}
         \label{fig:NUSQualClusterSmall25}
     \end{subfigure}
        \caption{\textbf{Cluster Reduction Patterns - NUS-WIDE (Section~\ref{sec:qualitativeViz}).} Example of constructed clusters obtained at different keep rate $r$ values, on a random image from the NUS-WIDE dataset.}
        \label{fig:NUSQualClusterSmall}
\end{figure*}

\begin{figure*}
     \begin{subfigure}[b]{\textwidth}
         \centering
        \includegraphics[width=0.7\linewidth]{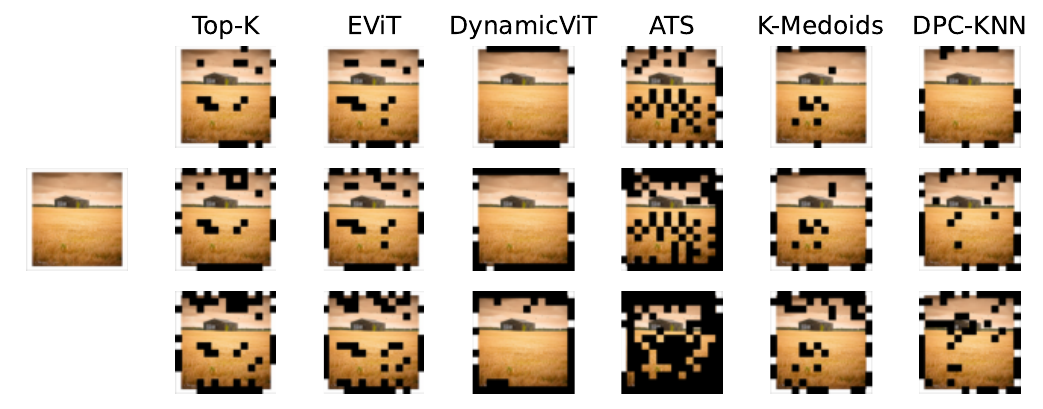}
         \caption{$r=0.90$}
         \label{fig:NUSQualPruneSmall90}
     \end{subfigure}
     \hfill
     \begin{subfigure}[b]{\textwidth}
         \centering
        \includegraphics[width=0.7\linewidth]{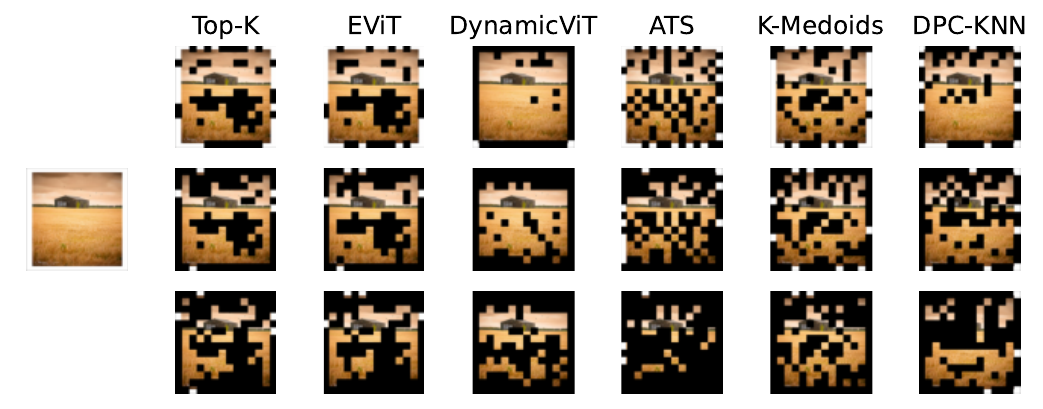}
         \caption{$r=0.70$}
         \label{fig:NUSQualPruneSmall70}
     \end{subfigure}
     \hfill
     \begin{subfigure}[b]{\textwidth}
         \centering
        \includegraphics[width=0.7\linewidth]{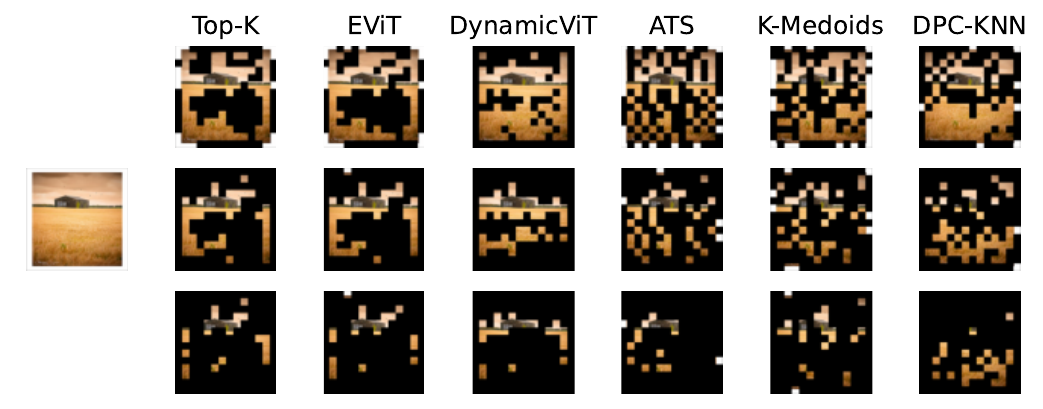}
         \caption{$r=0.50$}
         \label{fig:NUSQualPruneSmall50}
     \end{subfigure}
     \hfill
     \begin{subfigure}[b]{\textwidth}
         \centering
        \includegraphics[width=0.7\linewidth]{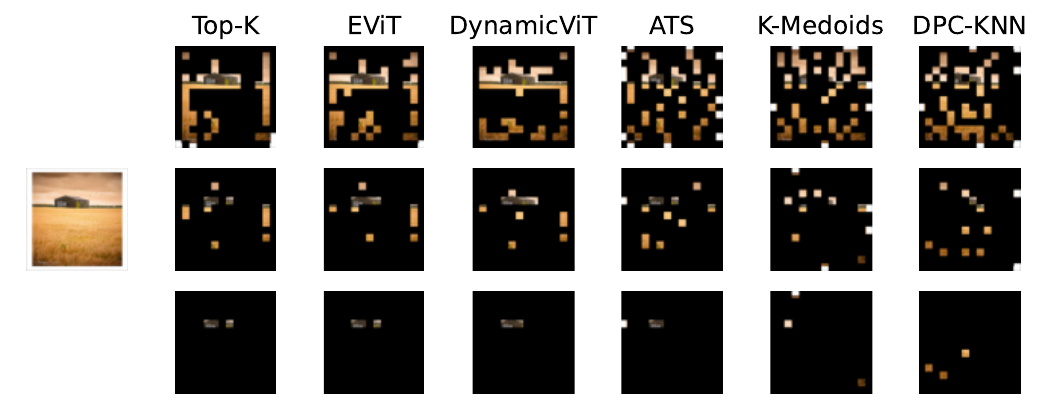}
         \caption{$r=0.25$}
         \label{fig:NUSQualPruneSmall25}
     \end{subfigure}
        \caption{\textbf{Pruning Reduction Patterns - NUS-WIDE (Section~\ref{sec:qualitativeViz}).} Example of pruning reduction patterns obtained at different keep rate $r$ values, on a random image from the NUS-WIDE dataset.}
        \label{fig:NUSQualPruneSmall}
\end{figure*}

As mentioned in Section~5, we present the reduction patterns obtained from a random sample of each dataset; see Figures~\ref{fig:imQualClusterSmall}-\ref{fig:NUSQualPruneSmall}. For brevity, we only show the reduction patterns obtained using the DeiT-S backbone. For the merging-based token reduction methods we observe that the hard-merging methods extract clusters which appear to be semantically coherent, whereas the soft-merging based methods more often extract clusters that are either incoherent or collapsed to a single cluster.

For the pruning-based methods we observe that the Top-K, EViT, and DynamicViT methods manage to focus in on the distinguishing features in a similar manner. Comparatively, the reduction patterns of the ATS method are distinctively different as they maintain a more global diversity of tokens, instead of larger coherent regions. Lastly, we observe the K-Medoids and DPC-KNN methods tend to not select tokens with distinctive features as cluster centers. This makes sense as these tokens may not be the best representative of a larger region.

\clearpage 
{\small
\clearpage
\bibliographystyle{ieee_fullname}
\bibliography{egbib}

\begin{thebibliography}{10}\itemsep=-1pt

\bibitem{VATT_2021}
Hassan Akbari, Liangzhe Yuan, Rui Qian, Wei-Hong Chuang, Shih-Fu Chang, Yin
  Cui, and Boqing Gong.
\newblock {VATT}: Transformers for multimodal self-supervised learning from raw
  video, audio and text.
\newblock In A. Beygelzimer, Y. Dauphin, P. Liang, and J.~Wortman Vaughan,
  editors, {\em Advances in Neural Information Processing Systems}, 2021.

\bibitem{Becker}
Hila Becker.
\newblock Identification and characterization of events in social media.
\newblock 2011.

\bibitem{MultiGrain}
Maxim Berman, Hervé Jégou, Andrea Vedaldi, Iasonas Kokkinos, and Matthijs
  Douze.
\newblock Multigrain: a unified image embedding for classes and instances.
\newblock {\em arXiv preprint arXiv:1902.05509}, 2019.

\bibitem{FlexiViT_2022}
Lucas Beyer, Pavel Izmailov, Alexander Kolesnikov, Mathilde Caron, Simon
  Kornblith, Xiaohua Zhai, Matthias Minderer, Michael Tschannen, Ibrahim
  Alabdulmohsin, and Filip Pavetic.
\newblock Flexivit: One model for all patch sizes.
\newblock {\em arXiv preprint arXiv:2212.08013}, 2022.

\bibitem{ToMe_2023}
Daniel Bolya, Cheng-Yang Fu, Xiaoliang Dai, Peizhao Zhang, Christoph
  Feichtenhofer, and Judy Hoffman.
\newblock Token merging: Your vit but faster.
\newblock In {\em International Conference on Learning Representations}, 2023.

\bibitem{SaliencyMetrics}
Zoya Bylinskii, Tilke Judd, Aude Oliva, Antonio Torralba, and Frédo Durand.
\newblock What do different evaluation metrics tell us about saliency models?
\newblock {\em IEEE Transactions on Pattern Analysis and Machine Intelligence},
  41(3), 2019.

\bibitem{ViTSlim_2022}
Arnav Chavan, Zhiqiang Shen, Zhuang Liu, Zechun Liu, Kwang-Ting Cheng, and Eric
  Xing.
\newblock Vision transformer slimming: Multi-dimension searching in continuous
  optimization space.
\newblock In {\em Proceedings of the IEEE Conference on Computer Vision and
  Pattern Recognition (CVPR)}, 2022.

\bibitem{CFVIT_2022}
Mengzhao Chen, Mingbao Lin, Ke Li, Yunhang Shen, Yongjian Wu, Fei Chao, and
  Rongrong~Ji Cf-vit.
\newblock A general coarse-to-fine method for vision transformer.
\newblock {\em arXiv preprint arXiv:2203.03821}, 2022.

\bibitem{SViTE_2021}
Tianlong Chen, Yu Cheng, Zhe Gan, Lu Yuan, Lei Zhang, and Zhangyang Wang.
\newblock Chasing sparsity in vision transformers: An end-to-end exploration.
\newblock In M. Ranzato, A. Beygelzimer, Y. Dauphin, P.S. Liang, and J.~Wortman
  Vaughan, editors, {\em Advances in Neural Information Processing Systems},
  volume~34, pages 19974--19988. Curran Associates, Inc., 2021.

\bibitem{NUS_WIDE}
Tat-Seng Chua, Jinhui Tang, Richang Hong, Haojie Li, Zhiping Luo, and Yan-Tao
  Zheng.
\newblock Nus-wide: A real-world web image database from national university of
  singapore.
\newblock In {\em Proc. of ACM Conf. on Image and Video Retrieval (CIVR'09)},
  Santorini, Greece., July 8-10, 2009.

\bibitem{RandAugment}
Ekin~Dogus Cubuk, Barret Zoph, Jon Shlens, and Quoc Le.
\newblock Randaugment: Practical automated data augmentation with a reduced
  search space.
\newblock In H. Larochelle, M. Ranzato, R. Hadsell, M.F. Balcan, and H. Lin,
  editors, {\em Advances in Neural Information Processing Systems}, volume~33,
  pages 18613--18624. Curran Associates, Inc., 2020.

\bibitem{efficiency_2022}
Mostafa Dehghani, Yi Tay, Anurag Arnab, Lucas Beyer, and Ashish Vaswani.
\newblock The efficiency misnomer.
\newblock In {\em International Conference on Learning Representations}, 2022.

\bibitem{CutOut}
Terrance DeVries and Graham~W. Taylor.
\newblock Improved regularization of convolutional neural networks with cutout.
\newblock {\em arXiv preprint arXiv:1708.04552}, 2017.

\bibitem{Procrustes_2021}
Frances Ding, Jean-Stanislas Denain, and Jacob Steinhardt.
\newblock Grounding representation similarity through statistical testing.
\newblock In A. Beygelzimer, Y. Dauphin, P. Liang, and J.~Wortman Vaughan,
  editors, {\em Advances in Neural Information Processing Systems}, 2021.

\bibitem{ViT_2021}
Alexey Dosovitskiy, Lucas Beyer, Alexander Kolesnikov, Dirk Weissenborn,
  Xiaohua Zhai, Thomas Unterthiner, Mostafa Dehghani, Matthias Minderer, Georg
  Heigold, Sylvain Gelly, Jakob Uszkoreit, and Neil Houlsby.
\newblock An image is worth 16x16 words: Transformers for image recognition at
  scale.
\newblock In {\em International Conference on Learning Representations}, 2021.

\bibitem{ToRe_2022}
Zhiyang Dou, Qingxuan Wu, Cheng Lin, Zeyu Cao, Qiangqiang Wu, Weilin Wan, Taku
  Komura, and Wenping Wang.
\newblock Tore: Token reduction for efficient human mesh recovery with
  transformer.
\newblock {\em arXiv preprint arXiv:2211.10705}, 2022.

\bibitem{DPCKNN_2016}
Mingjing Du, Shifei Ding, and Hongjie Jia.
\newblock Study on density peaks clustering based on k-nearest neighbors and
  principal component analysis.
\newblock {\em Knowledge-Based Systems}, 99:135--145, 2016.

\bibitem{ATS_2022}
Mohsen Fayyaz, Soroush~Abbasi Koohpayegani, Farnoush~Rezaei Jafari, Sunando
  Sengupta, Hamid Reza~Vaezi Joze, Eric Sommerlade, Hamed Pirsiavash, and
  Juergen Gall.
\newblock Adaptive token sampling for efficient vision transformers.
\newblock In {\em Proceedings of the European Conference on Computer Vision
  (ECCV)}, 2022.

\bibitem{LRNormFactor}
Priya Goyal, Piotr Dollár, Ross Girshick, Pieter Noordhuis, Lukasz Wesolowski,
  Aapo Kyrola, Andrew Tulloch, Yangqing Jia, and Kaiming He.
\newblock Accurate, large minibatch sgd: Training imagenet in 1 hour.
\newblock {\em arXiv preprint arXiv:1706.02677}, 2017.

\bibitem{LeViT_2021}
Benjamin Graham, Alaaeldin El-Nouby, Hugo Touvron, Pierre Stock, Armand Joulin,
  Herve Jegou, and Matthijs Douze.
\newblock Levit: A vision transformer in convnet's clothing for faster
  inference.
\newblock In {\em Proceedings of the IEEE/CVF International Conference on
  Computer Vision (ICCV)}, October 2021.

\bibitem{Sinkhorn_2022}
Joakim~Bruslund Haurum, Meysam Madadi, Sergio Escalera, and Thomas~B. Moeslund.
\newblock Multi-scale hybrid vision transformer and sinkhorn tokenizer for
  sewer defect classification.
\newblock {\em Automation in Construction}, 144:104614, 2022.

\bibitem{TransFG_2022}
Ju He, Jie-Neng Chen, Shuai Liu, Adam Kortylewski, Cheng Yang, Yutong Bai, and
  Changhu Wang.
\newblock Transfg: A transformer architecture for fine-grained recognition.
\newblock {\em Proceedings of the AAAI Conference on Artificial Intelligence},
  36(1):852--860, Jun. 2022.

\bibitem{PiT_2021}
Byeongho Heo, Sangdoo Yun, Dongyoon Han, Sanghyuk Chun, Junsuk Choe, and
  Seong~Joon Oh.
\newblock Rethinking spatial dimensions of vision transformers.
\newblock In {\em Proceedings of the International Conference on Computer
  Vision (ICCV)}, 2021.

\bibitem{hinkle2003applied}
Dennis~E Hinkle, William Wiersma, and Stephen~G Jurs.
\newblock {\em Applied statistics for the behavioral sciences}, volume 663.
\newblock Houghton Mifflin college division, 2003.

\bibitem{RepeatedAug}
Elad Hoffer, Tal Ben-Nun, Itay Hubara, Niv Giladi, Torsten Hoefler, and Daniel
  Soudry.
\newblock Augment your batch: Improving generalization through instance
  repetition.
\newblock In {\em Proceedings of the IEEE/CVF Conference on Computer Vision and
  Pattern Recognition (CVPR)}, pages 8126--8135, 2020.

\bibitem{HAS_2022}
Chao Hu, Liqiang Zhu, Weibin Qiu, and Weijie Wu.
\newblock Data augmentation vision transformer for fine-grained image
  classification.
\newblock {\em arXiv preprint arXiv:2211.12879}, 2022.

\bibitem{RAMSTrans_2021}
Yunqing Hu, Xuan Jin, Yin Zhang, Haiwen Hong, Jingfeng Zhang, Yuan He, and Hui
  Xue.
\newblock Rams-trans: Recurrent attention multi-scale transformer
  forfine-grained image recognition.
\newblock {\em arXiv preprint arXiv:2107.08192}, 2021.

\bibitem{StochasticDepth}
Gao Huang, Yu Sun, Zhuang Liu, Daniel Sedra, and Kilian~Q. Weinberger.
\newblock Deep networks with stochastic depth.
\newblock In Bastian Leibe, Jiri Matas, Nicu Sebe, and Max Welling, editors,
  {\em Proceedings of the European Conference on Computer Vision (ECCV)}, pages
  646--661, Cham, 2016. Springer International Publishing.

\bibitem{GumbelSoftmax}
Eric Jang, Shixiang Gu, and Ben Poole.
\newblock Categorical reparameterization with gumbel-softmax.
\newblock In {\em International Conference on Learning Representations}, 2017.

\bibitem{SPViT_2022}
Zhenglun Kong, Peiyan Dong, Xiaolong Ma, Xin Meng, Wei Niu, Mengshu Sun, Xuan
  Shen, Geng Yuan, Bin Ren, Hao Tang, Minghai Qin, and Yanzhi Wang.
\newblock Spvit: Enabling faster vision transformers via latency-aware soft
  token pruning.
\newblock In {\em Proceedings of the European Conference on Computer Vision
  (ECCV)}, 2022.

\bibitem{CKA_2019}
Simon Kornblith, Mohammad Norouzi, Honglak Lee, and Geoffrey Hinton.
\newblock Similarity of neural network representations revisited.
\newblock In Kamalika Chaudhuri and Ruslan Salakhutdinov, editors, {\em
  Proceedings of the 36th International Conference on Machine Learning},
  volume~97 of {\em Proceedings of Machine Learning Research}, pages
  3519--3529. PMLR, 09--15 Jun 2019.

\bibitem{VisualGenome}
Ranjay Krishna, Yuke Zhu, Oliver Groth, Justin Johnson, Kenji Hata, Joshua
  Kravitz, Stephanie Chen, Yannis Kalantidis, Li-Jia Li, David~A. Shamma,
  Michael~S. Bernstein, and Li Fei-Fei.
\newblock Visual genome: Connecting language and vision using crowdsourced
  dense image annotations.
\newblock {\em International Journal of Computer Vision}, 123(1):32--73, Feb.
  2017.

\bibitem{OpenImages}
Alina Kuznetsova, Hassan Rom, Neil Alldrin, Jasper Uijlings, Ivan Krasin, Jordi
  Pont-Tuset, Shahab Kamali, Stefan Popov, Matteo Malloci, Alexander
  Kolesnikov, Tom Duerig, and Vittorio Ferrari.
\newblock The open images dataset v4.
\newblock {\em International Journal of Computer Vision}, 128(7):1956--1981,
  Mar. 2020.

\bibitem{SaiT_2022}
Ling Li, David Thorsley, and Joseph Hassoun.
\newblock Sait: Sparse vision transformers through adaptive token pruning.
\newblock {\em arXiv preprint arXiv:2210.05832}, 2022.

\bibitem{QVIT_2022}
Zhexin Li, Tong Yang, Peisong Wang, and Jian Cheng.
\newblock Q-vit: Fully differentiable quantization for vision transformer.
\newblock {\em arXiv preprint arXiv:2201.07703}, 2022.

\bibitem{EViT_2022}
Youwei Liang, Chongjian GE, Zhan Tong, Yibing Song, Jue Wang, and Pengtao Xie.
\newblock {EV}it: Expediting vision transformers via token reorganizations.
\newblock In {\em International Conference on Learning Representations}, 2022.

\bibitem{SuperViT_2022}
Mingbao Lin, Mengzhao Chen, Yuxin Zhang, Ke Li, Yunhang Shen, Chunhua Shen, and
  Rongrong Ji.
\newblock Super vision transformer.
\newblock {\em arXiv preprint arXiv:2205.11397}, 2022.

\bibitem{COCO}
Tsung-Yi Lin, Michael Maire, Serge Belongie, James Hays, Pietro Perona, Deva
  Ramanan, Piotr Doll{\'a}r, and C.~Lawrence Zitnick.
\newblock Microsoft coco: Common objects in context.
\newblock In David Fleet, Tomas Pajdla, Bernt Schiele, and Tinne Tuytelaars,
  editors, {\em Proceedings of the European Conference on Computer Vision
  (ECCV)}, Cham, 2014. Springer International Publishing.

\bibitem{FQViT_2022}
Yang Lin, Tianyu Zhang, Peiqin Sun, Zheng Li, and Shuchang Zhou.
\newblock Fq-vit: Post-training quantization for fully quantized vision
  transformer.
\newblock In {\em Proceedings of the Thirty-First International Joint
  Conference on Artificial Intelligence, {IJCAI-22}}, pages 1173--1179, 2022.

\bibitem{PatchDropout_2023}
Yue Liu, Christos Matsoukas, Fredrik Strand, Hossein Azizpour, and Kevin Smith.
\newblock Patchdropout: Economizing vision transformers using patch dropout.
\newblock In {\em Proceedings of the IEEE/CVF Winter Conference on Applications
  of Computer Vision (WACV)}, pages 3953--3962, January 2023.

\bibitem{SWIN_2021}
Ze Liu, Yutong Lin, Yue Cao, Han Hu, Yixuan Wei, Zheng Zhang, Stephen Lin, and
  Baining Guo.
\newblock Swin transformer: Hierarchical vision transformer using shifted
  windows.
\newblock In {\em Proceedings of the IEEE/CVF International Conference on
  Computer Vision (ICCV)}, 2021.

\bibitem{DiversityEVIT_2022}
Sifan Long, Zhen Zhao, Jimin Pi, Shengsheng Wang, and Jingdong Wang.
\newblock Beyond attentive tokens: Incorporating token importance and diversity
  for efficient vision transformers.
\newblock {\em arXiv preprint arXiv:2211.11315}, 2022.

\bibitem{CosineLR}
Ilya Loshchilov and Frank Hutter.
\newblock {SGDR}: Stochastic gradient descent with warm restarts.
\newblock In {\em International Conference on Learning Representations}, 2017.

\bibitem{AdamW}
Ilya Loshchilov and Frank Hutter.
\newblock Decoupled weight decay regularization.
\newblock In {\em International Conference on Learning Representations}, 2019.

\bibitem{Marin_2023}
Dmitrii Marin, Jen-Hao~Rick Chang, Anurag Ranjan, Anish Prabhu, Mohammad
  Rastegari, and Oncel Tuzel.
\newblock Token pooling in vision transformers for image classification.
\newblock In {\em Proceedings of the IEEE/CVF Winter Conference on Applications
  of Computer Vision (WACV)}, pages 12--21, January 2023.

\bibitem{AdaVIT_2022}
Lingchen Meng, Hengduo Li, Bor-Chun Chen, Shiyi Lan, Zuxuan Wu, Yu-Gang Jiang,
  and Ser-Nam Lim.
\newblock Adavit: Adaptive vision transformers for efficient image recognition.
\newblock In {\em Proceedings of the IEEE/CVF Conference on Computer Vision and
  Pattern Recognition (CVPR)}, pages 12309--12318, June 2022.

\bibitem{CCA_2018}
Ari Morcos, Maithra Raghu, and Samy Bengio.
\newblock Insights on representational similarity in neural networks with
  canonical correlation.
\newblock In S. Bengio, H. Wallach, H. Larochelle, K. Grauman, N. Cesa-Bianchi,
  and R. Garnett, editors, {\em Advances in Neural Information Processing
  Systems}, volume~31. Curran Associates, Inc., 2018.

\bibitem{IARED2_2021}
Bowen Pan, Rameswar Panda, Yifan Jiang, Zhangyang Wang, Rogerio Feris, and Aude
  Oliva.
\newblock Ia-red\^{}2: Interpretability-aware redundancy reduction for vision
  transformers.
\newblock In M. Ranzato, A. Beygelzimer, Y. Dauphin, P.S. Liang, and J.~Wortman
  Vaughan, editors, {\em Advances in Neural Information Processing Systems},
  volume~34, pages 24898--24911. Curran Associates, Inc., 2021.

\bibitem{ViTSeeCNN}
Maithra Raghu, Thomas Unterthiner, Simon Kornblith, Chiyuan Zhang, and Alexey
  Dosovitskiy.
\newblock Do vision transformers see like convolutional neural networks?
\newblock In A. Beygelzimer, Y. Dauphin, P. Liang, and J.~Wortman Vaughan,
  editors, {\em Advances in Neural Information Processing Systems}, 2021.

\bibitem{DyViT_2021}
Yongming Rao, Wenliang Zhao, Benlin Liu, Jiwen Lu, Jie Zhou, and Cho-Jui Hsieh.
\newblock Dynamicvit: Efficient vision transformers with dynamic token
  sparsification.
\newblock In A. Beygelzimer, Y. Dauphin, P. Liang, and J.~Wortman Vaughan,
  editors, {\em Advances in Neural Information Processing Systems}, 2021.

\bibitem{PatchMerger_2022}
Cedric Renggli, André~Susano Pinto, Neil Houlsby, Basil Mustafa, Joan
  Puigcerver, and Carlos Riquelme.
\newblock Learning to merge tokens in vision transformers.
\newblock {\em arXiv preprint arXiv:2202.12015}, 2022.

\bibitem{ASL}
Tal Ridnik, Emanuel Ben-Baruch, Nadav Zamir, Asaf Noy, Itamar Friedman, Matan
  Protter, and Lihi Zelnik-Manor.
\newblock Asymmetric loss for multi-label classification.
\newblock In {\em Proceedings of the IEEE/CVF International Conference on
  Computer Vision (ICCV)}, October 2021.

\bibitem{Homogeneity}
Andrew Rosenberg and Julia Hirschberg.
\newblock {V}-measure: A conditional entropy-based external cluster evaluation
  measure.
\newblock In {\em Proceedings of the 2007 Joint Conference on Empirical Methods
  in Natural Language Processing and Computational Natural Language Learning
  ({EMNLP}-{C}o{NLL})}, pages 410--420, Prague, Czech Republic, June 2007.
  Association for Computational Linguistics.

\bibitem{ILSVRC}
Olga Russakovsky, Jia Deng, Hao Su, Jonathan Krause, Sanjeev Satheesh, Sean Ma,
  Zhiheng Huang, Andrej Karpathy, Aditya Khosla, Michael Bernstein,
  Alexander~C. Berg, and Li Fei-Fei.
\newblock {ImageNet} large scale visual recognition challenge.
\newblock {\em International Journal of Computer Vision}, 115(3):211--252, Apr.
  2015.

\bibitem{TokenLearner_2022}
Michael~S Ryoo, AJ Piergiovanni, Anurag Arnab, Mostafa Dehghani, and Anelia
  Angelova.
\newblock Tokenlearner: Adaptive space-time tokenization for videos.
\newblock In A. Beygelzimer, Y. Dauphin, P. Liang, and J.~Wortman Vaughan,
  editors, {\em Advances in Neural Information Processing Systems}, 2021.

\bibitem{CP-ViT}
Zhuoran Song, Yihong Xu, Zhezhi He, Li Jiang, Naifeng Jing, and Xiaoyao Liang.
\newblock Cp-vit: Cascade vision transformer pruning via progressive sparsity
  prediction.
\newblock {\em arXiv preprint arXiv:2203.04570}, 2022.

\bibitem{NMI}
Alexander Strehl and Joydeep Ghosh.
\newblock Cluster ensembles --- a knowledge reuse framework for combining
  multiple partitions.
\newblock {\em J. Mach. Learn. Res.}, 3:583–617, mar 2003.

\bibitem{DPSViT_2022}
Yehui Tang, Kai Han, Yunhe Wang, Chang Xu, Jianyuan Guo, Chao Xu, and Dacheng
  Tao.
\newblock Patch slimming for efficient vision transformers.
\newblock In {\em Proceedings of the IEEE/CVF Conference on Computer Vision and
  Pattern Recognition (CVPR)}, pages 12155--12164, 2022.

\bibitem{DeiT_2021}
Hugo Touvron, Matthieu Cord, Matthijs Douze, Francisco Massa, Alexandre
  Sablayrolles, and Herve Jegou.
\newblock Training data-efficient image transformers \& distillation through
  attention.
\newblock In Marina Meila and Tong Zhang, editors, {\em Proceedings of the 38th
  International Conference on Machine Learning}, volume 139 of {\em Proceedings
  of Machine Learning Research}, pages 10347--10357. PMLR, 18--24 Jul 2021.

\bibitem{NABirds}
Grant Van~Horn, Steve Branson, Ryan Farrell, Scott Haber, Jessie Barry, Panos
  Ipeirotis, Pietro Perona, and Serge Belongie.
\newblock Building a bird recognition app and large scale dataset with citizen
  scientists: The fine print in fine-grained dataset collection.
\newblock In {\em Proceedings of the IEEE Conference on Computer Vision and
  Pattern Recognition (CVPR)}, June 2015.

\bibitem{Transformer_2017}
Ashish Vaswani, Noam Shazeer, Niki Parmar, Jakob Uszkoreit, Llion Jones,
  Aidan~N Gomez, \L~ukasz Kaiser, and Illia Polosukhin.
\newblock Attention is all you need.
\newblock In I. Guyon, U.~Von Luxburg, S. Bengio, H. Wallach, R. Fergus, S.
  Vishwanathan, and R. Garnett, editors, {\em Advances in Neural Information
  Processing Systems}, volume~30. Curran Associates, Inc., 2017.

\bibitem{ViTSUpervision}
Matthew Walmer, Saksham Suri, Kamal Gupta, and Abhinav Shrivastava.
\newblock Teaching matters: Investigating the role of supervision in vision
  transformers.
\newblock In {\em Proceedings of the IEEE/CVF Conference on Computer Vision and
  Pattern Recognition (CVPR)}, 2023.

\bibitem{MAWS_2021}
Jun Wang, Xiaohan Yu, and Yongsheng Gao.
\newblock Feature fusion vision transformer for fine-grained visual
  categorization.
\newblock In {\em British Machine Vision Conference}, 2021.

\bibitem{MSViT_2022}
Yunke Wang, Bo Du, and Chang Xu.
\newblock Multi-tailed vision transformer for efficient inference.
\newblock {\em arXiv preprint arXiv:2203.01587}, 2022.

\bibitem{DVT_2021}
Yulin Wang, Rui Huang, Shiji Song, Zeyi Huang, and Gao Huang.
\newblock Not all images are worth 16x16 words: Dynamic transformers for
  efficient image recognition.
\newblock In A. Beygelzimer, Y. Dauphin, P. Liang, and J.~Wortman Vaughan,
  editors, {\em Advances in Neural Information Processing Systems}, 2021.

\bibitem{R2Trans_2022}
Yu Wang, Shuo Ye, Shujian Yu, and Xinge You.
\newblock R2-trans:fine-grained visual categorization with redundancy
  reduction.
\newblock {\em arXiv preprint arXiv:2204.10095}, 2022.

\bibitem{CentroidViT_2021}
Lemeng Wu, Xingchao Liu, and Qiang Liu.
\newblock Centroid transformers: Learning to abstract with attention, 2021.

\bibitem{GroupViT_2022}
Jiarui Xu, Shalini De~Mello, Sifei Liu, Wonmin Byeon, Thomas Breuel, Jan Kautz,
  and Xiaolong Wang.
\newblock Groupvit: Semantic segmentation emerges from text supervision.
\newblock In {\em Proceedings of the IEEE/CVF Conference on Computer Vision and
  Pattern Recognition (CVPR)}, pages 18134--18144, June 2022.

\bibitem{EVoViT_2022}
Yifan Xu, Zhijie Zhang, Mengdan Zhang, Kekai Sheng, Ke Li, Weiming Dong, Liqing
  Zhang, Changsheng Xu, and Xing Sun.
\newblock Evo-vit: Slow-fast token evolution for dynamic vision transformer.
\newblock In {\em Proceedings of the AAAI Conference on Artificial
  Intelligence}, volume~36, pages 2964--2972, 2022.

\bibitem{AViT_2022}
Hongxu Yin, Arash Vahdat, Jose Alvarez, Arun Mallya, Jan Kautz, and Pavlo
  Molchanov.
\newblock {A}-{V}i{T}: {A}daptive tokens for efficient vision transformer.
\newblock In {\em Proceedings of the IEEE/CVF Conference on Computer Vision and
  Pattern Recognition}, 2022.

\bibitem{UPViT_2021}
Hao Yu and Jianxin Wu.
\newblock A unified pruning framework for vision transformers.
\newblock {\em arXiv preprint arXiv:2111.15127}, 2021.

\bibitem{PSViT_2021}
Xiaoyu Yue, Shuyang Sun, Zhanghui Kuang, Meng Wei, Philip Torr, Wayne Zhang,
  and Dahua Lin.
\newblock Vision transformer with progressive sampling.
\newblock {\em Proceedings of the IEEE/CVF International Conference on Computer
  Vision (ICCV)}, 2021.

\bibitem{CutMix}
Sangdoo Yun, Dongyoon Han, Seong~Joon Oh, Sanghyuk Chun, Junsuk Choe, and
  Youngjoon Yoo.
\newblock Cutmix: Regularization strategy to train strong classifiers with
  localizable features.
\newblock In {\em Proceedings of the IEEE/CVF International Conference on
  Computer Vision (ICCV)}, October 2019.

\bibitem{DPCKNN_2022}
Wang Zeng, Sheng Jin, Wentao Liu, Chen Qian, Ping Luo, Wanli Ouyang, and
  Xiaogang Wang.
\newblock Not all tokens are equal: Human-centric visual analysis via token
  clustering transformer.
\newblock In {\em Proceedings of the IEEE/CVF Conference on Computer Vision and
  Pattern Recognition}, pages 11101--11111, 2022.

\bibitem{MixUp}
Hongyi Zhang, Moustapha Cisse, Yann~N. Dauphin, and David Lopez-Paz.
\newblock mixup: Beyond empirical risk minimization.
\newblock In {\em International Conference on Learning Representations}, 2018.

\bibitem{RandomErasing}
Zhun Zhong, Liang Zheng, Guoliang Kang, Shaozi Li, and Yi Yang.
\newblock Random erasing data augmentation.
\newblock {\em Proceedings of the AAAI Conference on Artificial Intelligence},
  34(07):13001--13008, Apr. 2020.

\bibitem{ReViT_2021}
Yichen Zhu, Yuqin Zhu, Jie Du, Yi Wang, Zhicai Ou, Feifei Feng, and Jian Tang.
\newblock Make a long image short: Adaptive token length for vision
  transformers.
\newblock {\em arXiv preprint arXiv:2112.01686}, 2021.

\bibitem{SiT_2022}
Zhuofan Zong, Kunchang Li, Guanglu Song, Yali Wang, Yu Qiao, Biao Leng, and Yu
  Liu.
\newblock Self-slimmed vision transformer.
\newblock In {\em Proceedings of the European Conference on Computer Vision
  (ECCV)}, 2022.

\end{thebibliography}
}
\end{document}